\documentclass[10pt,twocolumn,letterpaper]{article}

\usepackage[pagenumbers]{cvpr} 

\usepackage{graphicx}
\usepackage{amsmath}
\usepackage{amssymb}
\usepackage{booktabs}
\usepackage{hhline}
\usepackage{mathtools}
\makeatletter
\@namedef{ver@everyshi.sty}{}
\makeatother
\usepackage{tikz}

\usepackage{siunitx}
\newcolumntype{P}[1]{>{\centering\arraybackslash}p{#1}}
\newcolumntype{M}[1]{>{\centering\arraybackslash}m{#1}}

\DeclarePairedDelimiter\norm{\lVert}{\rVert}%

\usepackage[pagebackref,breaklinks,colorlinks]{hyperref}

\usepackage[capitalize]{cleveref}
\crefname{section}{Sec.}{Secs.}
\Crefname{section}{Section}{Sections}
\Crefname{table}{Table}{Tables}
\crefname{table}{Tab.}{Tabs.}

\def\cvprPaperID{5022} 
\def\confName{CVPR}
\def\confYear{2022}

\begin{document}

\title{Generalizing Interactive Backpropagating Refinement for Dense Prediction Networks}

\author{Fanqing Lin\\
Brigham Young University\\
{\tt\small flin2@byu.edu}
\and
Brian Price\\
Adobe Research\\
{\tt\small bprice@adobe.com}
\and 
Tony Martinez\\
Brigham Young University\\
{\tt\small martinez@cs.byu.edu}
}
\maketitle

\begin{abstract}
As deep neural networks become the state-of-the-art approach in the field of computer vision for dense prediction tasks, many methods have been developed for automatic estimation of the target outputs given the visual inputs. Although the estimation accuracy of the proposed automatic methods continues to improve, interactive refinement is oftentimes necessary for further correction. Recently, feature backpropagating refinement scheme~\cite{Sofiiuk20} (\text{\textit{f}-BRS}) has been proposed for the task of interactive segmentation, which enables efficient optimization of a small set of auxiliary variables inserted into the pretrained network to produce object segmentation that better aligns with user inputs. However, the proposed auxiliary variables only contain channel-wise scale and bias, limiting the optimization to global refinement only. In this work, in order to generalize backpropagating refinement for a wide range of dense prediction tasks, we introduce a set of G-BRS (Generalized Backpropagating Refinement Scheme) layers that enable both global and localized refinement for the following tasks: interactive segmentation, semantic segmentation, image matting and monocular depth estimation. Experiments on SBD, Cityscapes, Mapillary Vista, Composition-1k and NYU-Depth-V2 show that our method can successfully generalize and significantly improve performance of existing pretrained state-of-the-art models with only a few clicks.
\end{abstract}
\section{Introduction}\label{sec:introduction}
Deep learning has revolutionized the task of dense prediction, allowing a breakthrough for pixel-classification problems such as semantic segmentation~\cite{Long15,Lin17,Zhao17,Chen17} and pixel-regression problems such as depth estimation~\cite{Fu14, Eigen14,Eigen15,Laina16}. While these automatic methods are constantly improving in performance, a user has no resource to make corrections on the estimated output other than using external tools that do not leverage any learned features. To enable user interactions, dense prediction tasks such as interactive segmentation~\cite{Xu16,Le18,Liew17,Mahadevan18,Hu19} and image matting~\cite{Xu17,Lutz18,Cai19,Zhang19,Forte20} use user inputs in forms of distance maps and trimap respectively as network input. Although the additional information can be helpful during forward propagation, deep networks are still free to generate predictions inconsistent with the user-provided inputs. \\ 
\indent In this work, we investigate whether a pretrained automatic dense prediction method can be effectively converted into an efficient interactive method without any additional retraining. This is a significant task as deep networks are commonly applied in interactive ways for photography~\cite{Zhang17, XHuang17, Zhu17, Zhou19, Yu19}, videography~\cite{WugOh19, Miao20, Xu19}, special effects~\cite{Haouchine18, Thies19, Fried19}, etc. Two prior works, both focused primarily on interactive segmentation, have inspired our method. Backpropagating Refinement Scheme (BRS)~\cite{Jang19} performs interactive segmentation using an initial forward pass given the input image and distance maps generated from a set of clicks as in~\cite{Xu16}. To additionally refine the prediction and encourage consistency with the input clicks, it sets the input distance maps as the trainable parameters and performs backpropagation using loss computed from the prediction and the clicked labels. BRS also briefly extends this idea to a few other applications: semantic segmentation, saliency detection and medical image segmentation, showing potential use of BRS for CNNs in general. A follow-on work, \text{\textit{f}-BRS}~\cite{Sofiiuk20} later argues that due to the need for online backpropagation through the entire network, BRS has slow inference speed and is computationally expensive. To this end, instead of using the input distance maps as trainable parameters, \text{\textit{f}-BRS} inserts a pair of auxiliary parameters that act as channel-wise scale and bias after an intermediate network layer, requiring backpropagation through a subpart of the network while achieving nearly equivalent performance. \\
\indent Despite the improved efficiency of \text{\textit{f}-BRS}, it comes with a major disadvantage: the proposed auxiliary channel-wise scale and bias are only capable of global modification. This not only neglects the need for localized refinement in many vision applications, but also makes the modified output susceptible to undesired global changes while correcting for existing clicks. To make efficient and effective refinement generalized for dense prediction models, we propose to expand the idea of auxiliary channel-wise scale and bias to a set of G-BRS (Generalized Backpropagating Refinement Scheme) layers with more advanced layer architectures. Our approach enables both global and localized refinement using a channel-weighted bias map in various settings. In addition, we propose a novel consistency loss with an attention mechanism that stabilizes the refinement process and enables more user control. To demonstrate the generality of our approach, we implement G-BRS on four state-of-the-art models for a wide range of dense prediction tasks including interactive segmentation, semantic segmentation, image matting and depth estimation. We perform thorough evaluation on five benchmark datasets: SBD, Cityscapes, Mapillary Vista, Composition-1k and NYU-Depth-V2. Results show that our method enables existing models to achieve significant improvement with interactive clicks and opens up promising directions for equipping automatic methods with interactive features in general. 
\section{Method}\label{sec:method}
\subsection{Background}
\noindent\textbf{Backpropagating refinement scheme.} BRS was initially proposed by Jang \etal~\cite{Jang19} for interactive segmentation, which is a task to segment the foreground object and the background given the user inputs. First, the input clicks are used to generate the foreground and background interaction map using distance transform. At inference time, the input image concatenated with the interaction maps is forward propagated in a CNN to produce an output segmentation. Although information of the clicked locations are encoded in the input interaction maps, it is possible that the annotated locations are still mislabeled in the output segmentation. To address this issue, BRS proposes to use backpropagation to refine the input interaction maps to enforce consistency between the input clicks and output segmentation. An alternative method of finetuning the entire model is not ideal since it is computationally inefficient and the model would lose the pretrained knowledge needed for intelligent refinement. With the network defined as $f$, given a set of input clicks $\{({u}_{i}, {v}_{i}, {l}_{i})\}_{i=1}^{n}$ where $(u, v)$ and $l \in \{0, 1\}$ denote the clicked location and label respectively, BRS refines the initial interaction maps $x$ by solving for $\Delta x$ in the following optimization problem: \\
\begin{equation}\label{eq:brs}
E(x) = \min_{\Delta x} \left(\lambda\norm{\Delta x}_{2}  + \sum\limits_{i=1}^n (f(x + \Delta x)_{{u}_{i}, {v}_{i}} - {l}_{i})^2\right).
\end{equation}
The first term represents the inertial energy used to prevent excessive modification, where $\lambda$ is a scaling constant that regulates the trade-off. The second term represents the corrective energy used to enforce correct output segmentation at the clicked locations.\\
\noindent\textbf{Feature Backpropagating Refinement Scheme.} Despite the improvement in accuracy, BRS is computationally expensive as it requires gradient computation through the entire network. Consequently, Sofiiuk \etal~\cite{Sofiiuk20} propose \text{\textit{f}-BRS} to modify a small set of inserted auxiliary parameters instead of the input interaction maps, leading to a faster algorithm that requires gradient computation through only a small part of the network. It defines $\hat{f}(x, p)$ as the function that accepts the additionally inserted auxiliary parameter $p$. The optimization problem is then presented as the following: \\
\begin{equation}\label{eq:f-brs}
E(x) = \min_{\Delta x} \left(\lambda\norm{\Delta p}_{2}  + \sum\limits_{i=1}^n (\hat{f}(x, p + \Delta p)_{{u}_{i}, {v}_{i}} - {l}_{i})^2\right).
\end{equation}
To avoid minor localized refinement near the clicked locations and encourage global refinement, Sofiiuk \etal propose to use channel-wise scale $s \in \mathbb{R}^{\mathcal{C}}$ and bias $b \in \mathbb{R}^{\mathcal{C}}$ as the auxiliary parameters, where $\mathcal{C}$ denotes the number of channels for the corresponding intermediate feature map of the network. Let us define the inserted auxiliary parameters as a G-BRS layer. The proposed layer that utilizes channel-wise scale and bias can then be formulated as, 
\begin{equation}\label{eq:g-brs-sb}
\mathcal{G}_{sb}(m) = m \ \dot{\times} \ s \ \dot{+} \ b
\end{equation}
where $m \in \mathbb{R}^{\mathcal{H} \times \mathcal{W} \times \mathcal{C}}$ is the intermediate feature map with $\mathcal{H}$, $\mathcal{W}$ and $\mathcal{C}$ denoting the height, width and number of channels respectively. Channel-wise multiplication and addition are represented as $\dot{\times}$ and $\dot{+}$. Since the inserted G-BRS layer should not interfere with the initial network prediction, its initial parameters need to perform the identity operation such that $\mathcal{G}_{0}(m) = m$. This can be fulfilled with the initialization of $s_{0} = \textbf{1}$ and $b_{0} = \textbf{0}$. We will refer to this G-BRS layer as the G-BRS-sb layer. 
\subsection{Global and Localized Refinement}
As the G-BRS-sb layer enables channel-wise scaling and shifting of the original feature maps, it solely focuses on global refinement since $s$ and $b$ are invariant to position in the selected feature. This limitation can result in unstable and undesirable effects as an attempt to fix a localized error could lead to unpredictable global changes across the image. To additionally enable positional modification of the selected feature map for precise localized refinement, we propose three novel G-BRS layer architectures with better performance on numerous applications below.\\
\indent First, we introduce the G-BRS-bmsb layer that contains an additional bias map ${b}_{m} \in \mathbb{R}^{\mathcal{H} \times \mathcal{W}}$ prior to the channel-wise scale and bias. To enable all channels of the feature map to shift freely in different directions, we also introduce a channel weight variable ${w}_{c} \in \mathbb{R}^{\mathcal{C}}$ to perform channel-wise scaling for the bias map. We formulate the G-BRS-bmsb layer as follows: \\
\begin{equation}\label{eq:g-brs-bmsb}
\mathcal{G}_{bmsb}(m) = (m + ({b}_{m} \ \dot{\times} {w}_{c})) \ \dot{\times} \ s \ \dot{+} \ b
\end{equation}
where $({b}_{m}\ \dot{\times}\ {w}_{c})\in \mathbb{R}^{\mathcal{H} \times \mathcal{W} \times \mathcal{C}}$. Similar to $s$ and $b$, we initialize ${b}_{m}$ as $\textbf{0}$ and ${w}_{c}$ as $\textbf{1}$. Since the size of ${b}_{m}$ depends on the resolution of the selected feature map, we apply the G-BRS insertion(s) in deeper feature space where the feature resolution is a fraction of the output resolution. This setting also prevents the aforementioned drawback that leads to trivial localized refinement.\\
\indent As the channel-weighted bias map and the channel-wise scale and bias apply localized and global changes respectively, the G-BRS-bmsb layer modifies the input feature through the two variables sequentially. To explore feature fusion where the G-BRS layer merges feature maps from the global branch and the localized branch, we introduce the G-BRS-bmsb-m layer formulated as follows:\\
\begin{equation}\label{eq:g-brs-bmsb-m}
\begin{gathered}
{g}_{1}(m) = m \ \dot{\times} \ s \ \dot{+} \ b \\
{g}_{2}(m) = m + ({b}_{m} \ \dot{\times} {w}_{c})\\
\mathcal{G}_{bmsb-m}(m) = w\cdot{g}_{1}(m) + (1 - w)\cdot{g}_{2}(m)
\end{gathered}
\end{equation}
where $w \in [0, 1]$ is a learnable parameter (initialized to 0.5) used to regulate the trade-off between global and localized changes in the input feature.\\ 
\indent In addition to the channel-wise scale and bias, we explore a more powerful representation by replacing $s$ and $b$ with a convolutional layer, which we refer to as the G-BRS-bmconv layer. For kernel size $k = 1$, the convolutional layer essentially learns to combine features from different input channels for each output channel. With ${\mathcal{C}}_{in} = {\mathcal{C}_{out}}$, we initialize the kernel weight ${w}_{conv} \in \mathbb{R}^{\mathcal{C} \times \mathcal{C} \times 1 \times 1}$ as an identity matrix and the bias ${b}_{conv} \in \mathbb{R}^{\mathcal{C}}$ as $\textbf{0}$. Initially, each output channel represents exactly the corresponding input channel and $\mathcal{G}_{bmconv}(m) = m$. We formulate the G-BRS-bmconv layer as follows: \\
\begin{equation}\label{eq:g-brs-bmconv}
\mathcal{G}_{bmconv}(m) = (m + \beta({b}_{m} \ \dot{\times} {w}_{c})) \cdot {w}_{conv} \ \dot{+} \ {b}_{conv}
\end{equation}
where the $1\times1$ convolutional operation is represented as matrix multiplication and channel-wise bias. $\beta = 10$ is used as a scalar for amplifying the gradient of the bias map.\\
\subsection{Attention Mechanism}
For optimization using backpropagating refinement, intelligent refinement without inaccurate excessive modification is important. Previous methods~\cite{Jang19, Sofiiuk20} propose to rely on the minimization of the inertial energy $\lambda\norm{\Delta p}_{2}$. Instead of simply enforcing a small $\norm{\Delta p}_{2}$, we propose to punish excessive perturbation in the output estimation outside of a user-defined attention region, which becomes achievable with the proposed G-BRS layer capable of both global and localized feature map modification. In the following sections, we define each input click as $(u, v, r, l)$ with $r$ and $l$ denoting the attention radius centered at $(u, v)$ and the target label respectively. We introduce a consistency loss with the following general formulation:\\
\begin{equation}
\mathcal{L}_{c} = {\lambda}\ \mathcal{E}((\hat{f}(x, {p}_{prev}) - \hat{f}(x, p))\ \mathcal{M})
\end{equation}
where $\mathcal{E}$ is a function that computes the pixel-wise error using the current prediction $\hat{f}(x, p)$ and the initial prediction $\hat{f}(x, {p}_{prev})$ with ${p}_{prev}$ denoting the auxiliary variables from the previous click. $\mathcal{M}$ represents a pixel-wise scaling mask generated using the newest click, which selects the region outside of the $r$ for error computation. In all our experiments, we perform backpropagation for $I = 20$ iterations.
\subsection{Generalization}
In this work, we use existing pretrained state-of-the-art architectures for a wide range of dense prediction problems. The selected applications include binary-label (interactive segmentation) and multi-label (semantic segmentation) pixel-wise segmentation tasks, bounded (interactive image matting) and unbounded (depth estimation) pixel-wise regression tasks. Our goal is to demonstrate the generality of our approach in both interactive and automatic settings for dense prediction models.\\
\indent We introduce the corresponding G-BRS layer configuration for each architecture. Options for multiple G-BRS layer insertions are explored to leverage combination of feature modification at different levels. Since architectures for different tasks also drastically differ, it is worth mentioning that designing an effective G-BRS layout requires thought and experimentation to obtain optimal performance.
\begin{figure*}[t]
 \centering
  \begin{subfigure}[t]{0.24\linewidth}
    \includegraphics[width=\linewidth]{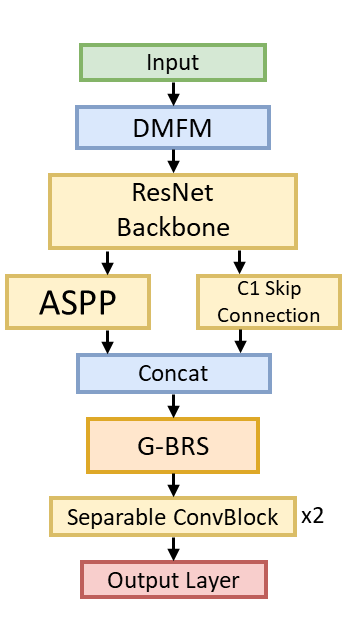}
    \caption{Interactive Segmentation.}\label{fig:fbrs_is}
  \end{subfigure}
  \begin{subfigure}[t]{0.24\linewidth}
    \includegraphics[width=\linewidth]{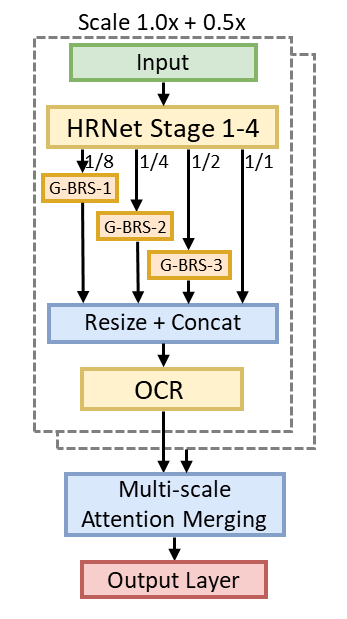}
    \caption{Semantic Segmentation.}\label{fig:fbrs_ss}
  \end{subfigure}
  \begin{subfigure}[t]{0.24\linewidth}
    \includegraphics[width=\linewidth]{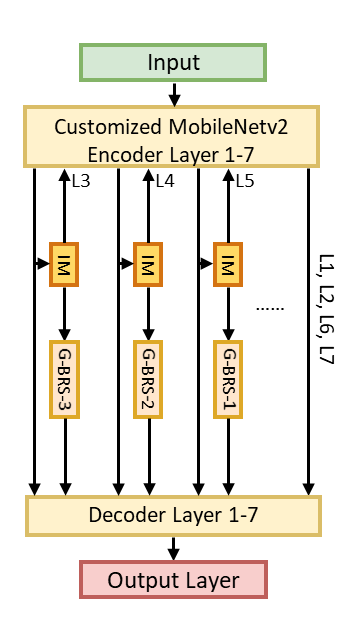}
    \caption{Image Matting.}\label{fig:fbrs_mt}
  \end{subfigure}
  \begin{subfigure}[t]{0.24\linewidth}
    \includegraphics[width=\linewidth]{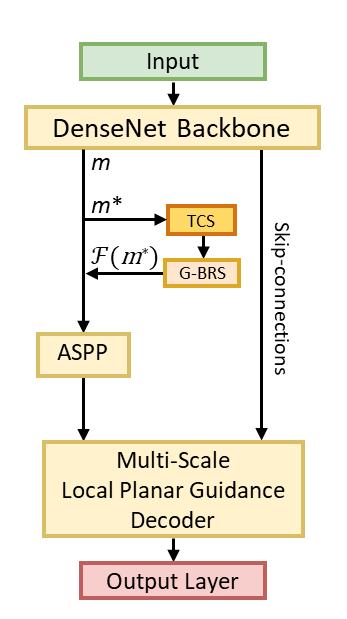}
    \caption{Depth Estimation.}\label{fig:fbrs_de}
  \end{subfigure}
  \caption{G-BRS configurations on four state-of-the-art architectures for various computer vision applications.}
  \label{fig:fbrs_configurations}
  \vspace{-3mm}
\end{figure*}
\subsubsection{Interactive Segmentation}
Interactive segmentation is a binary segmentation task that separates any target foreground object and the background using user inputs. Since prior methods \cite{Jang19, Sofiiuk20} primarily focused on this task, we make a direct comparison with the \text{\textit{f}-BRS}~\cite{Sofiiuk20} (equivalent to G-BRS-sb) layer and use the standard DeepLabV3+ with ResNet-101 and the proposed Distance Maps Fusion Module as the architecture. The G-BRS layer is also inserted at the position shown in Figure \ref{fig:fbrs_is}, where the best performance is reported by~\cite{Sofiiuk20}. We formulate the optimization as a minimization problem for the click refinement loss $\mathcal{L}_{r}$ and the consistency loss $\mathcal{L}_{c}$:
\begin{equation}\label{eq:f-brs-is}
\begin{gathered}
\mathcal{L}_{r} = \sum\limits_{i=1}^n \begin{cases}
max(1 - \hat{f}(x, p)_{{u}_{i}, {v}_{i}}, 0)^{2} & {l}_{i} = 1\\
max(1 + \hat{f}(x, p)_{{u}_{i}, {v}_{i}}, 0)^{2} & {l}_{i} = -1\\
\end{cases}\\
\mathcal{L}_{c} = {\lambda}_{is} \frac{1}{HW} \norm{(\hat{f}(x, {p}_{prev}) - \hat{f}(x, p))\ \mathcal{M}}_{2}^{2}
\end{gathered}
\end{equation}
Since the selected architecture produces unbounded output values, $\mathcal{L}_{r}$ enables backpropagation only on positive clicks with values less than $1$ and negative clicks with values larger than $-1$, allowing positive predictions to exceed $1$ and vice versa. $\mathcal{L}_{c}$ uses the Mean Squared Error (MSE) and punishes excessive output deviation outside of the attention region. $\mathcal{M} \in \{0, 1\}^{H \times W}$ defines a binary attention mask with the value of 0 within the circular attention region. We use ${\lambda}_{is} = \num{1e2}$ as the weight for this term.\\
\indent For each click, the network makes an inference using the updated interaction maps and performs backpropagation. Note that all provided clicks are used for $\mathcal{L}_{r}$ while only the most recent click is used for $\mathcal{L}_{c}$. Using all clicks in the computation of $\mathcal{L}_{r}$ allows correction for the newly provided click without losing knowledge gained from previous clicks. As the threshold for binary segmentation is 0, to avoid overfitting and achieve a faster response time, the refinement does early stopping when $max(\norm{{l}_{i} - \hat{f}(x, p)_{{u}_{i}, {v}_{i}}}_{1}: i = 1, ..., n) < 0.8$. 
\subsubsection{Semantic Segmentation}
Semantic segmentation is a multi-label segmentation task with predefined classes. To enable interactive refinement on the output segmentation, we configure multiple G-BRS layer insertions on the architecture proposed by Tao \etal~\cite{Tao20}, a multi-scale attention network with HRNet-OCR~\cite{Yuan20} as the backbone. As shown in Figure \ref{fig:fbrs_ss}, we make three insertions in stage $4$ of the HRNet backbone~\cite{Sun19} for each scale branch, where the feature resolution is $\frac{1}{2}$, $\frac{1}{4}$ and $\frac{1}{8}$ of the input resolution. For practicality in user applications using a single GPU, we omit the branch with 2.0x scale and use two branches with 1.0x and 0.5x scale. In addition, we introduce two refinement modes: the click mode and the stroke mode. First, we formulate the optimization problem for the click mode below:\\
\vspace{-2mm}
\begin{equation}\label{eq:g-brs-ss-1}
\begin{gathered}
\mathcal{L}_{r} = \frac{1}{n}\sum\limits_{i=1}^n log \frac{{e}^{\hat{f}(x, p)_{{u}_{i}, {v}_{i}, {c}_{l}}}}{\sum_{c=1}^{C} {e}^{\hat{f}(x, p)_{{u}_{i}, {v}_{i}, {c}}}} \\
\mathcal{L}_{c} = {\lambda}_{ss}\frac{1}{HW} \sum\limits_{h=1}^H\sum\limits_{w=1}^W log \frac{{e}^{\hat{f}(x, p)_{h, w, {c}_{p}}}}{\sum_{c=1}^{C} {e}^{\hat{f}(x, p)_{h, w, {c}}}}
\end{gathered}
\end{equation}
We compute the cross entropy loss for both $\mathcal{L}_{r}$ and $\mathcal{L}_{c}$ with $\mathcal{L}_{r}$ only using the clicked locations. $C$, ${c}_{l}$ and ${c}_{p}$ denote the number of classes, the clicked target class and the previously predicted class respectively. ${c}_{p}$ is set as the ignored label within the circular attention region for the computation of $\mathcal{L}_{c}$. \\
\indent In the stroke mode, we enable the user to draw strokes for different target classes with arbitrary radius and create a finetune mask $\mathcal{T} \in \{0, ..., C\}^{H \times W}$, where the value of $C$ is used as the ignored label for initialization. In this mode, we update $\mathcal{L}_{r}$ in Equation \ref{eq:g-brs-ss-1} as:\\
\begin{equation}\label{eq:f-brs-ss-2}
\mathcal{L}_{r} = \frac{1}{HW} \sum\limits_{h=1}^H\sum\limits_{w=1}^W log \frac{{e}^{\hat{f}(x, p)_{h, w, {c}_{\mathcal{T}}}}}{\sum_{c=1}^{C} {e}^{\hat{f}(x, p)_{h, w, {c}}}}\text{.}
\end{equation}
For the weighting of $\mathcal{L}_{c}$, we use ${\lambda}_{ss} = 10$ for the click mode and ${\lambda}_{ss} = 1$ for the stroke mode.
\subsubsection{Image Matting}
Image matting is a task to predict dense alpha matte for the target foreground given the input image and the user-defined trimap. Although interactive refinement can be performed by modifying the trimap, such modification does not guarantee an output matte consistent with the trimap. More importantly, the input trimap lacks the necessary precision for alpha values as it only contains three labels that denote the foreground, the background and the unsure region. To enable backpropagating refinement, IndexNet~\cite{Lu19} with the backbone of MobileNetv2~\cite{Sandler18} is selected as the architecture. We observe that the index maps generated by the IndexNet Module (IM) for the decoder layers contain the best features and insert the G-BRS layers as shown in Figure \ref{fig:fbrs_mt}. The optimization problem is formulated as follows:\\
\begin{equation}\label{eq:f-brs-mt-1}
\begin{gathered}
\mathcal{L}_{r} = \frac{1}{n}\sum\limits_{i=1}^n ({l}_{i} - \hat{f}(x, p)_{{u}_{i}, {v}_{i}})^2\\
\mathcal{L}_{c} = {\lambda}_{mt} \frac{1}{HW} \norm{(\hat{f}(x, {p}_{prev}) - \hat{f}(x, p))\ \mathcal{M}}_{2}^{2}
\end{gathered}
\end{equation}
where ${l}_{i} \in [0, 1]$ represents the target alpha value for click $i$. MSE loss is computed for $\mathcal{L}_{r}$ and $\mathcal{L}_{c}$ with $\mathcal{L}_{r}$ only using the clicked locations. $\mathcal{L}_{c}$ punishes perturbation far from the attention region using an element-wise weighting mask $\mathcal{M}$, which is defined using an inverse gaussian kernel at the newest clicked location with $\sigma = r$. ${\lambda}_{mt} = \num{1e3}$ is used as the weight for $\mathcal{L}_{c}$. We refer to this refinement mode as the click mode.\\
\indent Since it is challenging for the user to determine the exact alpha value for the target pixels, for practicality, we introduce the push mode that allows the user to left/right click to push the alpha values up/down. We define $l \in \{0, 1\}$ for the left/right click and formulate the optimization problem as below:\\
\begin{equation}\label{eq:f-brs-mt-2}
\begin{gathered}
\mathcal{L}_{r} = \begin{cases}
((\hat{f}(x, {p}_{prev})_{u, v} + \epsilon) - \hat{f}(x, p)_{u, v})^2 & l = 1\\
((\hat{f}(x, {p}_{prev})_{u, v} - \epsilon) - \hat{f}(x, p)_{u, v})^2 & l = 0\\
\end{cases}
\end{gathered}
\end{equation}
where $\epsilon = 0.1$ denotes the push distance. The push mode contains no memory of previous clicks and omits $\mathcal{L}_{c}$. Backpropagation is applied for only 1 iteration since the required modification is marginal.
\subsubsection{Depth Estimation}
Depth estimation is a task to produce an accurate depth map from a single image. To enable interactive refinement, we select BTSNet~\cite{Lee19} with the backbone of DenseNet-161~\cite{GHuang17} as the architecture. We insert the G-BRS layer after the final DenseNet Block of the encoder as shown in Figure \ref{fig:fbrs_de}. Since feature map $m$ at this location has a large number of channels $\mathcal{C} = 2208$, applying G-BRS on excessive number of parameters can lead to overfitting on the target clicks. Additionally, a large $\mathcal{C}$ is inefficient for the G-BRS-bmconv layer with the parameter ${w}_{conv} \in \mathbb{R}^{\mathcal{C} \times \mathcal{C} \times 1 \times 1}$. To this end, we perform top-$k$ channel selection (TCS) that selects the $\mathcal{K} = 256$ channels of $m$ with the highest mean activation for G-BRS. The resulting selected feature map ${m}^{\ast} \in \mathbb{R}^{\mathcal{H} \times \mathcal{W} \times \mathcal{K}}$ is used as the input for the G-BRS layer and the unselected channels in $m$ are not modified. We formulate the optimization problem as below:\\
\begin{equation}\label{eq:f-brs-de-1}
\mathcal{L}_{r} = \sum\limits_{i=1}^n ({l}_{i} - \hat{f}(x, p)_{{u}_{i}, {v}_{i}})^2
\end{equation}
We compute the Sum Squared Error (SSE) for $\mathcal{L}_{r}$ and formulate $\mathcal{L}_{c}$ the same as Equation \ref{eq:f-brs-mt-1} using ${\lambda}_{de} = \num{1e-1}$. The push mode is also formulated following Equation \ref{eq:f-brs-mt-2}.
\section{Experiments}\label{sec:experiments}
\begin{figure*}[h]
  \centering
   \begin{subfigure}[t]{0.16\linewidth}
    \includegraphics[width=\linewidth]{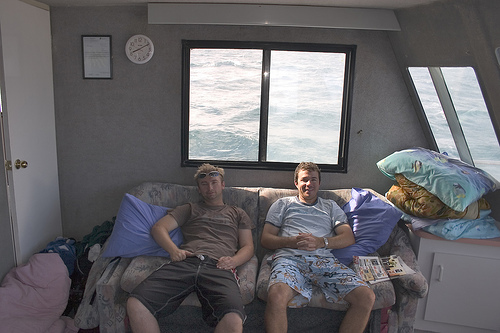}
  \end{subfigure}
  \begin{subfigure}[t]{0.16\linewidth}
    \includegraphics[width=\linewidth]{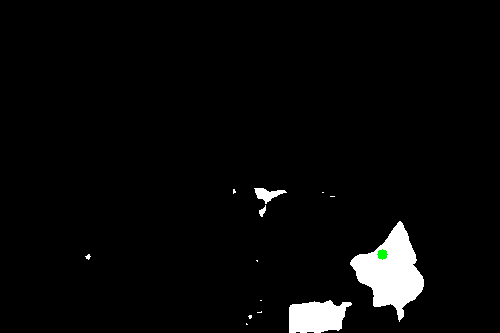}
  \end{subfigure}
  \begin{subfigure}[t]{0.16\linewidth}
    \includegraphics[width=\linewidth]{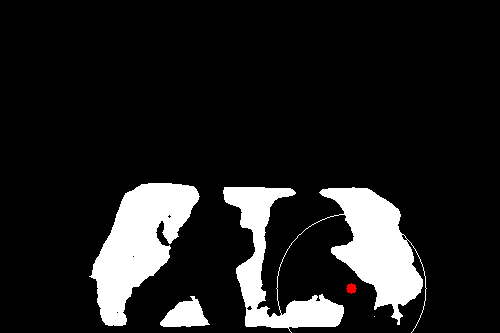}
  \end{subfigure}
  \begin{subfigure}[t]{0.16\linewidth}
    \includegraphics[width=\linewidth]{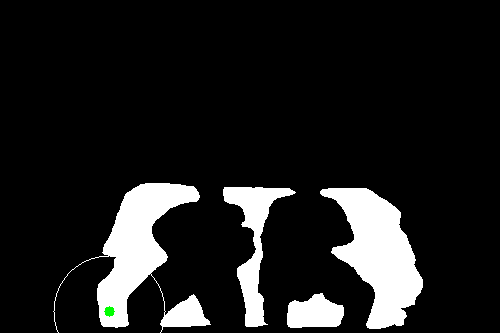}
  \end{subfigure}
  \begin{subfigure}[t]{0.16\linewidth}
    \includegraphics[width=\linewidth]{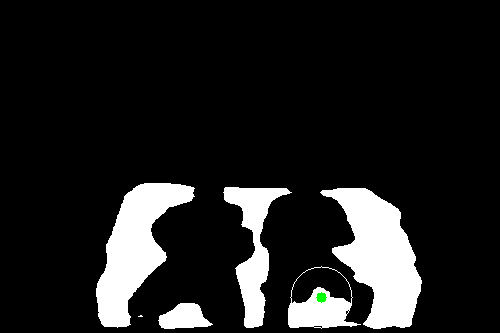}
  \end{subfigure}
  \vspace{1mm}
  \begin{subfigure}[t]{0.16\linewidth}
    \includegraphics[width=\linewidth]{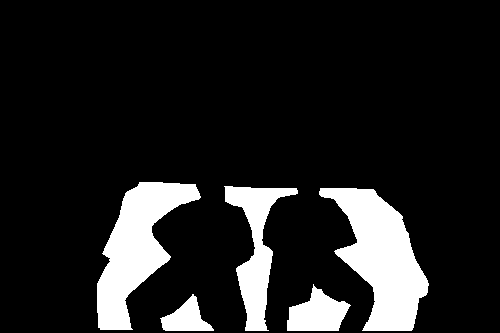}
  \end{subfigure}
   \begin{subfigure}[t]{0.16\linewidth}
    \includegraphics[width=\linewidth]{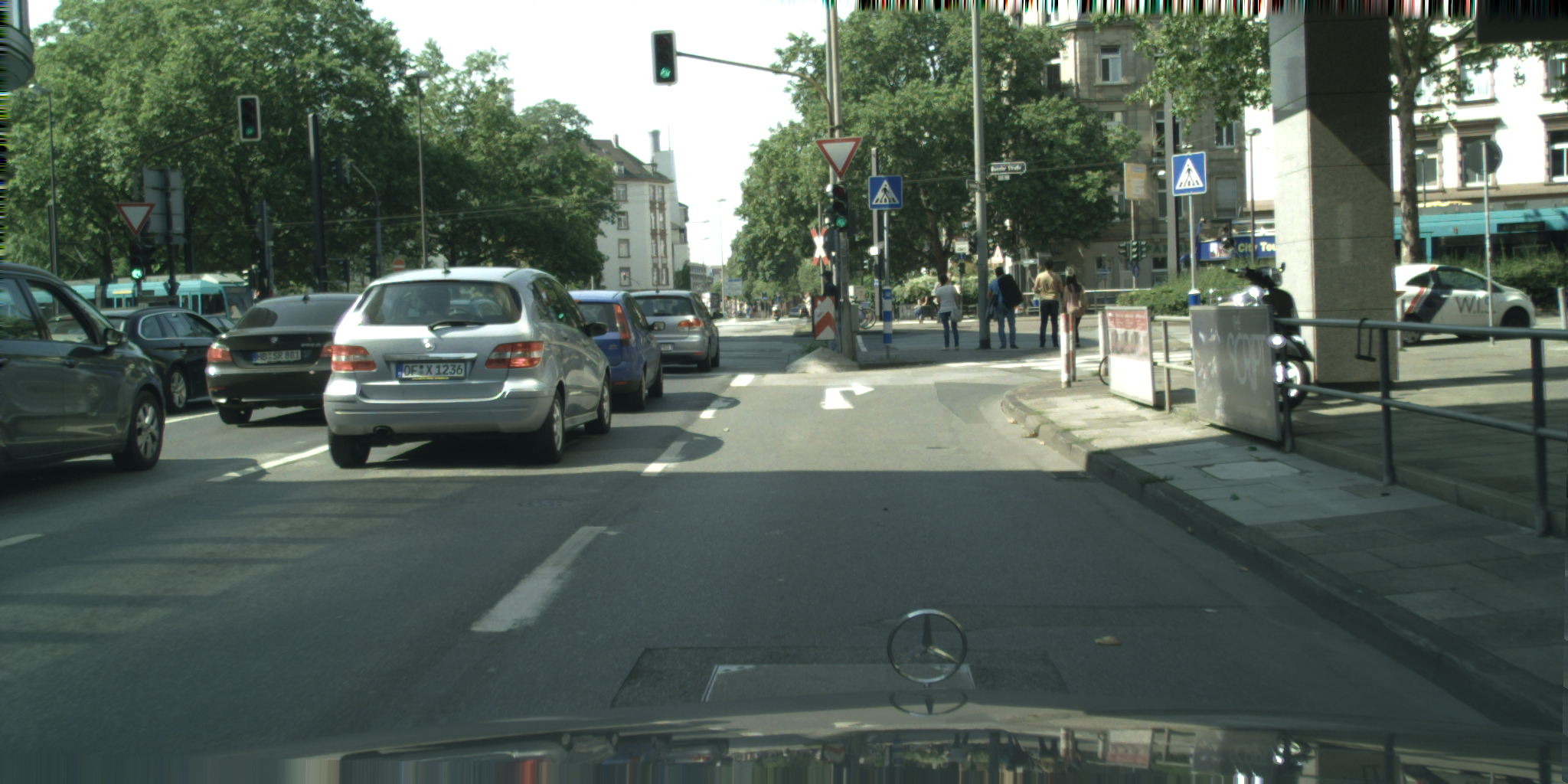}
  \end{subfigure}
  \begin{subfigure}[t]{0.16\linewidth}
    \includegraphics[width=\linewidth]{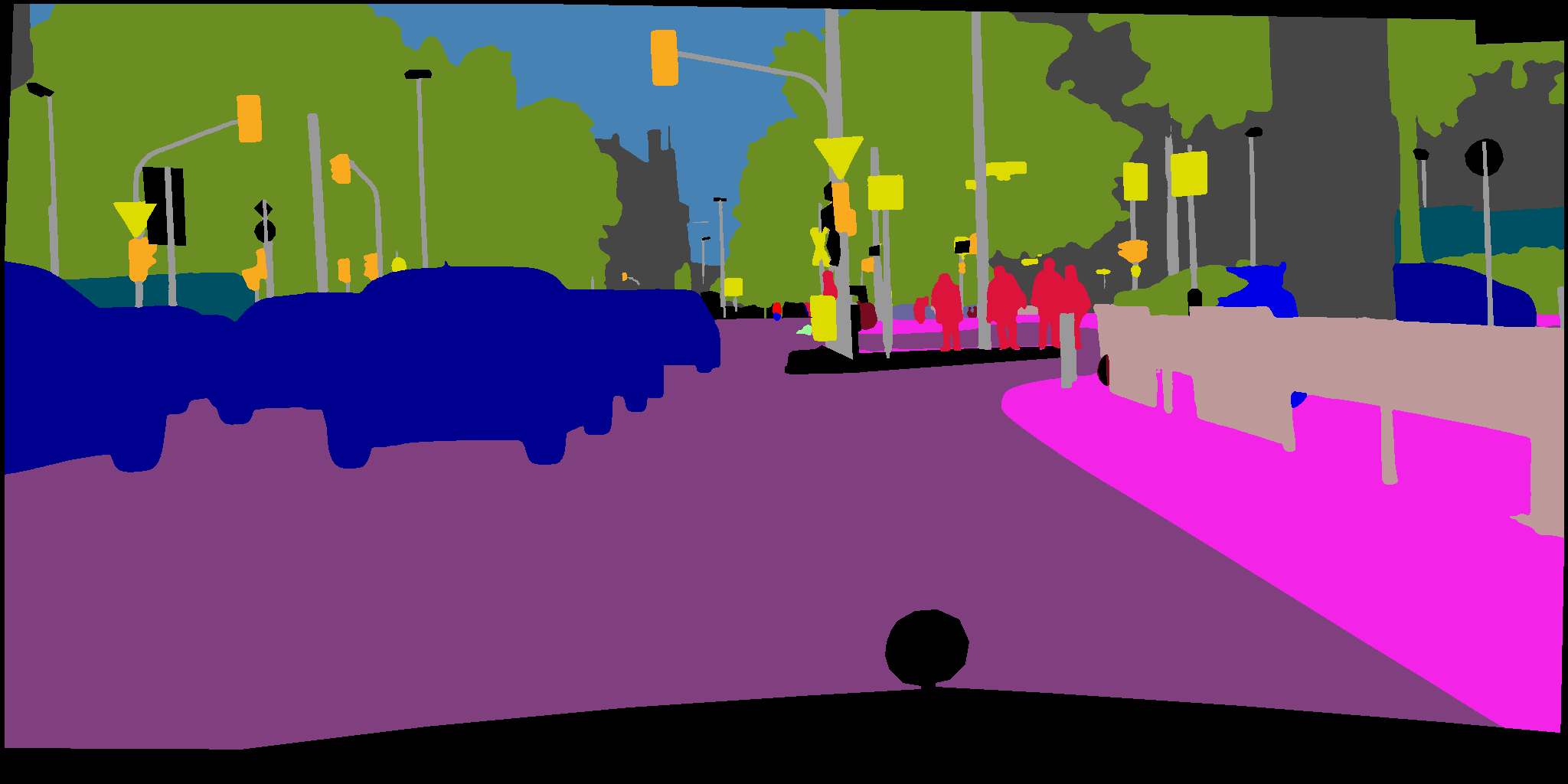}
  \end{subfigure}
  \begin{subfigure}[t]{0.16\linewidth}
    \includegraphics[width=\linewidth]{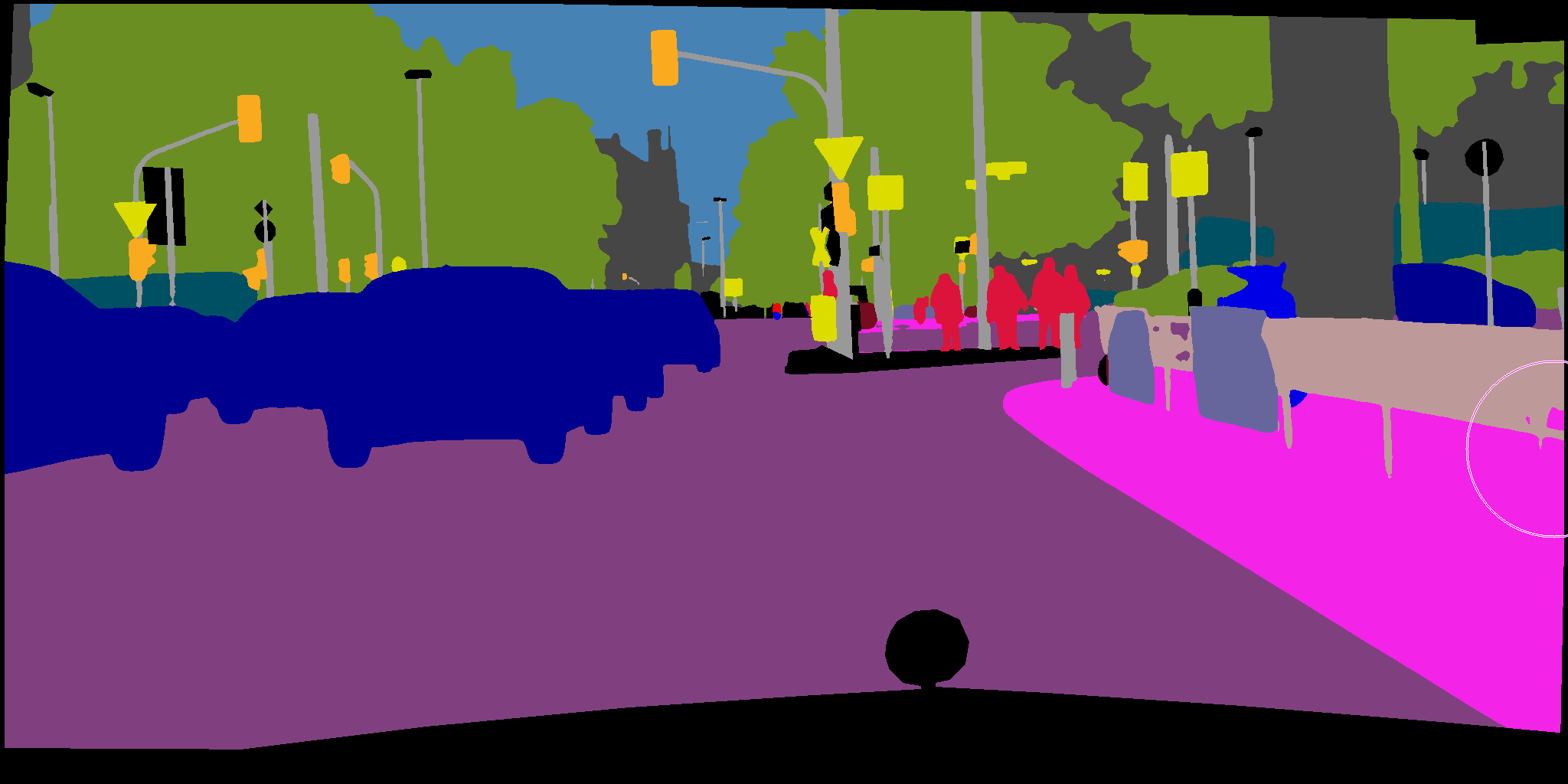}
  \end{subfigure}
  \begin{subfigure}[t]{0.16\linewidth}
    \includegraphics[width=\linewidth]{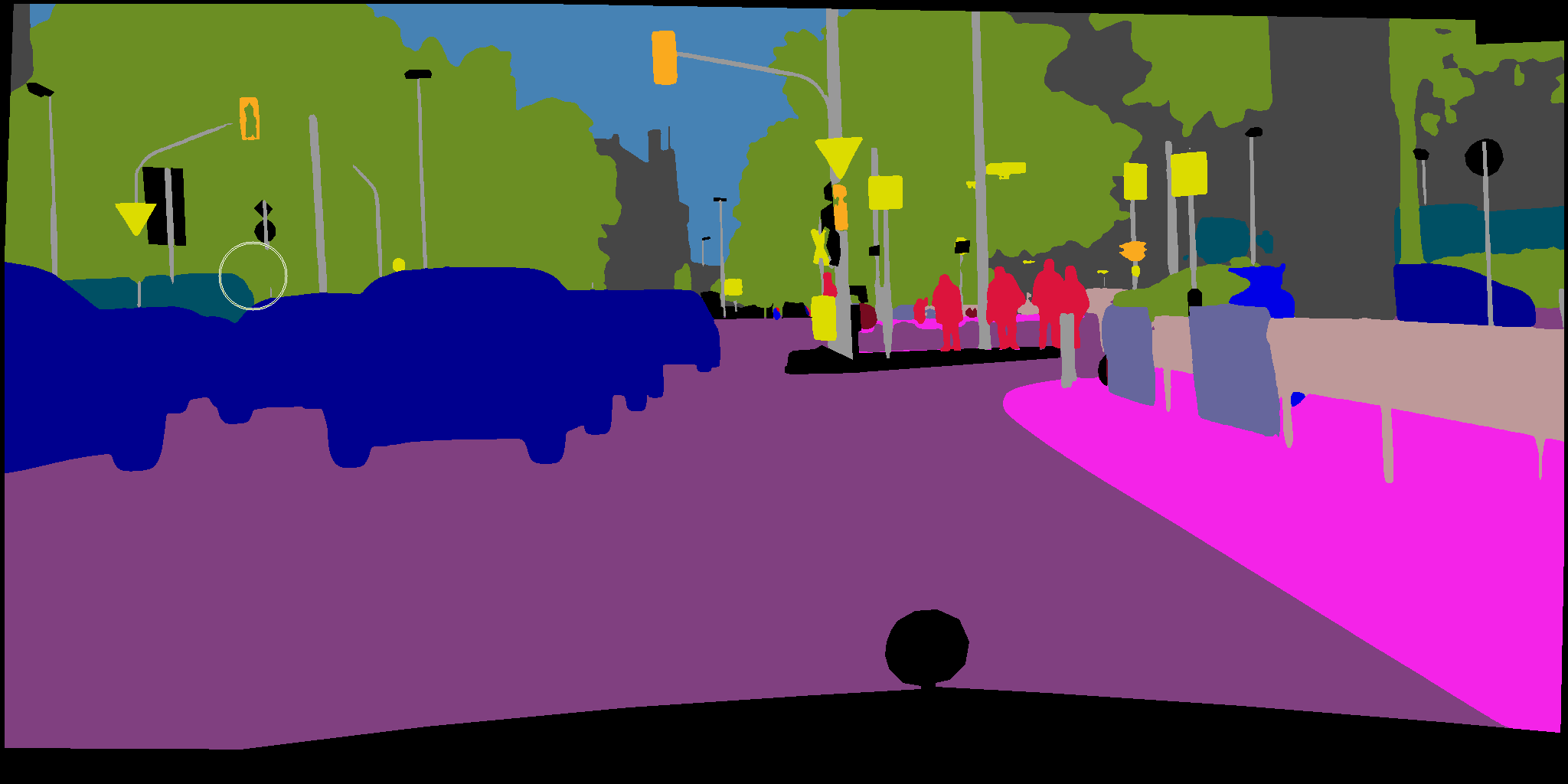}
  \end{subfigure}
  \begin{subfigure}[t]{0.16\linewidth}
    \includegraphics[width=\linewidth]{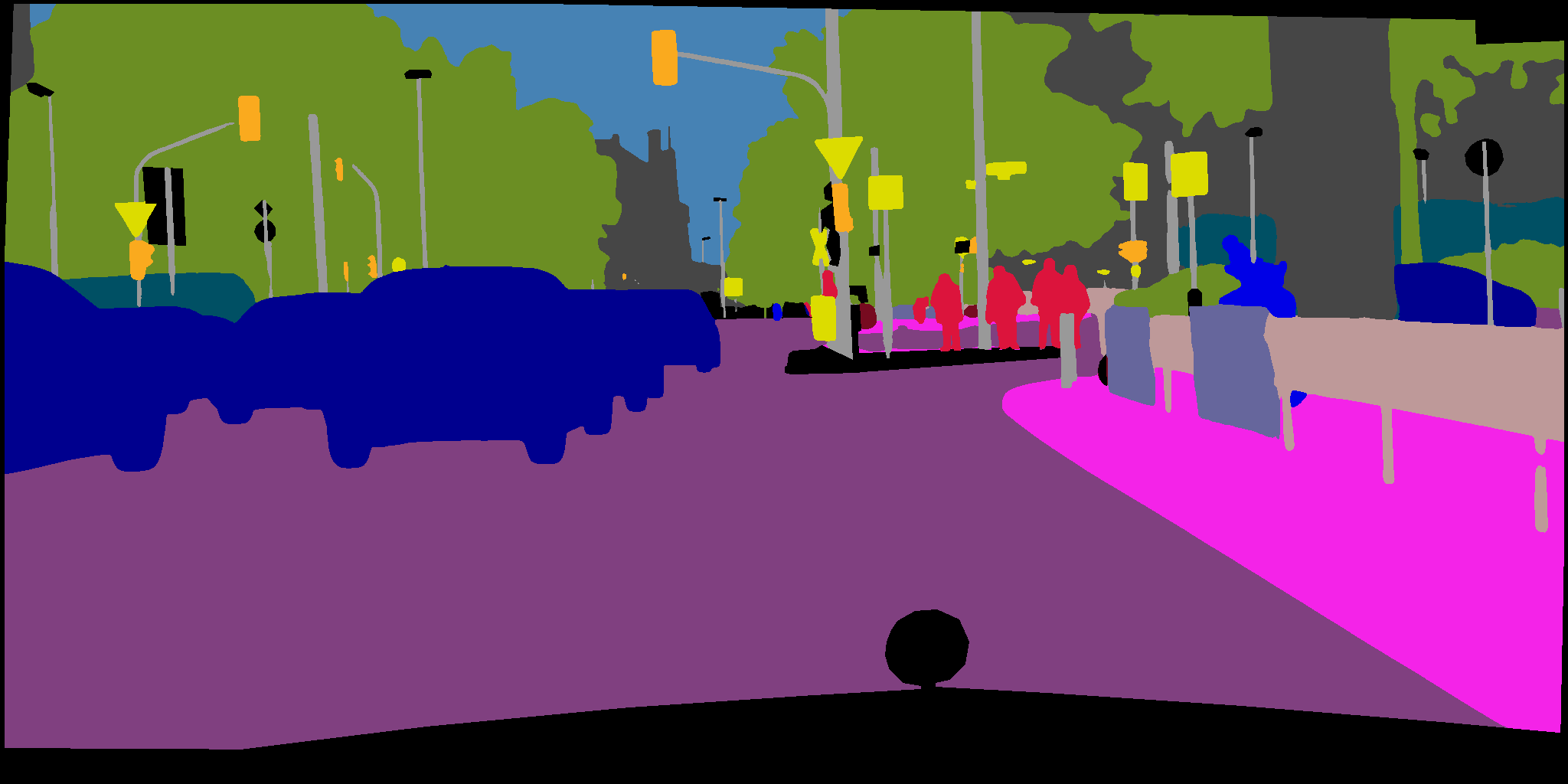}
  \end{subfigure}
  \vspace{1mm}
  \begin{subfigure}[t]{0.16\linewidth}
    \includegraphics[width=\linewidth]{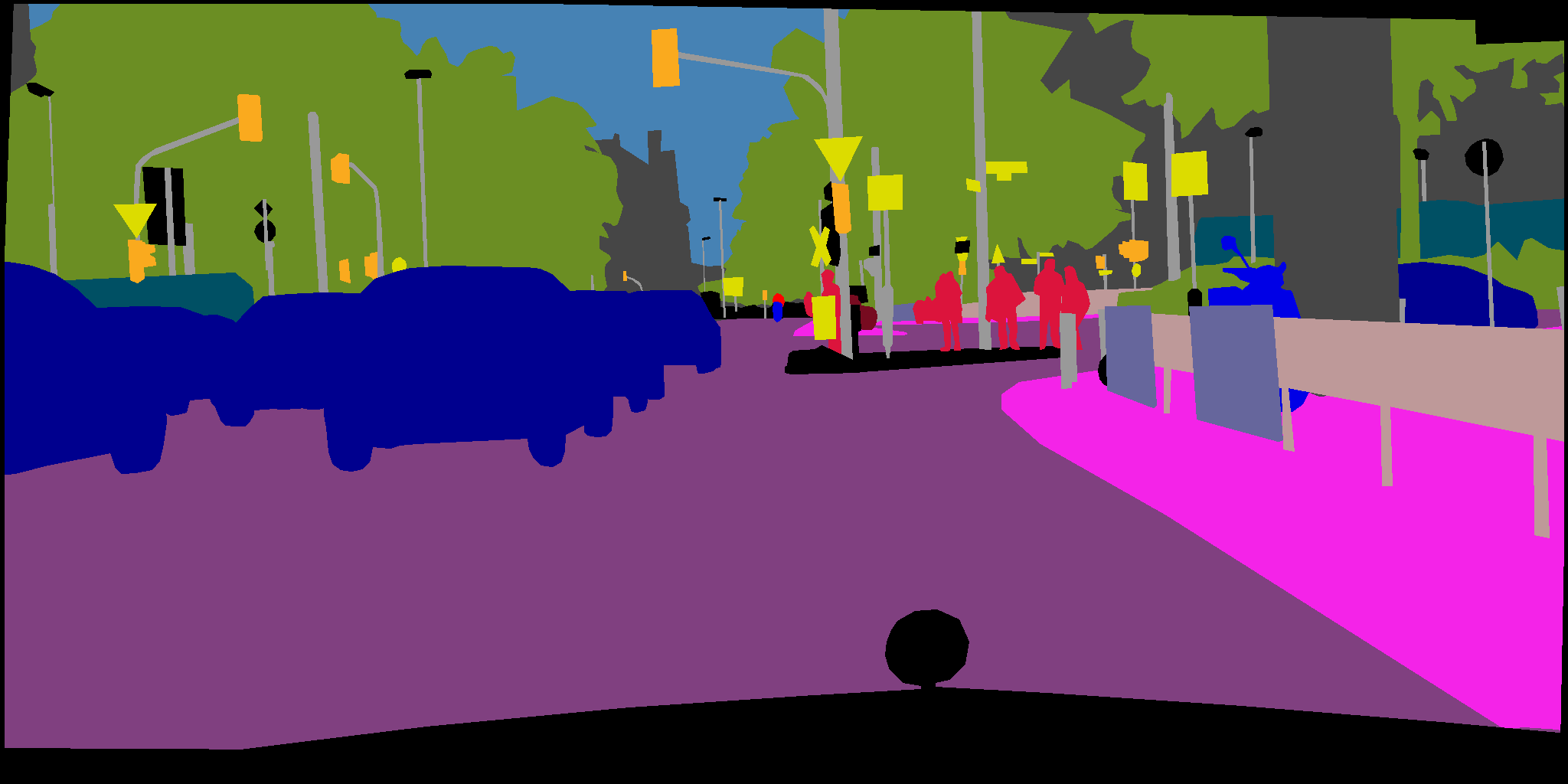}
  \end{subfigure}
   \begin{subfigure}[t]{0.16\linewidth}
    \includegraphics[width=\linewidth]{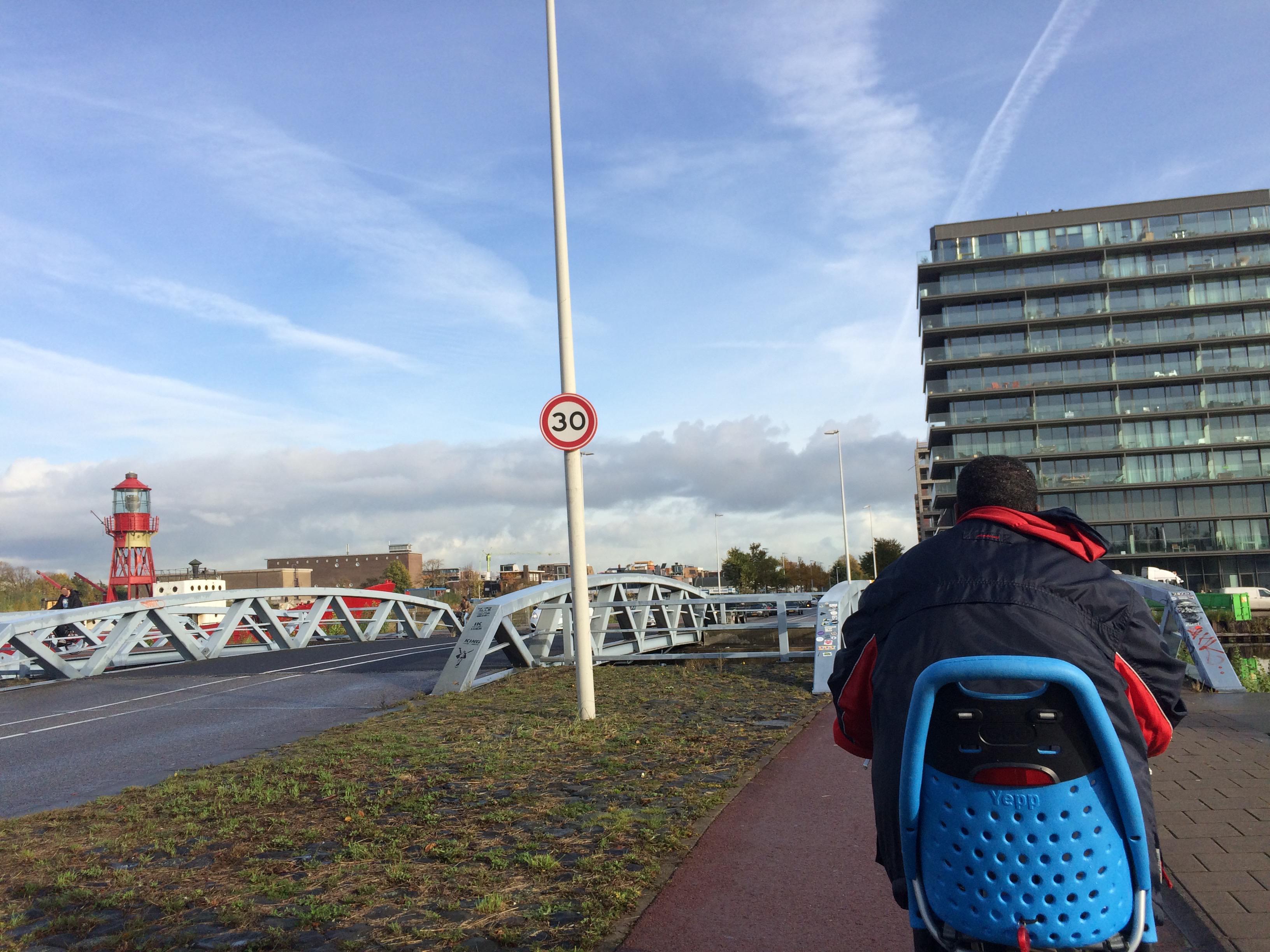}
  \end{subfigure}
  \begin{subfigure}[t]{0.16\linewidth}
    \includegraphics[width=\linewidth]{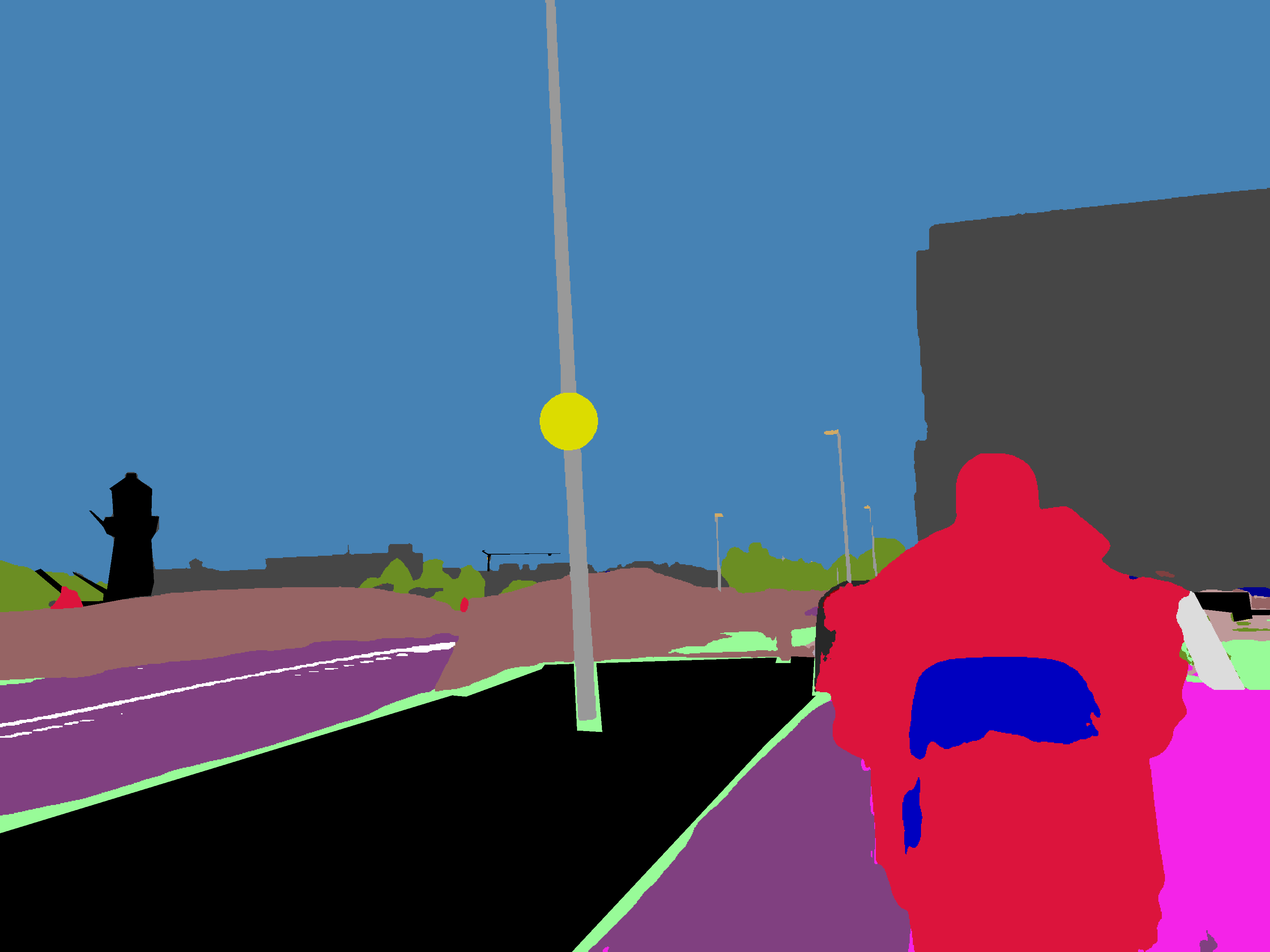}
  \end{subfigure}
  \begin{subfigure}[t]{0.16\linewidth}
    \includegraphics[width=\linewidth]{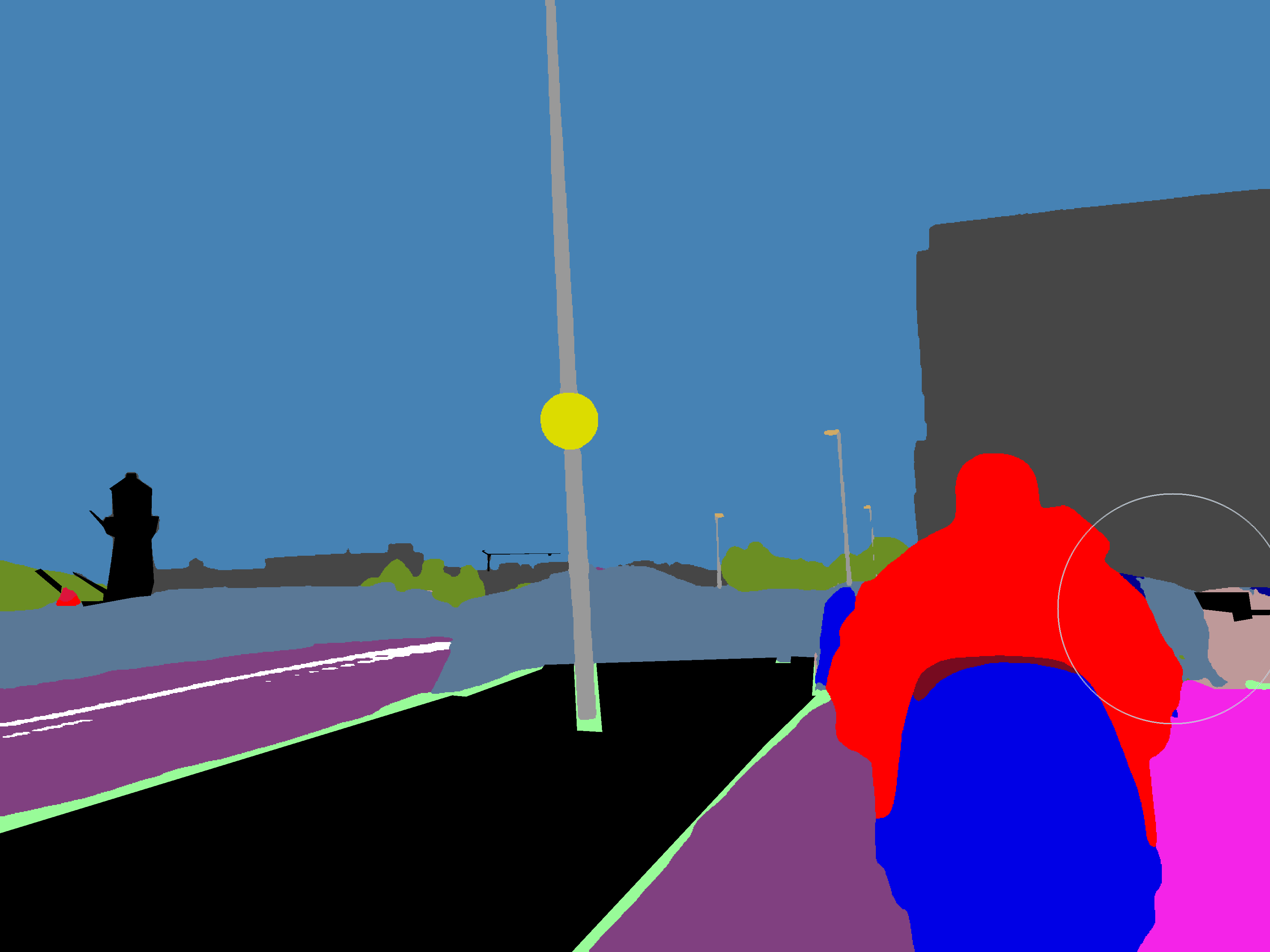}
  \end{subfigure}
  \begin{subfigure}[t]{0.16\linewidth}
    \includegraphics[width=\linewidth]{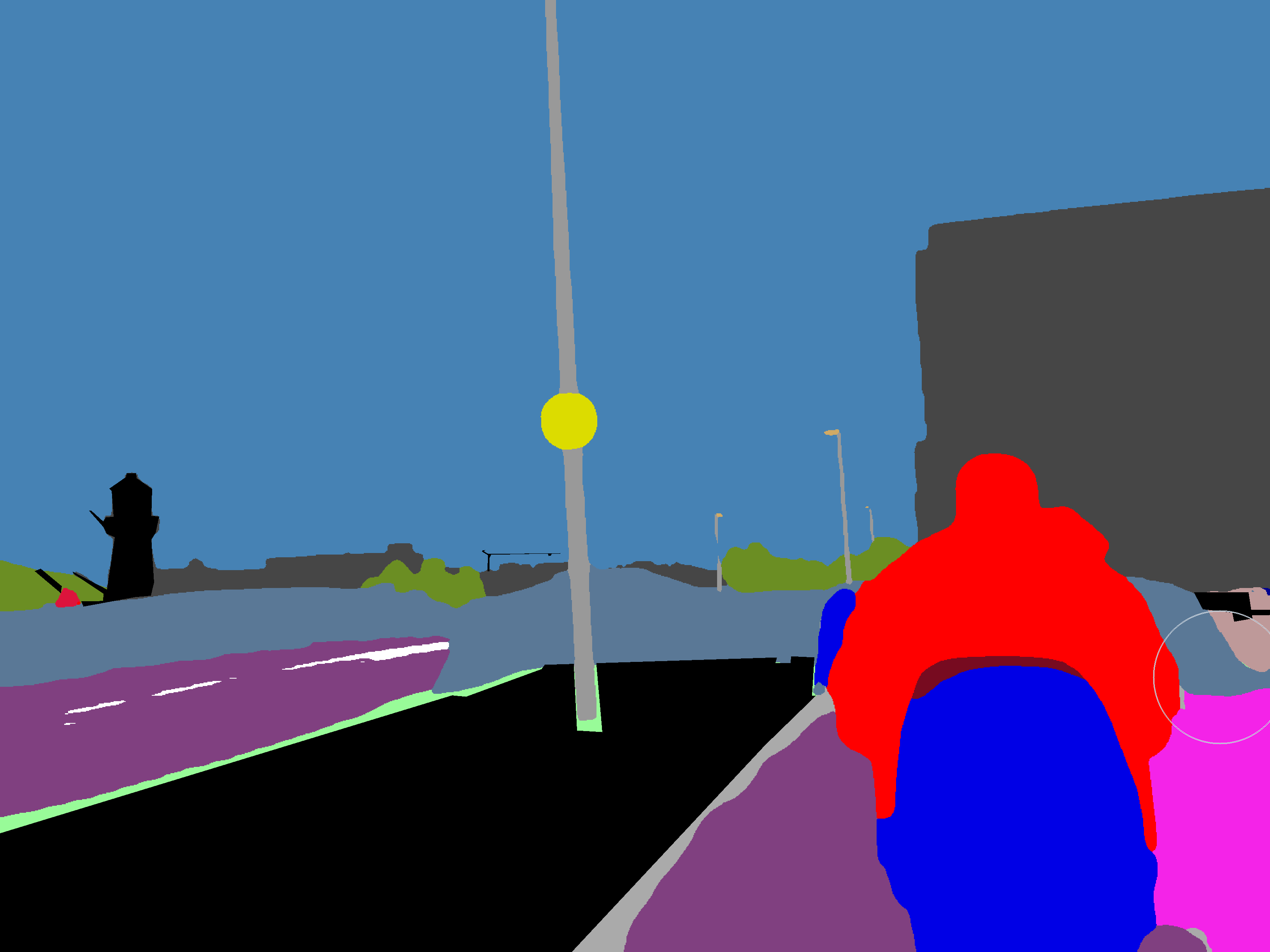}
  \end{subfigure}
  \begin{subfigure}[t]{0.16\linewidth}
    \includegraphics[width=\linewidth]{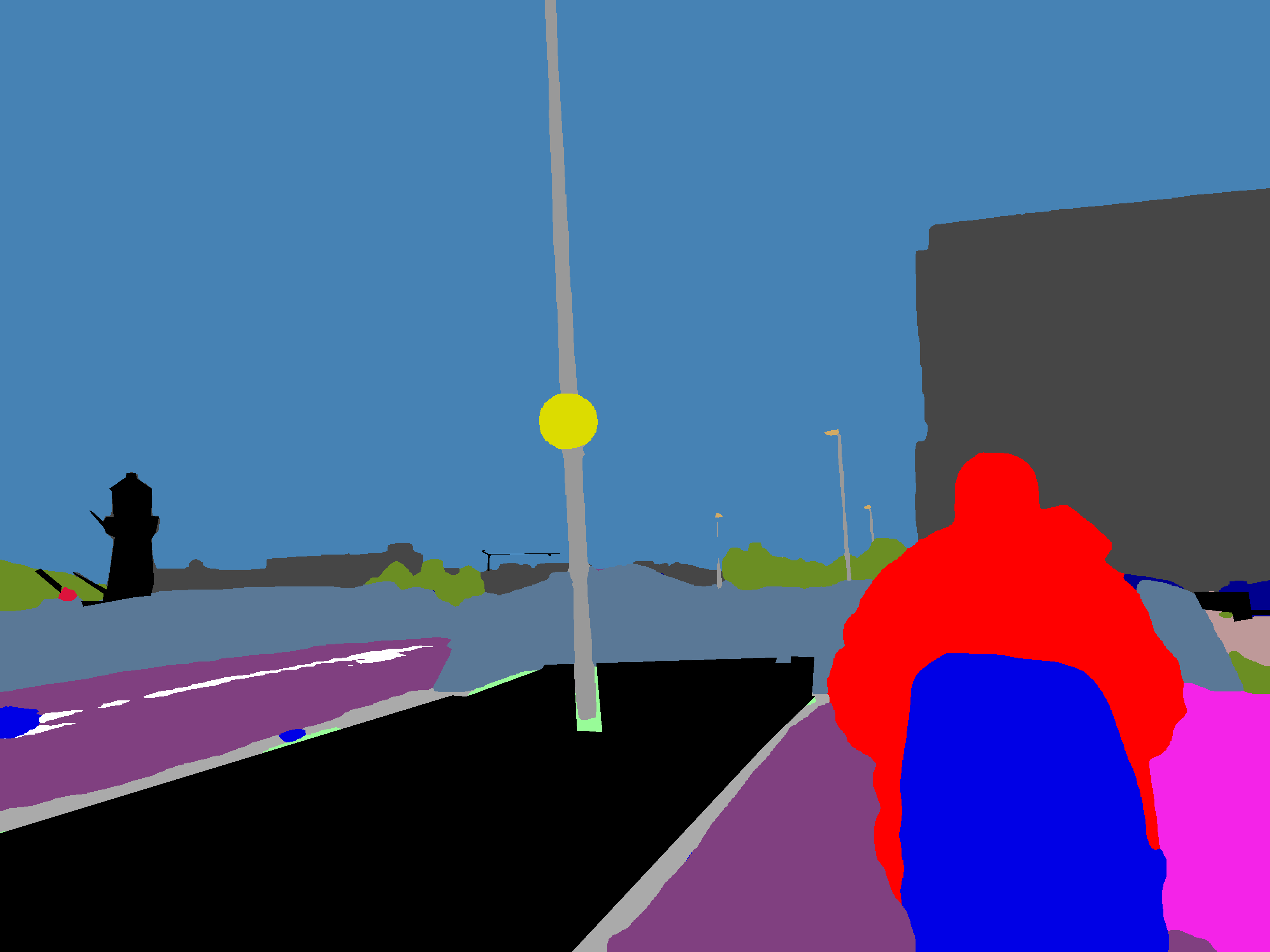}
  \end{subfigure}
  \vspace{1mm}
  \begin{subfigure}[t]{0.16\linewidth}
    \includegraphics[width=\linewidth]{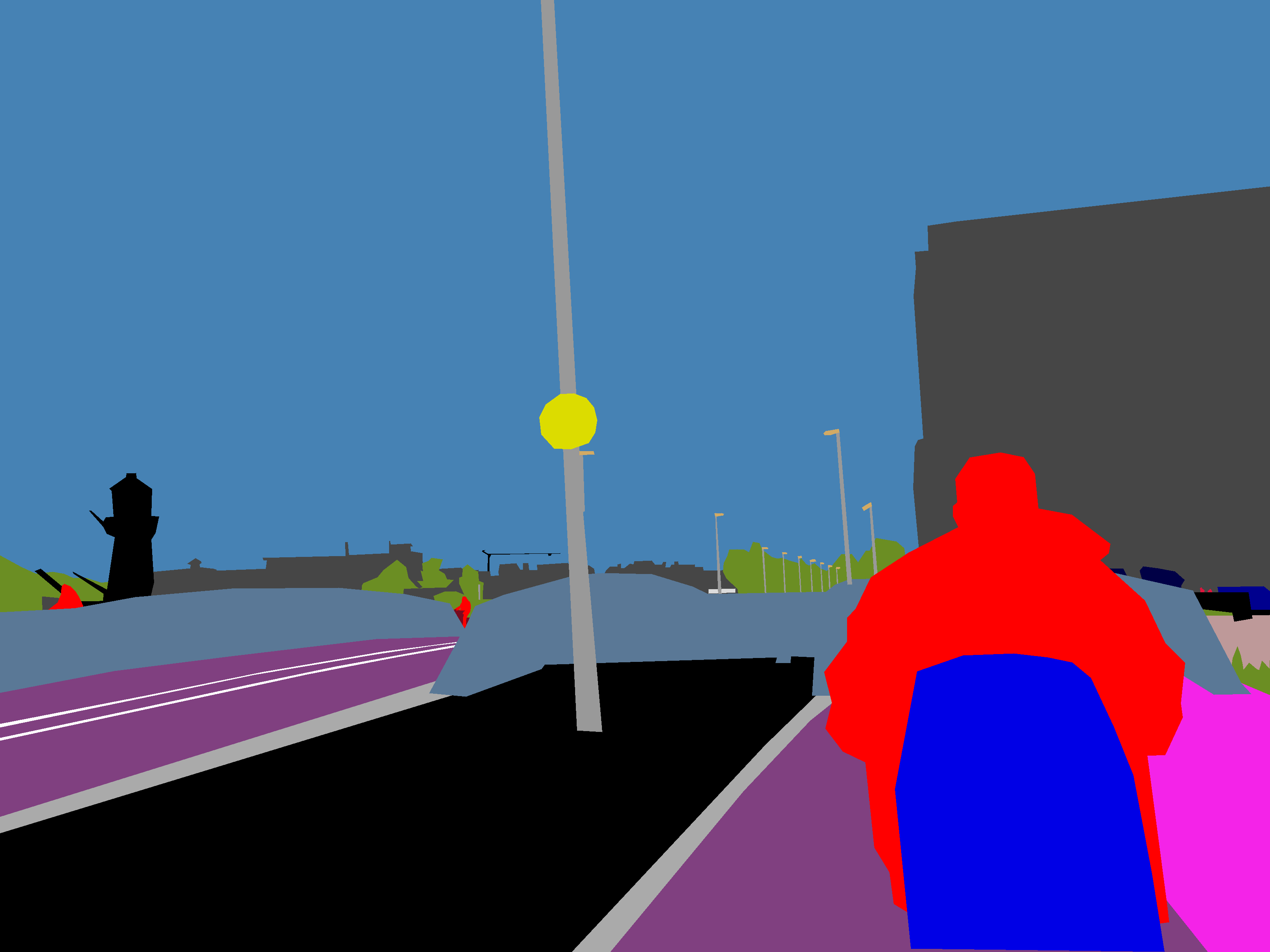}
  \end{subfigure}
   \begin{subfigure}[t]{0.16\linewidth}
    \includegraphics[width=\linewidth]{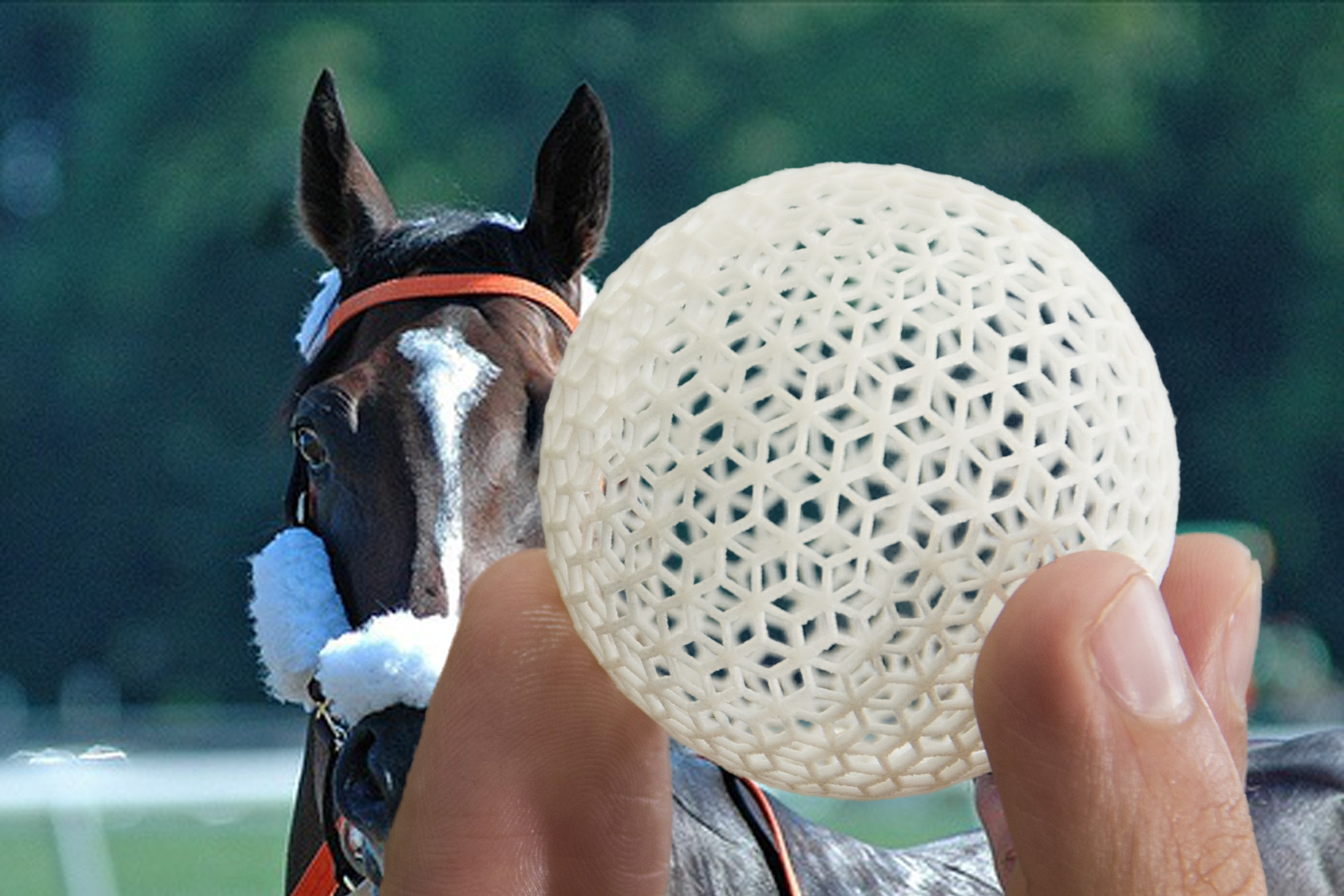}
  \end{subfigure}
  \begin{subfigure}[t]{0.16\linewidth}
    \includegraphics[width=\linewidth]{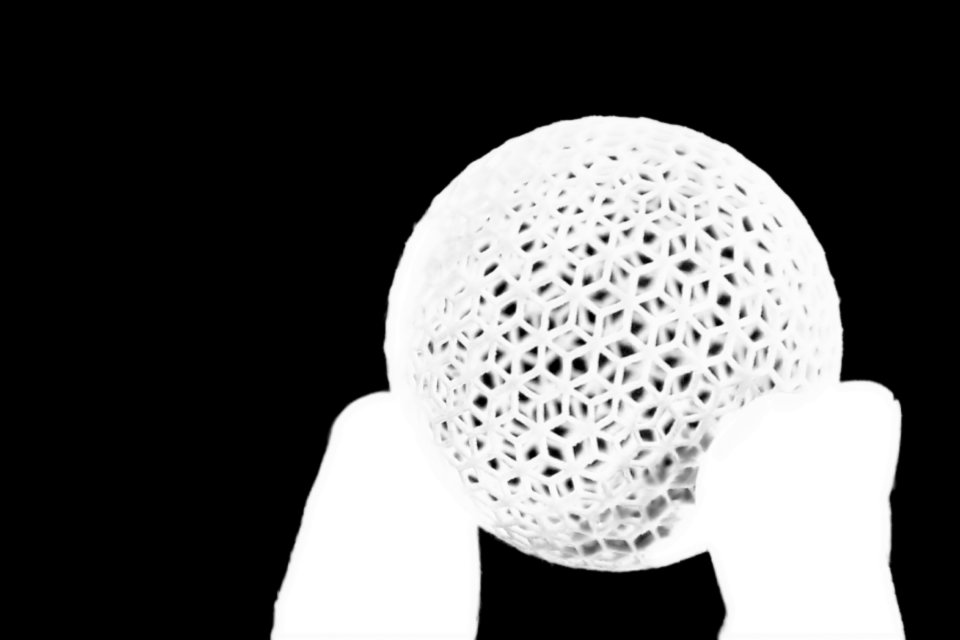}
  \end{subfigure}
  \begin{subfigure}[t]{0.16\linewidth}
    \includegraphics[width=\linewidth]{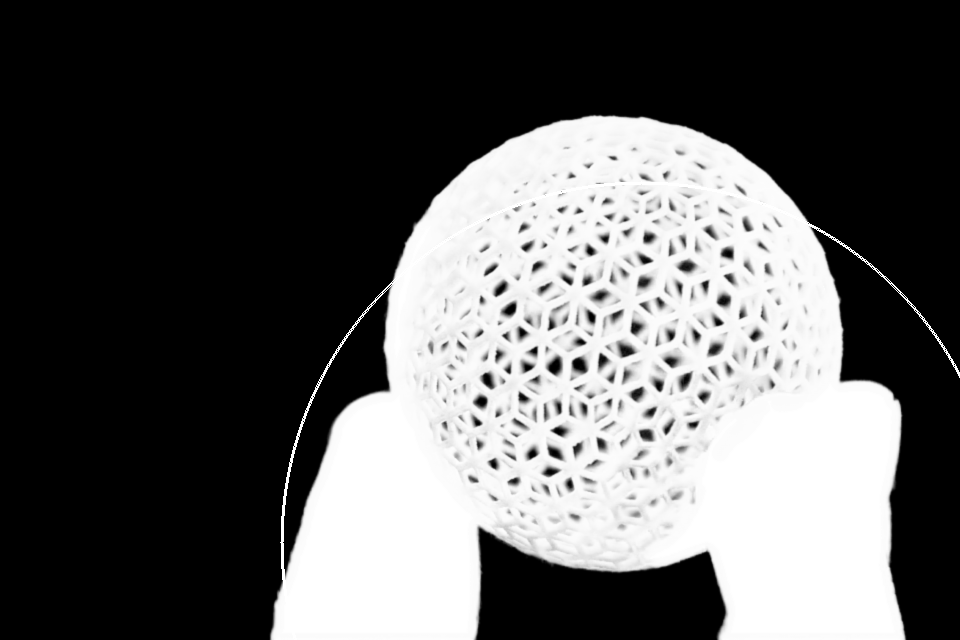}
  \end{subfigure}
  \begin{subfigure}[t]{0.16\linewidth}
    \includegraphics[width=\linewidth]{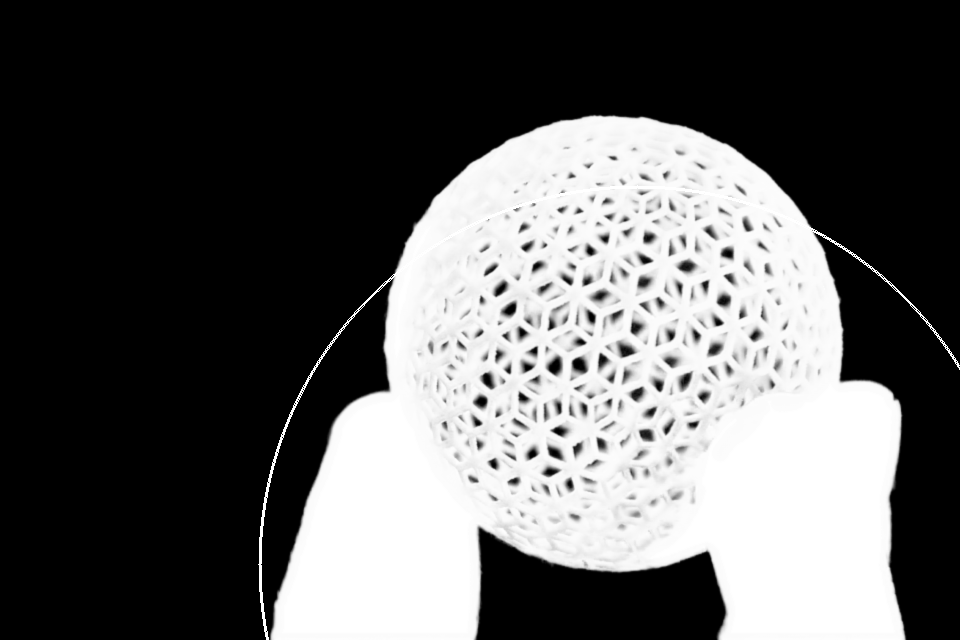}
  \end{subfigure}
  \begin{subfigure}[t]{0.16\linewidth}
    \includegraphics[width=\linewidth]{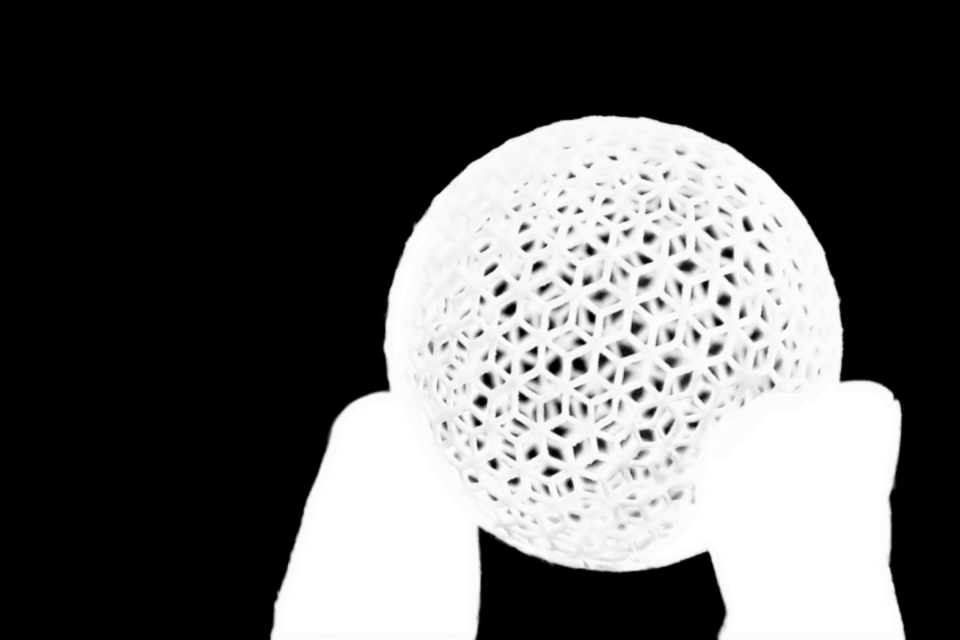}
  \end{subfigure}
  \vspace{1mm}
  \begin{subfigure}[t]{0.16\linewidth}
    \includegraphics[width=\linewidth]{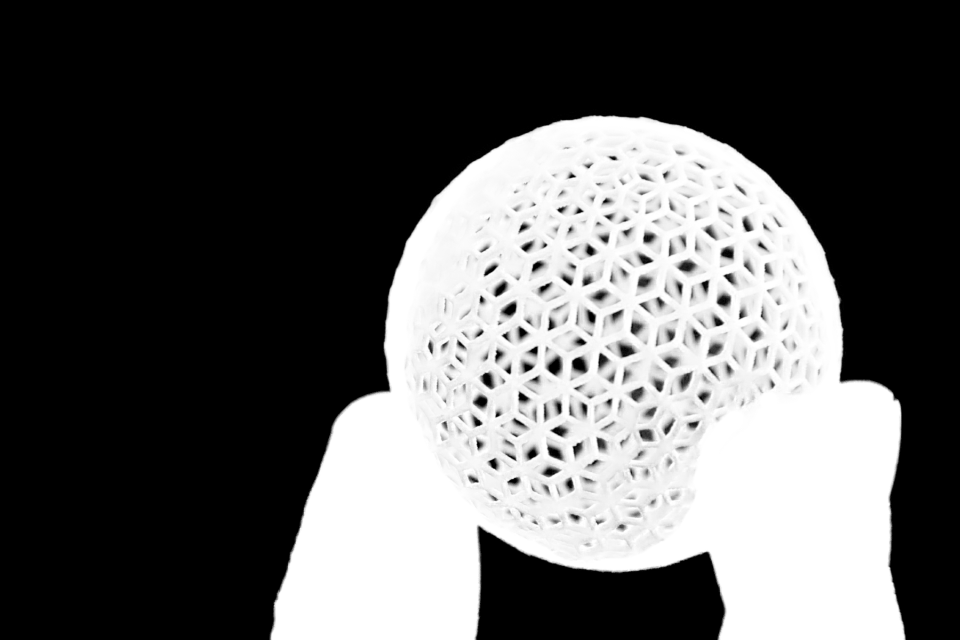}
  \end{subfigure}
   \begin{subfigure}[t]{0.16\linewidth}
    \includegraphics[width=\linewidth]{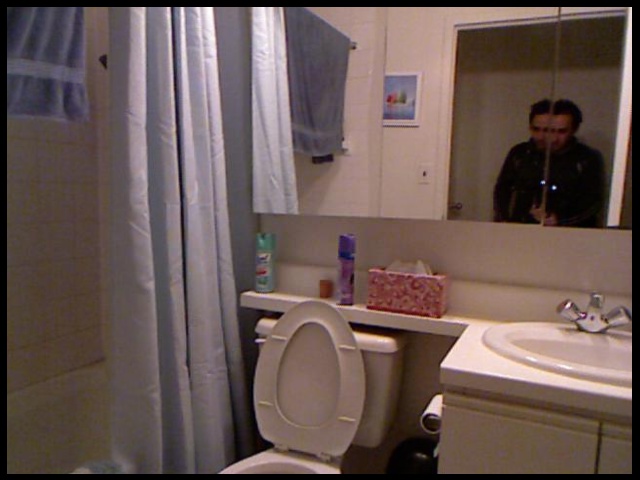}
    \caption*{Input image}
  \end{subfigure}
  \begin{subfigure}[t]{0.16\linewidth}
    \includegraphics[width=\linewidth]{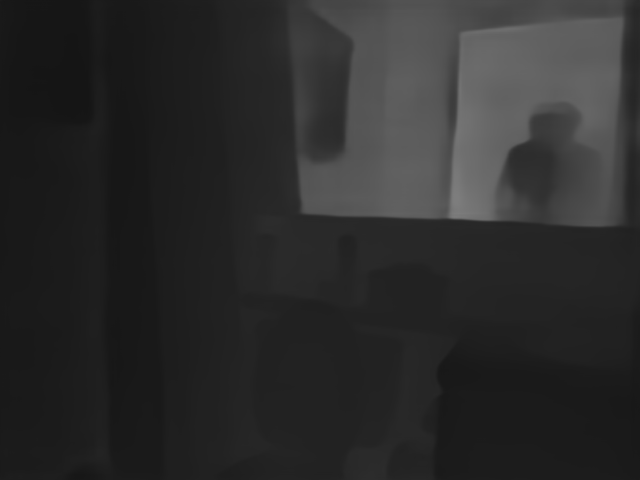}
    \caption*{Initial estimation}
  \end{subfigure}
  \begin{subfigure}[t]{0.16\linewidth}
    \includegraphics[width=\linewidth]{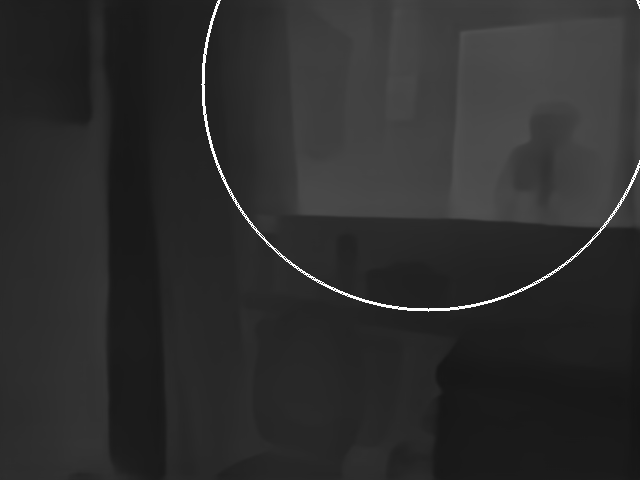}
    \caption*{$\text{Click}_{5}$}
  \end{subfigure}
  \begin{subfigure}[t]{0.16\linewidth}
    \includegraphics[width=\linewidth]{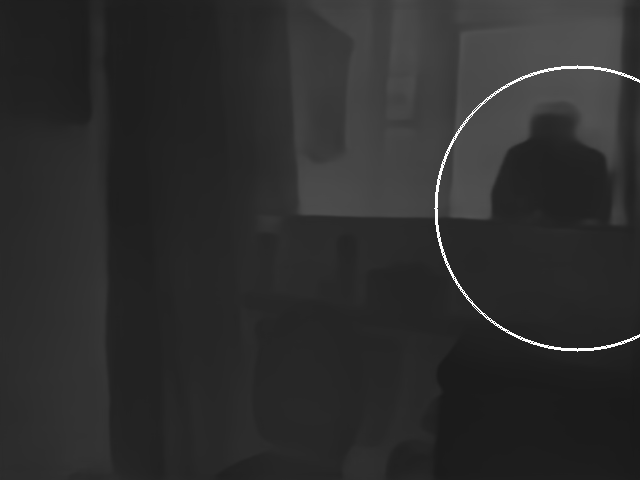}
    \caption*{$\text{Click}_{10}$}
  \end{subfigure}
  \begin{subfigure}[t]{0.16\linewidth}
    \includegraphics[width=\linewidth]{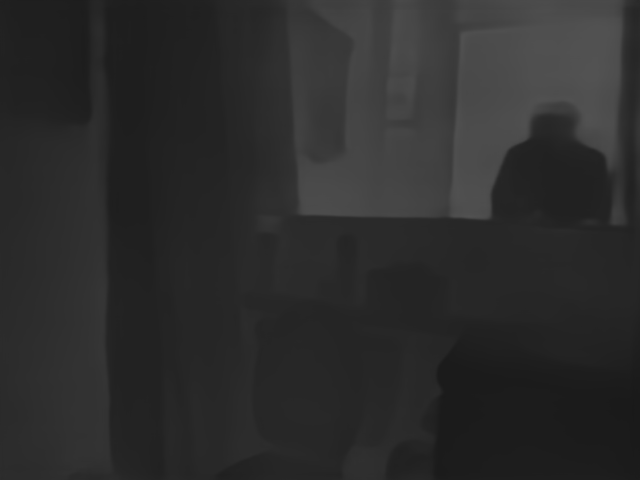}
    \caption*{Final output}
  \end{subfigure}
  \vspace{1mm}
  \begin{subfigure}[t]{0.16\linewidth}
    \includegraphics[width=\linewidth]{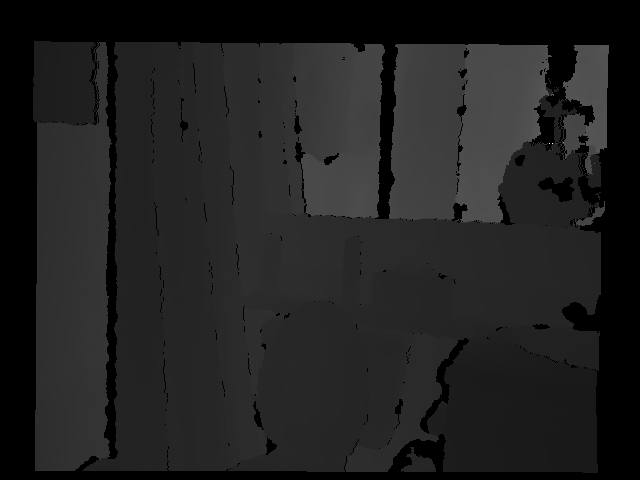}
    \caption*{Ground truth}
  \end{subfigure}
  \caption{Qualitative examples on SBD, Cityscapes, Mapillary Vista, Composition-1k and NYU-Depth-V2 using G-BRS-bmconv. Clicks with attention radius are visualized. Black region for semantic segmentation and depth estimation is invalid. \textbf{Best viewed in magnification.}}
  \label{fig:qualitative_results}
  \vspace{-4mm}
\end{figure*}

We perform experiments on five benchmark datasets and evaluate on the test/validation sets with publicly available ground truth that enables automatic click generation. We compare the quantitative results of the four types of G-BRS layers. For architectures with multiple G-BRS insertions, we incrementally include insertions for features with higher resolution. In addition to results on the complete test/validation sets, we report results for the 10\% of the instances with the lowest initial scores for two reasons: first, since the selected state-of-the-art models can already achieve high average initial accuracy, separate evaluation can better demonstrate the effectiveness of G-BRS on instances with more prominent error. Second, for real-world applications, instances that are high-priority targets for refinement are instances with the worst initial estimation. \\
\indent For additional analysis, we perform ablation study on the effectiveness of the proposed consistency loss. Since ~\cite{Sofiiuk20} suggested that backpropagating refinement can also be applied using the RGB input as parameters instead of features, we include results using RGB-BRS. Qualitative examples of interactive refinement for all applications are shown in Figure \ref{fig:qualitative_results}. Additional qualitative comparisons between different settings are included in the supplementary document.\\
\begin{figure*}[t]
 \centering
  \begin{subfigure}[t]{0.32\linewidth}
    \includegraphics[width=\linewidth]{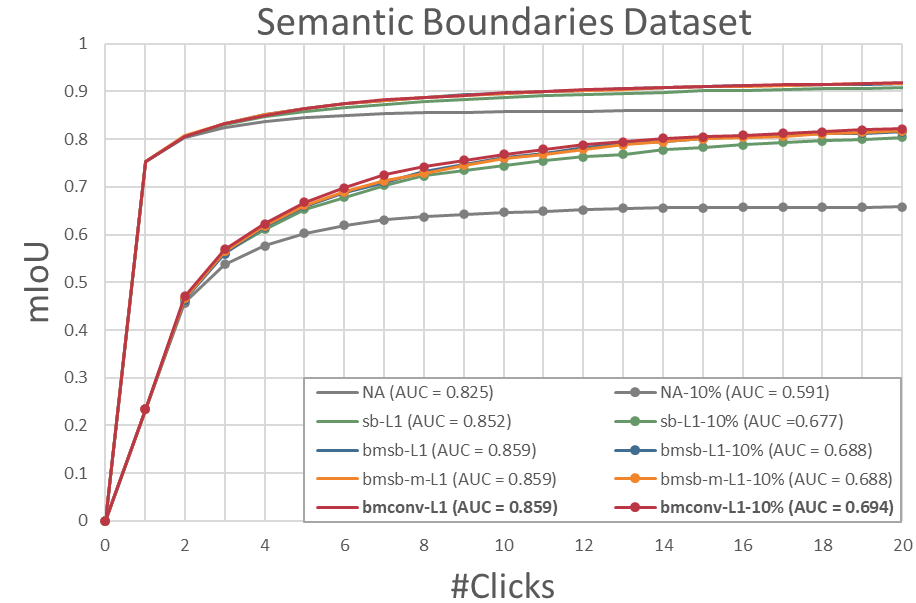}
    \caption{Interactive Segmentation}\label{fig:is_sbd}
  \end{subfigure}
  \begin{subfigure}[t]{0.32\linewidth}
    \includegraphics[width=\linewidth]{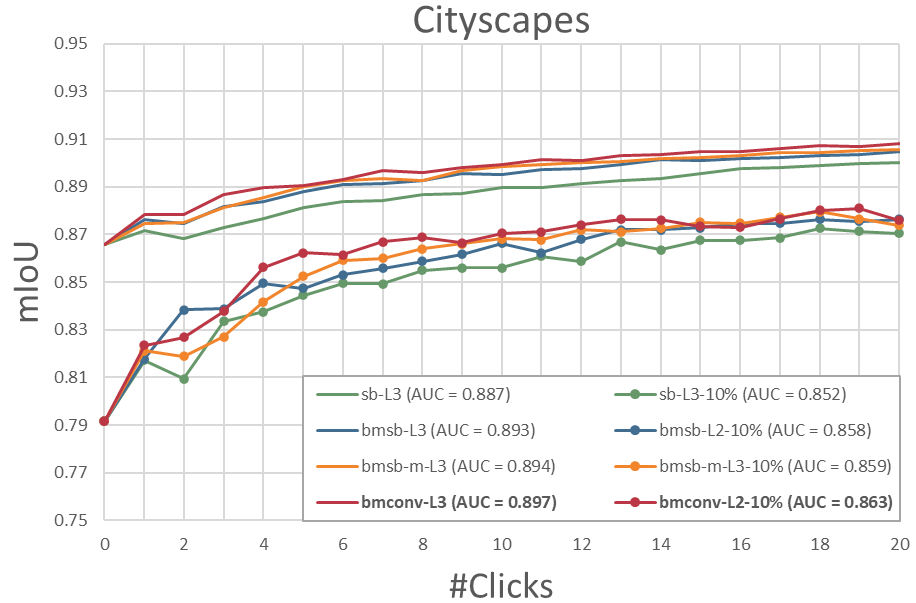}
    \caption{Semantic Segmentation - Cityscapes}\label{fig:ss_cityscapes}
  \end{subfigure}
  \begin{subfigure}[t]{0.32\linewidth}
    \includegraphics[width=\linewidth]{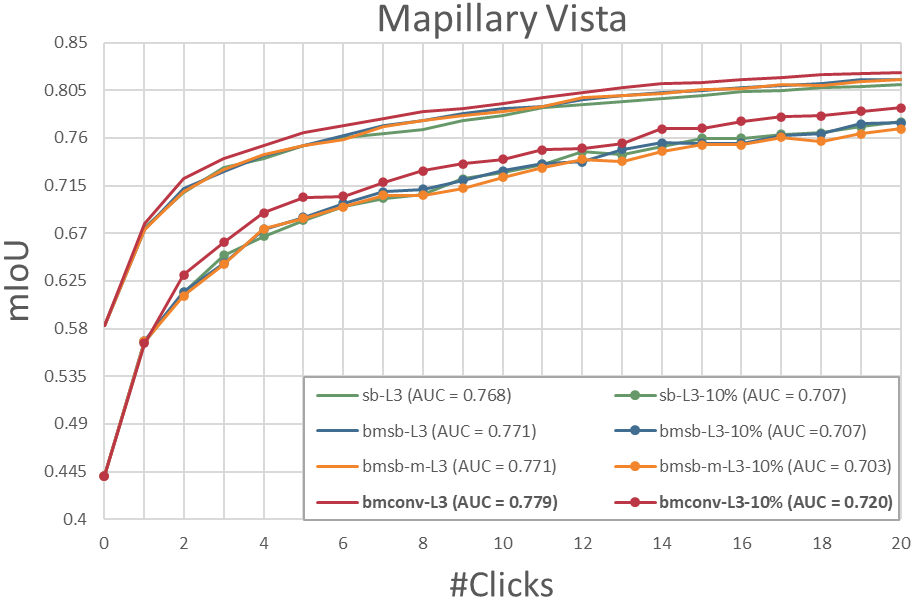}
    \caption{Semantic Segmentation - Mapillary Vista}\label{fig:ss_mapillary}
  \end{subfigure}
  \caption{Quantitative results on interactive segmentation and semantic segmentation using various G-BRS settings with consistency loss. The number of layers that achieve the best scores are reported for each type of layer (e.g. L3 indicates 3 active layers.)}
  \vspace{-0.4cm}
  \label{fig:is_ss}
\end{figure*}
\subsection{Evaluation Protocol}
We compute the standard metrics for all four tasks on each provided click. For a thorough analysis, we additionally compute the following metrics: (1) Area Under Curve (AUC) of the selected metrics to account for convergence time, (2) best score achieved in total number of clicks. We first report results obtained using the consistency loss and provide ablation study in a later section. To find the optimal learning rate for each G-BRS setting, we select a subset of each test set to evaluate using 10 learning rates ranging from $0.1$ to $0.1\times{0.5}^{9}$. We report top scores achieved for each type of G-BRS layer and include all experimental results, learning rates used and run time analysis in the supplementary document due to the space limit.\\
\indent To enable quantitative evaluation of our refinement procedures, we use two different automatic click generation strategies. For interactive segmentation and semantic segmentation that requires pixel-wise classification, let us define the binary error mask ${\xi}_{c} \in \{0, 1\}^{H \times W}$ that represents the misclassified region for class $c$. We generate the next click with the target label at the location defined below,
\begin{equation}\label{eq:auto-click-1}
\begin{gathered}
{c}^{*} = \underset{c}{\arg\max}\ (max(\mathcal{D}({\xi}_{c})))\\
(u, v) = \underset{u, v}{\arg\max}\ \mathcal{D}({\xi}_{{c}^{*}})
\end{gathered}
\end{equation}
where $\mathcal{D}$ denotes a distance transform function and ${c}^{*}$ is the selected class. Note the region with the ignored label is excluded from the error mask computation. To enable automatic radius generation, we select the connected component ${\xi}_{e}$ from ${\xi}_{c}$ that contains $(u, v)$ and compute the maximum Euclidian distance between $(u, v)$ and the boundary of ${\xi}_{e}$.\\
\indent For image matting and depth estimation that requires pixel-wise regression, we use a similar click generation strategy that first transforms the regression error mask to segmentation error mask $\xi$ using Otsu thresholding. Second, as ${\xi}_{c}$ can be computed for each class in segmentation tasks, we divide the error mask $\xi$ with positive and negative error into ${\xi}_{+}$ and ${\xi}_{-}$. The clicking location $(u, v)$ can then be generated by following the same strategy as Equation \ref{eq:auto-click-1}. For radius generation, we observe that an insufficient radius is counterproductive as it prevents accurate refinement outside of the small attention region and drastically impacts performance. To this end, we apply dilation with a kernel size of $15$ to the selected ${\xi}_{+/-}$ and compute the radius following the aforementioned strategy for segmentation.
\subsection{Evaluation - Interactive Segmentation}\label{sec:eval_is}
We evaluate on the Semantic Boundaries Dataset (SBD), which is currently the largest dataset for interactive segmentation with 2,820 test images and 6,671 instance-level object masks. Since the input clicks that generate the interaction maps also achieve improvement without backpropagating refinement, we run experiments without G-BRS as a baseline comparison. Figure \ref{fig:is_sbd} shows the mean Intersection over Union (mIoU) computed over all object instances for 20 clicks on various settings. We compute AUC using mIoU for segmentation tasks. It is shown that the baseline approach (denoted as NA) has limited refinement capability comparing to methods that utilize backpropagating refinement. Note that the G-BRS-sb layer in this task is equivalent to auxiliary variables used in \text{\textit{f}-BRS}~\cite{Sofiiuk20}. Since \text{\textit{f}-BRS} is not implemented on the other applications we tackle, we refer to this layer archicture as G-BRS-sb in our experiments. Results show that all three G-BRS layers proposed in this work outperform the G-BRS-sb layer (\text{\textit{f}-BRS}), with G-BRS-bmconv layer achieving the top $\text{AUC}_\text{mIoU}$ of 0.859 and 0.694 for the test set and the bottom 10\% instances. The G-BRS-bmconv layer also achieves the best peak mIoU obtained in the total number of clicks with a score of 0.918.
\begin{table}[t]
  \small
  \centering
  \begin{tabular}{c|c|c}
  \toprule
Methods& 
\begin{tabular}{@{}c@{}}$\mathcal{L}_{brs}$~\cite{Jang19}\\\hline
	\begin{tabular}{P{0.65cm}|P{1.0cm}}
		AUC&$\text{mIoU}_{max}$
	\end{tabular}
\end{tabular}& 
\begin{tabular}{@{}c@{}}$\mathcal{L}_{c}$ (Ours)\\\hline
	\begin{tabular}{P{0.65cm}|P{1.0cm}}
		AUC&$\text{mIoU}_{max}$
	\end{tabular}
\end{tabular}\\\hline
\midrule
DistMap-BRS & \begin{tabular}{P{0.65cm}|P{1.0cm}}0.832&0.894\end{tabular} & \begin{tabular}{P{0.65cm}|P{1.0cm}}0.845&0.891\end{tabular} \\\hline
RGB-BRS & \begin{tabular}{P{0.65cm}|P{1.0cm}}\textbf{0.853}&\textbf{0.908}\end{tabular} & \begin{tabular}{P{0.65cm}|P{1.0cm}}\textbf{0.851}&\textbf{0.905}\end{tabular} \\\hline
  \end{tabular}
\caption{Comparison between refinement settings using the input.}
\label{tab:table_is_rgb_brs}
\vspace{-7mm}
\end{table}
\begin{figure*}[t]
 \centering
 \begin{subfigure}[t]{0.45\linewidth}
    \includegraphics[width=\linewidth]{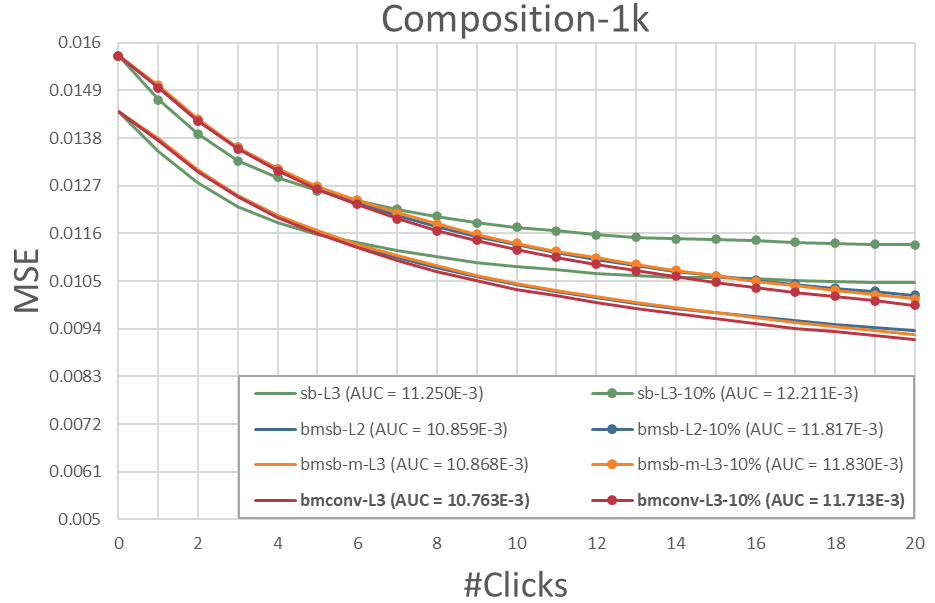}
    \caption{Image Matting}\label{fig:mt_composition_1k}
  \end{subfigure}
  \begin{subfigure}[t]{0.45\linewidth}
    \includegraphics[width=\linewidth]{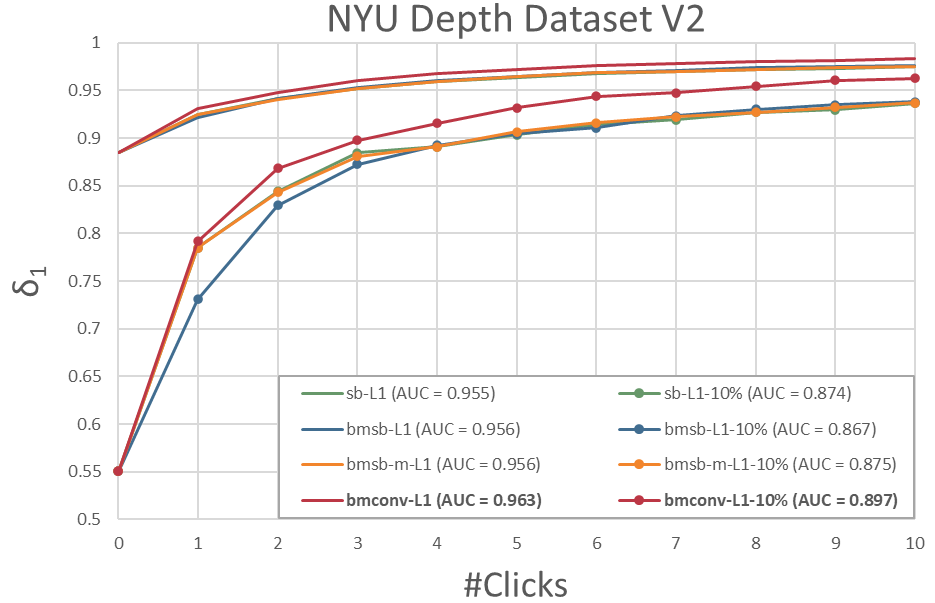}
    \caption{Depth Estimation}\label{fig:de_nyu_depth_v2}
  \end{subfigure}
  \vspace{-0.2cm}
  \caption{Quantitative results on image matting and depth estimation with consistency loss. }
  \vspace{-0.4cm}
  \label{fig:mt_de}
\end{figure*}\\
\indent To compare with backpropagating refinement settings that use the input as parameters, we first perform DistMap-BRS~\cite{Jang19} that uses the input distance maps as parameters. RGB-BRS that uses the RGB input is also performed, which should be an equivalent solution as suggested by~\cite{Sofiiuk20}. As the proposed original DistMap-BRS by~\cite{Jang19} uses a corrective energy and inertial energy with the L-BFGS optimizer, we refer to this loss minimization method as $\mathcal{L}_{brs}$ and compare it with our method that uses the consistency loss $\mathcal{L}_{c}$ with the Adam optimizer. Table \ref{tab:table_is_rgb_brs} shows that RGB-BRS outperforms DistMap-BRS and has a slightly higher AUC of $0.853$ when using $\mathcal{L}_{brs}$. However, the overall advantage of $\mathcal{L}_{brs}$ greatly diminishes as the L-BFGS optimizer is extremely memory intensive and therefore inapplicable for many applications. As a result, in a later section, we show results obtained using RGB-BRS with $\mathcal{L}_{c}$ for all applications to compare between backpropagating refinement using the input and the features. For interactive segmentation, Figure \ref{fig:is_sbd} shows that all four types of G-BRS layers outperform RGB-BRS.
\subsection{Evaluation - Semantic Segmentation}
Since the ground truth of the test sets is not publicly available for automatic click generation, we select the validation sets of Cityscapes and Mapillary Vista for the evaluation of this task. Cityscapes provides 500 test images with 19 classes while Mapillary Vista presents a more challenging task with 2,000 instances and 65 classes. We resize the input resolution for Mapillary Vista to match the area of $1920\times960$ due to GPU memory constraints. Figure \ref{fig:ss_cityscapes} shows that our G-BRS layers outperform the G-BRS-sb layer with the G-BRS-bmconv layer achieving the top $\text{AUC}_\text{mIoU}$ of $0.897$ and $0.863$ on Cityscapes and its bottom 10\% instances. Figure \ref{fig:ss_mapillary} shows that the G-BRS-bmconv layer also achieves the top $\text{AUC}_\text{mIoU}$ of $0.779$ and $0.720$ on Mapillary Vista and its bottom 10\% instances.\\
\indent We emphasize that our approach is capable of transforming existing state-of-the-art models into interactive methods that further achieve significant improvement. On Cityscapes, as the initial estimation from multi-scale HRNet-OCR~\cite{Yuan20} already achieves a high mIoU of $0.866$, the proposed G-BRS-bmconv layer is able to improve the mIoU to $0.9$ in only 10 clicks. For the Mapillary Vista dataset, despite a much lower mIoU of $0.582$ from the initial estimation, the G-BRS-bmconv layer achieves a mIoU of $0.822$ in 20 clicks, improving the initial score by 41.2\%. Additionally, for the bottom 10\% instances with a greater need for refinement, we achieve a 77.1\% improvement from a mIoU of $0.445$ to $0.788$ in 20 clicks.
\subsection{Evaluation - Image Matting}
We evaluate on the Composition-1k, which consists of 1,000 test images composited using 50 unique foreground objects. Standard metrics of Sum of Absolute Differences (SAD), Mean Squared Error (MSE), Gradient (Grad) and Connectivity (Conn) error are included in the supplementary document. For simplicity, we show the MSE for 20 clicks on various settings. Figure \ref{fig:mt_composition_1k} shows that the G-BRS-bmconv layer achieves the lowest $\text{AUC}_\text{mse}$ of $10.763\times10^{-3}$ and $11.713\times10^{-3}$ on Composition-1k and its bottom 10\% instances. It also decreases the MSE by 36.6\% from the initial score of $14.420\times10^{-3}$ to $9.146\times10^{-3}$ in 20 clicks. Our proposed G-BRS layers show a tendency for continuing improvement even after 20 clicks while the G-BRS-sb layer struggles to improve after 10 clicks due to the inability to make localized refinement.
\begin{table}[t]
  \small
  \centering
  \begin{tabular}{c|c|c|c|c}
  \toprule
Datasets&sb&bmsb&bmsb-m&bmconv 
\\\hline
\midrule
SBD & 0.843& 0.832& 0.853& 0.846 \\\hline
Cityscapes & 0.881& 0.889& 0.886& 0.883 \\\hline
Mapillary Vista & 0.737& 0.742& 0.739& 0.738 \\\hline
Composition-1k & 0.0125& \textbf{0.0108}& 0.0109& 0.0112 \\\hline
NYU-Depth-V2 & 0.955& \textbf{0.962}& 0.956& 0.955 \\\hline
  \end{tabular}
\caption{Top AUC using each G-BRS layer type without the consistency loss. Scores that outperform settings using $\mathcal{L}_{c}$ are in bold.}
\label{tab:table_loss_c}
\vspace{-3mm}
\end{table}
\subsection{Evaluation - Depth Estimation}
We evaluate on the test set of NYU-Depth-V2 dataset that consists of 654 RGB-D indoor images. We compute the standard metrics of ${\delta}_{1-3}$, Abs Rel, Sq Rel, RMSE and RMSE\textit{log} and include all results in the supplementary document. For simplicity, we report results for ${\delta}_{1}$ defined as ${\delta}_{t} = mean(max(\frac{{d}_{gt}}{d}, \frac{d}{{d}_{gt}}) < 1.25^{t})$, where ${d}_{gt}$ and $d$ denote the ground truth and predicted depth map respectively. Figure \ref{fig:de_nyu_depth_v2} shows that the G-BRS-bmconv layer achieves the best $\text{AUC}_{\delta}$ of $0.963$ and $0.897$ on the test set and its bottom 10\% instances. We improve the initial $\delta_{1}$ from $0.885$ to a near perfect score of $0.983$ in 10 clicks. For the bottom 10\% instances, there is also a drastic improvement of 74.8\% from $\delta_{1} = 0.551$ to $\delta_{1} = 0.963$.
\subsection{Ablation Study}
We perform the same experiments for all datasets without using the proposed consistency loss and show the top results in Table \ref{tab:table_loss_c}. By comparing Table \ref{tab:table_loss_c} with Figure \ref{fig:is_ss} and \ref{fig:mt_de}, we show that using consistency loss is beneficial for nearly all G-BRS settings. Results also show that settings of G-BRS-bmconv that achieve the top AUC for each dataset all utilize $\mathcal{L}_{c}$. Additional results for experiments with/without $\mathcal{L}_{c}$ are included in the supplementary document.
\begin{table}[t]
  \small
  \centering
  \begin{tabular}{c|c|c}
  \toprule
Datasets& 
\begin{tabular}{@{}c@{}}RGB-BRS\\\hline
	\begin{tabular}{P{0.85cm}|P{0.8cm}}
		AUC&SPC
	\end{tabular}
\end{tabular}& 
\begin{tabular}{@{}c@{}}G-BRS-bmconv\\\hline
	\begin{tabular}{P{0.85cm}|P{0.8cm}}
		AUC&SPC
	\end{tabular}
\end{tabular}\\\hline
\midrule
SBD & \begin{tabular}{P{0.85cm}|P{0.8cm}}0.851&1.542\end{tabular} & \begin{tabular}{P{0.85cm}|P{0.8cm}}0.859&0.584\end{tabular} \\\hline
Cityscapes & \begin{tabular}{P{0.85cm}|P{0.8cm}}0.882&8.361\end{tabular} & \begin{tabular}{P{0.85cm}|P{0.8cm}}0.869&5.727\end{tabular} \\\hline
Mapillary Vista & \begin{tabular}{P{0.85cm}|P{0.8cm}}0.675&8.125\end{tabular} & \begin{tabular}{P{0.85cm}|P{0.8cm}}0.673&5.130\end{tabular} \\\hline
Composition-1k & \begin{tabular}{P{0.85cm}|P{0.8cm}}0.0100&2.473\end{tabular} & \begin{tabular}{P{0.85cm}|P{0.8cm}}0.0108&1.383\end{tabular} \\\hline
NYU-Depth-V2 & \begin{tabular}{P{0.85cm}|P{0.8cm}}0.961&3.205\end{tabular} & \begin{tabular}{P{0.85cm}|P{0.8cm}}0.963&2.107\end{tabular} \\\hline
  \end{tabular}
\caption{Comparison between RGB-BRS and G-BRS-bmconv.}
\label{tab:table_rgb_brs}
\vspace{-4mm}
\end{table}
\subsection{Comparison with RGB-BRS}
We perform experiments using RGB-BRS with $\mathcal{L}_{c}$ for all datasets as discussed in Section \ref{sec:eval_is} and measure the AUC as well as the seconds per click (SPC). Experiments for speed measurement are run using a RTX 2080 Ti GPU. Table \ref{tab:table_rgb_brs} shows that despite the considerably higher inference time due to the need to backpropagate through the entire network, RGB-BRS and G-BRS-bmconv obtain comparable results. The additional memory consumption for RGB-BRS is also undesirable. For instance, RGB-BRS requires us to downsize the image resolution for semantic segmentation to $1024 \times 512$ to fit the memory limit (the same resolution is used for G-BRS-bmconv in this experiment for a fair comparison). As a result, we can see a drop of performance from the top AUC of 0.897 and 0.779 using G-BRS-bmconv (Figure \ref{fig:is_ss}) to an AUC of 0.882 and 0.675 using RGB-BRS for Cityscapes and Mapillary Vista respectively. RGB-BRS also has no flexibility for how backpropagating refinement is performed, preventing users from designing effective and efficient G-BRS layouts for different architectures.
\section{Conclusion}
In this work, we propose a novel set of Generalized Backpropagating Refinement Scheme (G-BRS) layers that bring significant improvement to the performance of state-of-the-art models with both global and localized modification of the intermediate features. By using a user-controlled attention mechanism during refinement, our proposed consistency loss achieves consistent improvement for various G-BRS settings. We show generality of our approach by targeting four different applications and converting the pretrained state-of-the-art architecture for each application into an interactive method with the corresponding G-BRS layer configuration. Our work shows promising directions for adding interactive capability to architectures used for many other computer vision applications.
{\small
\bibliographystyle{ieee_fullname}
\bibliography{egbib}
}

\clearpage

\begin{center}
       {\Large\textbf{Supplementary Document:\\
Generalizing Interactive Backpropagating Refinement for Dense Prediction Networks}}

       \vspace{0.5cm}
\end{center}
\setlength{\fboxsep}{0pt}
\pagenumbering{gobble}
\setcounter{section}{0}
\setcounter{figure}{0}
\setcounter{table}{0}
\section{Optimal learning rates for experiments}
Learning rate is one of the most important hyperparameter for backpropagation. Therefore, we search for the best learning rate settings for all experiments by using a subset of the test set from each dataset for evaluation using various learning rates. Specifically, we test for 10 learning rates ranging from $0.1\times{0.5}^0$ to $0.1\times{0.5}^{9}$. Tables \ref{tab:table_lr_sbd}-\ref{tab:table_lr_nyu} shows the obtained best learning rates used for our experiments. The number of layers included in the architecture is denoted as L1, L2 and L3. Architectures for semantic segmentation and image matting can contain up to 3 G-BRS layers, while architectures for interactive segmentation and depth estimation contain 1 G-BRS layer.
\section{Quantitative results for all experiments}
\subsection{Interactive and semantic segmentation}
We only report top scores achieved by each G-BRS layer type in the main paper due to the space limit. Tables \ref{tab:table_scores_sbd}-\ref{tab:table_scores_mapillary} show the AUC as well as maximum mIoU achieved in total number of clicks using all backpropagating refinement settings. We see that the proposed consistency loss $\mathcal{L}_{c}$ enables very consistent improvement for all G-BRS as well as RGB-BRS settings. The G-BRS-bmconv layer also consistently achieves the best AUC and maximum mIoU for segmentation tasks.
\subsection{Image matting}
We compute the following standard metrics for the task of image matting: Sum of Absolute Differences (SAD), Mean Squared Error (MSE), Gradient (Grad) and Connectivity (Conn) error. We report results on MSE in the main paper due to the space limit. For simplicity, here we report the metrics obtained for each G-BRS layer type with the layout (\#layers) that achieves the best scores. Figure \ref{fig:mt_supp} shows that the G-BRS-sb layer that uses the same layer architecture as \text{\textit{f}-BRS} has limited ability for refinement comparing to other G-BRS layers. This is due to G-BRS-sb's lack of ability for localized modification of the features. We also observe that settings that utilize the consistency loss $\mathcal{L}_{c}$ generally achieves more stable and accurate results. \\
\indent Interestingly, our results indicate that RGB-BRS achieves best overall quantitative scores for image matting with or without $\mathcal{L}_{c}$. This finding is helpful for the community as backpropagating refinement has not been implemented on the task of image matting before. However, RGB-BRS does introduce additional inference time and memory consumption. In a later section, we show a detailed report on Seconds Per Click (SPC) for all backpropagating refinement settings so the trade-off between accuracy and efficiency can be decided by the user.
\subsection{Depth estimation}
We compute the following standard metrics for the task of depth estimation: ${\delta}_{1-3}$, Abs Rel, Sq Rel, RMSE and RMSE\textit{log}. We only report results on ${\delta}_{1}$ in the main paper due to the space limit. Here we report the AUC computed on each metric over the total number of clicks as well as the best score achieved in the total number of clicks. Table \ref{tab:table_scores_nyu} shows that the G-BRS-bmconv layer achieves the best metrics in all settings and $\mathcal{L}_{c}$ provides consistent improvement. 
\section{Running time analysis}
Inference speed is an important factor for interactive applications. In this section, we provide the Seconds Per Click (SPC) measured for all backpropagating refinement settings using the proposed $\mathcal{L}_{c}$. We select 100 test instances from each dataset and perform 10 clicks on each instance. A PC with an AMD Ryzen Threadripper 1920X CPU and a RTX 2080 Ti GPU is used in this experiment. Tables \ref{tab:table_spc_sbd}-\ref{tab:table_spc_nyu} show the obtained SPC for all backpropagating refinement settings. First, We see that the overall difference between the inference time of different G-BRS layers is very small. In addition, the additional inference time for inserting multiple G-BRS layers is very small, which makes the usage of multiple G-BRS layers more desirable if it can lead to improvement in performance. As expected, RGB-BRS has considerably higher inference time and previous experiments show that it only achieves better performance in the task of image matting. For the task of semantic segmentation, we recognize that the obtained inference time is not ideal for real-time interactive responses. This is due to our selection of a state-of-the-art architecture that prioritizes accuracy over efficiency. Our experiments show that our approach can effectively improve results of top-performing models. As a result, users have the flexibility to apply our approach to other architectures and design the G-BRS configuration to accommodate the specific needs of their applications.
\section{Qualitative Comparisons}
\begin{figure*}[t]
 \centering
  \begin{subfigure}[t]{0.45\linewidth}
    \includegraphics[width=\linewidth]{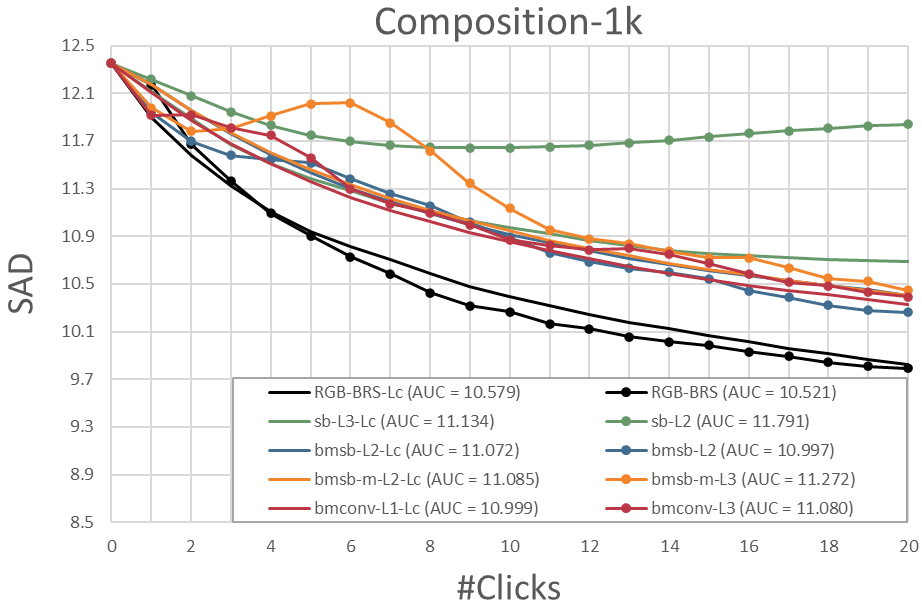}
  \end{subfigure}
  \begin{subfigure}[t]{0.45\linewidth}
    \includegraphics[width=\linewidth]{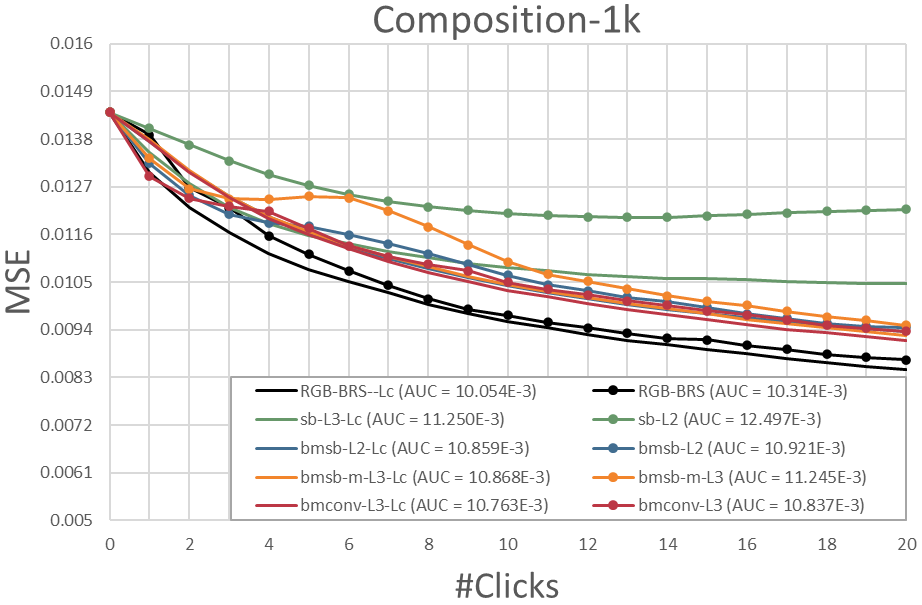}
  \end{subfigure}
  \begin{subfigure}[t]{0.45\linewidth}
    \includegraphics[width=\linewidth]{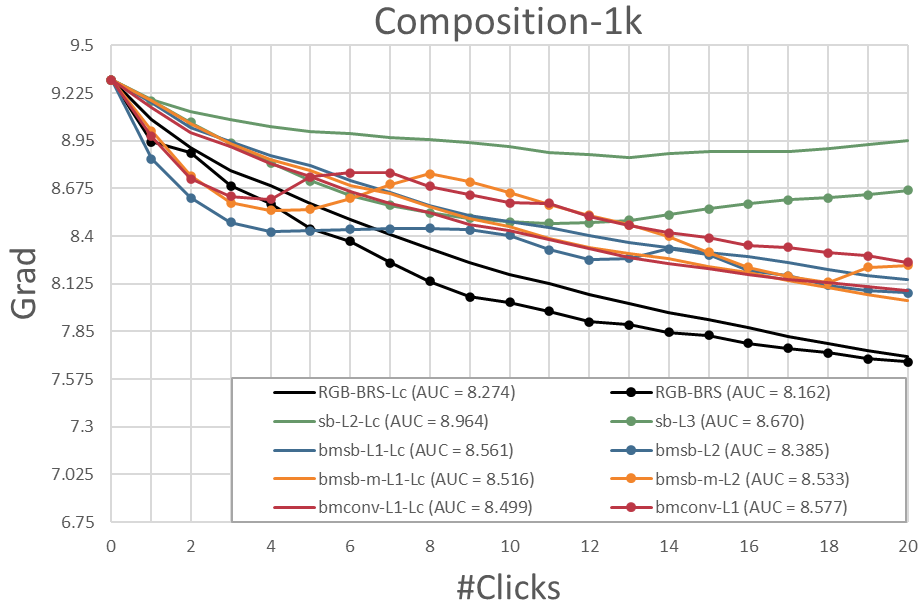}
  \end{subfigure}
  \begin{subfigure}[t]{0.45\linewidth}
    \includegraphics[width=\linewidth]{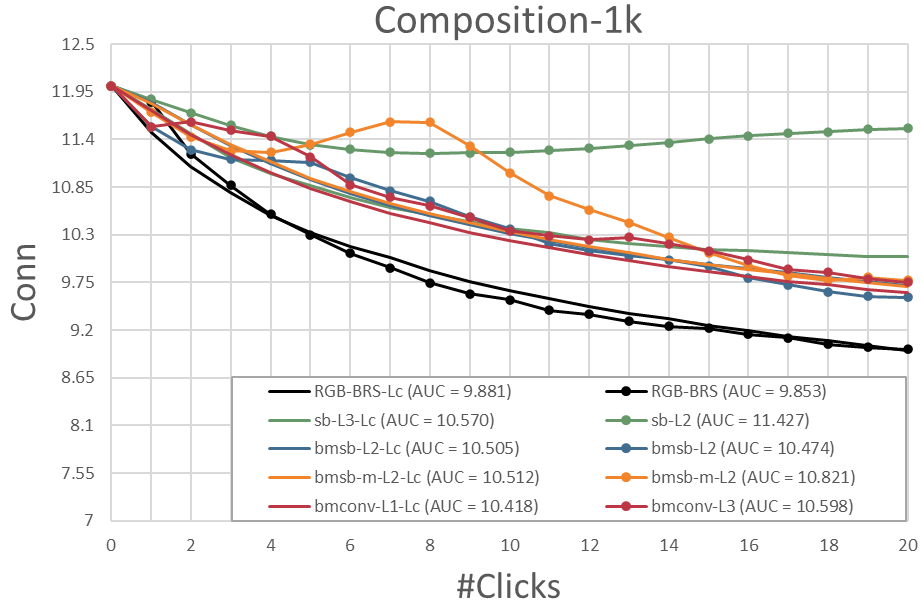}
  \end{subfigure}
  \vspace{-0.2cm}
  \caption{SAD, MSE, Grad and Conn on \textbf{Composition-1k} for image matting. Results denoted with $\mathcal{L}_{c}$ are obtained using the consistency loss.}
  \vspace{-0.4cm}
  \label{fig:mt_supp}
\end{figure*}

\begin{figure*}[h]
  \centering
   \begin{subfigure}[t]{0.24\linewidth}
    \includegraphics[width=\linewidth]{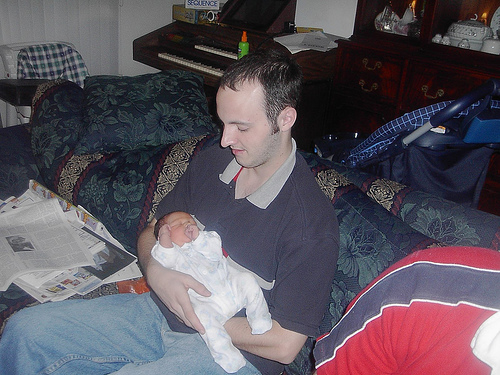}
  \end{subfigure}
  \begin{subfigure}[t]{0.24\linewidth}
    \includegraphics[width=\linewidth]{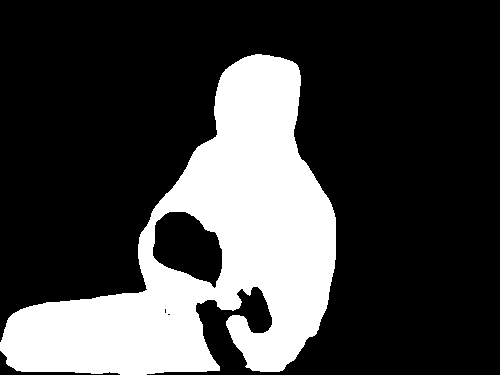}
  \end{subfigure}
  \begin{subfigure}[t]{0.24\linewidth}
    \includegraphics[width=\linewidth]{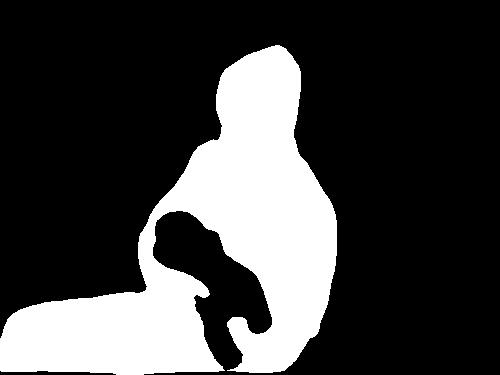}
  \end{subfigure}
  \vspace{1mm}
  \begin{subfigure}[t]{0.24\linewidth}
    \includegraphics[width=\linewidth]{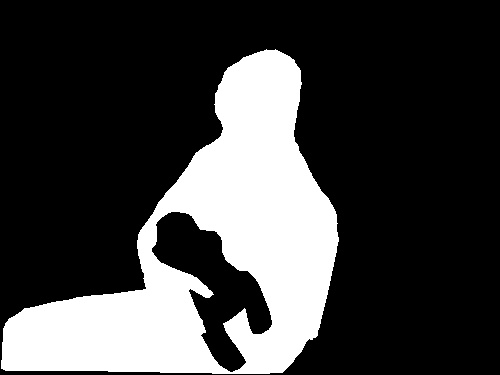}
  \end{subfigure}
   \begin{subfigure}[t]{0.24\linewidth}
    \includegraphics[width=\linewidth]{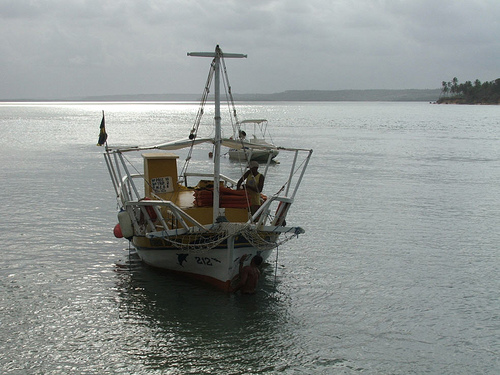}
  \end{subfigure}
  \begin{subfigure}[t]{0.24\linewidth}
    \includegraphics[width=\linewidth]{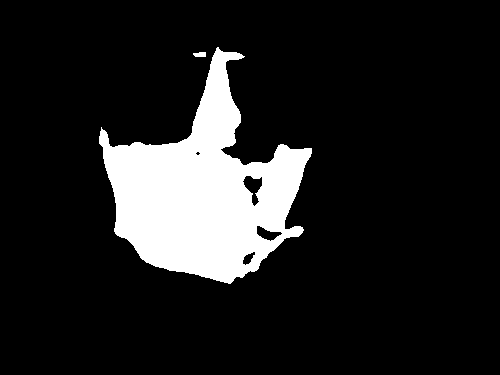}
  \end{subfigure}
  \begin{subfigure}[t]{0.24\linewidth}
    \includegraphics[width=\linewidth]{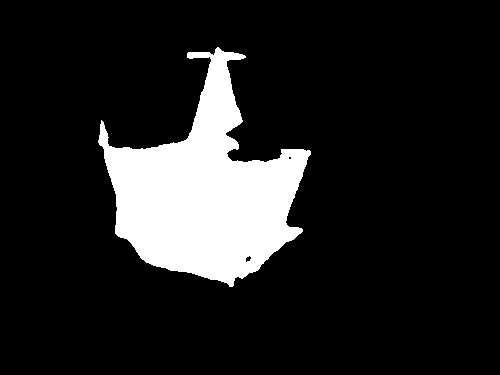}
  \end{subfigure}
  \vspace{1mm}
  \begin{subfigure}[t]{0.24\linewidth}
    \includegraphics[width=\linewidth]{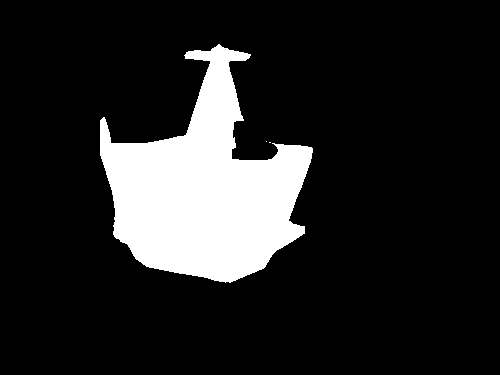}
  \end{subfigure}
   \begin{subfigure}[t]{0.24\linewidth}
    \includegraphics[width=\linewidth]{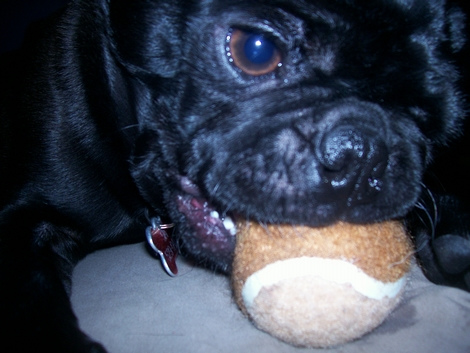}
  \end{subfigure}
  \begin{subfigure}[t]{0.24\linewidth}
    \includegraphics[width=\linewidth]{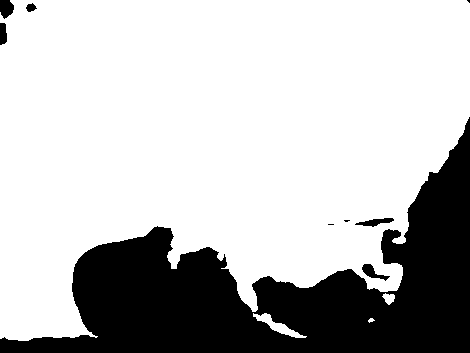}
  \end{subfigure}
  \begin{subfigure}[t]{0.24\linewidth}
    \includegraphics[width=\linewidth]{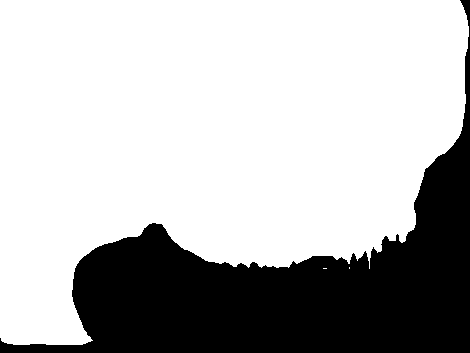}
  \end{subfigure}
  \vspace{1mm}
  \begin{subfigure}[t]{0.24\linewidth}
    \includegraphics[width=\linewidth]{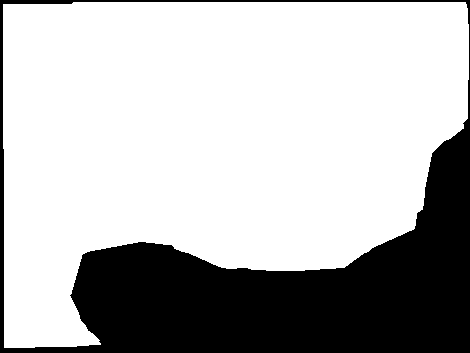}
  \end{subfigure}
   \begin{subfigure}[t]{0.24\linewidth}
    \includegraphics[width=\linewidth]{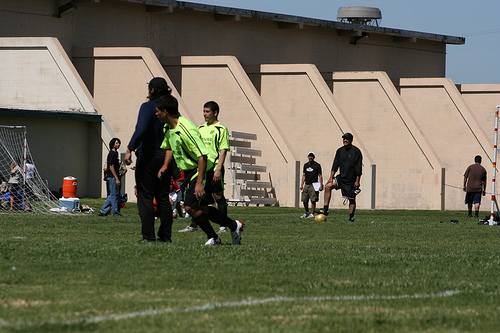}
  \end{subfigure}
  \begin{subfigure}[t]{0.24\linewidth}
    \includegraphics[width=\linewidth]{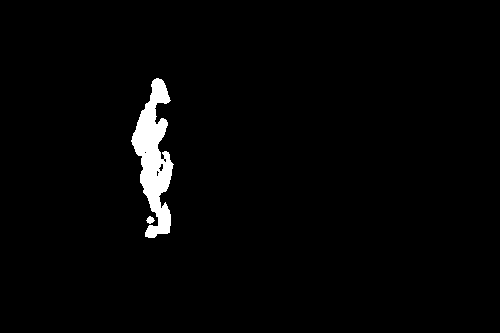}
  \end{subfigure}
  \begin{subfigure}[t]{0.24\linewidth}
    \includegraphics[width=\linewidth]{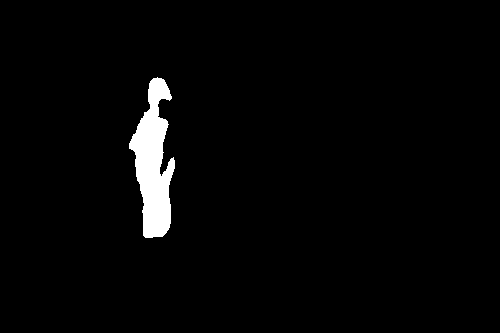}
  \end{subfigure}
  \vspace{1mm}
  \begin{subfigure}[t]{0.24\linewidth}
    \includegraphics[width=\linewidth]{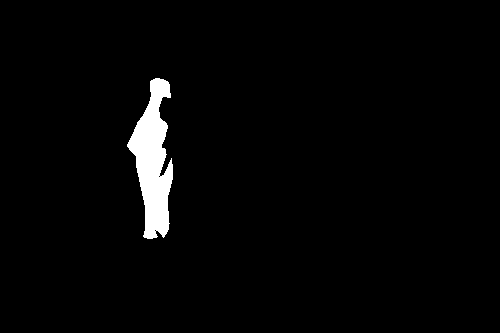}
  \end{subfigure}
   \begin{subfigure}[t]{0.24\linewidth}
    \includegraphics[width=\linewidth]{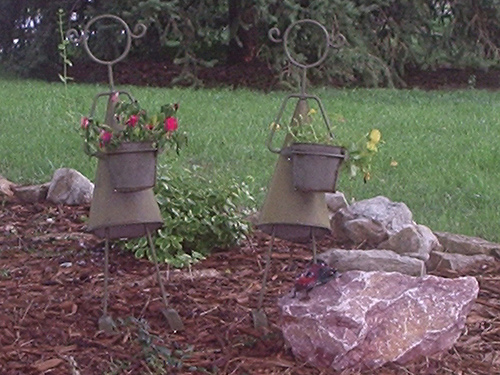}
  \end{subfigure}
  \begin{subfigure}[t]{0.24\linewidth}
    \includegraphics[width=\linewidth]{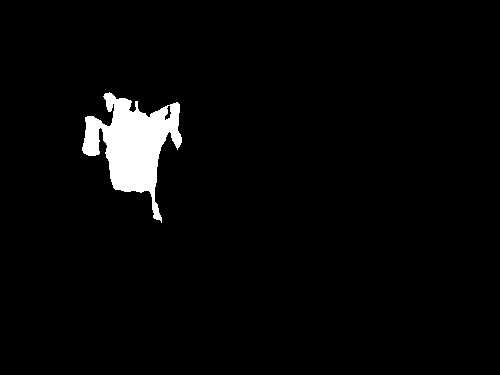}
  \end{subfigure}
  \begin{subfigure}[t]{0.24\linewidth}
    \includegraphics[width=\linewidth]{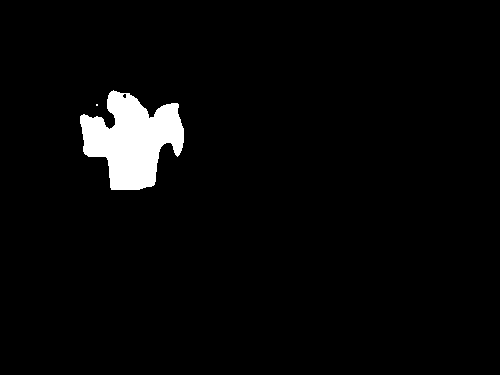}
  \end{subfigure}
  \vspace{1mm}
  \begin{subfigure}[t]{0.24\linewidth}
    \includegraphics[width=\linewidth]{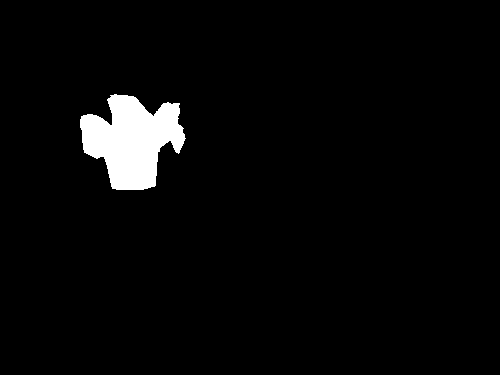}
  \end{subfigure}
   \begin{subfigure}[t]{0.24\linewidth}
    \includegraphics[width=\linewidth]{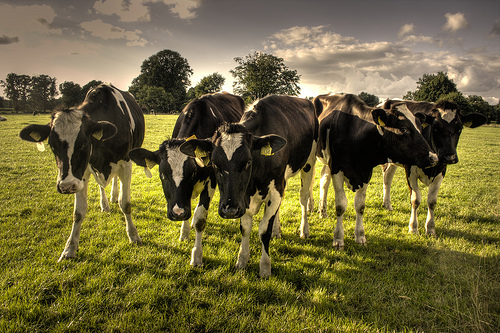}
    \caption*{Input image}
  \end{subfigure}
  \begin{subfigure}[t]{0.24\linewidth}
    \includegraphics[width=\linewidth]{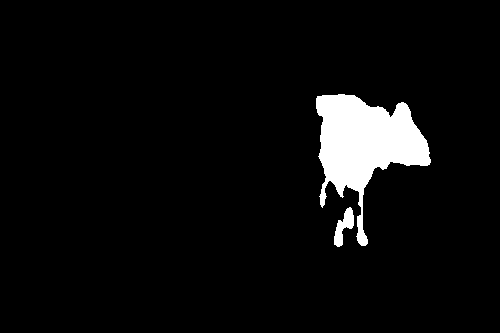}
    \caption*{G-BRS-sb (\textit{f}-BRS)}
  \end{subfigure}
  \begin{subfigure}[t]{0.24\linewidth}
    \includegraphics[width=\linewidth]{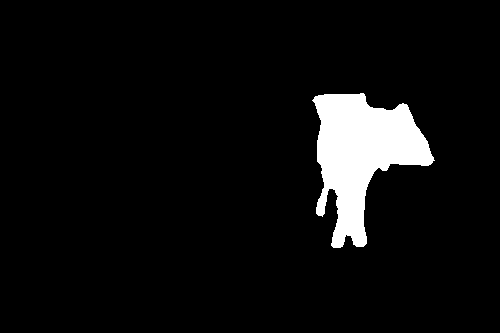}
    \caption*{G-BRS-bmconv}
  \end{subfigure}
  \begin{subfigure}[t]{0.24\linewidth}
    \includegraphics[width=\linewidth]{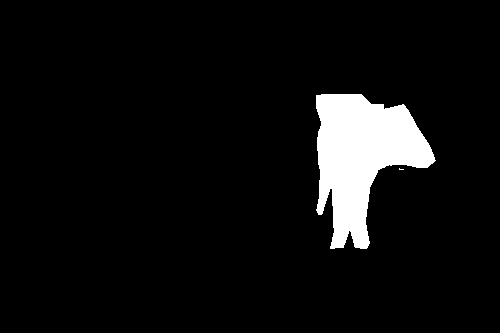}
    \caption*{Ground truth}
  \end{subfigure}
  \caption{Qualitative comparison between performance of the G-BRS-sb (\textit{f}-BRS) layer and our G-BRS-bmconv layer for the task of interactive segmentation on \textbf{SBD}.}
  \label{fig:qualitative_compare_sbd}
  \vspace{-4mm}
\end{figure*}

\begin{figure*}[h]
  \centering
   \begin{subfigure}[t]{0.24\linewidth}
    \includegraphics[width=\linewidth]{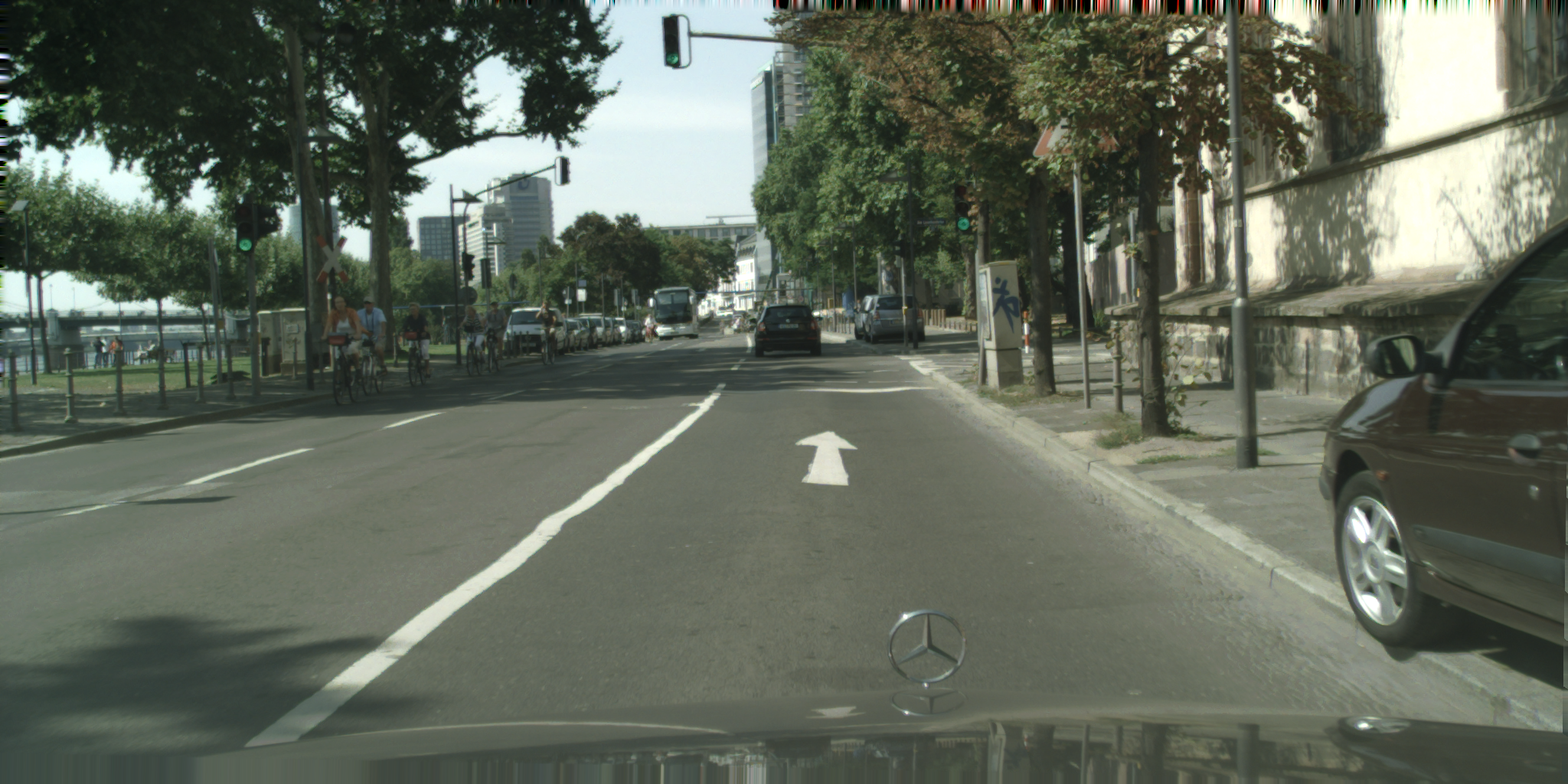}
  \end{subfigure}
  \begin{subfigure}[t]{0.24\linewidth}
    \includegraphics[width=\linewidth]{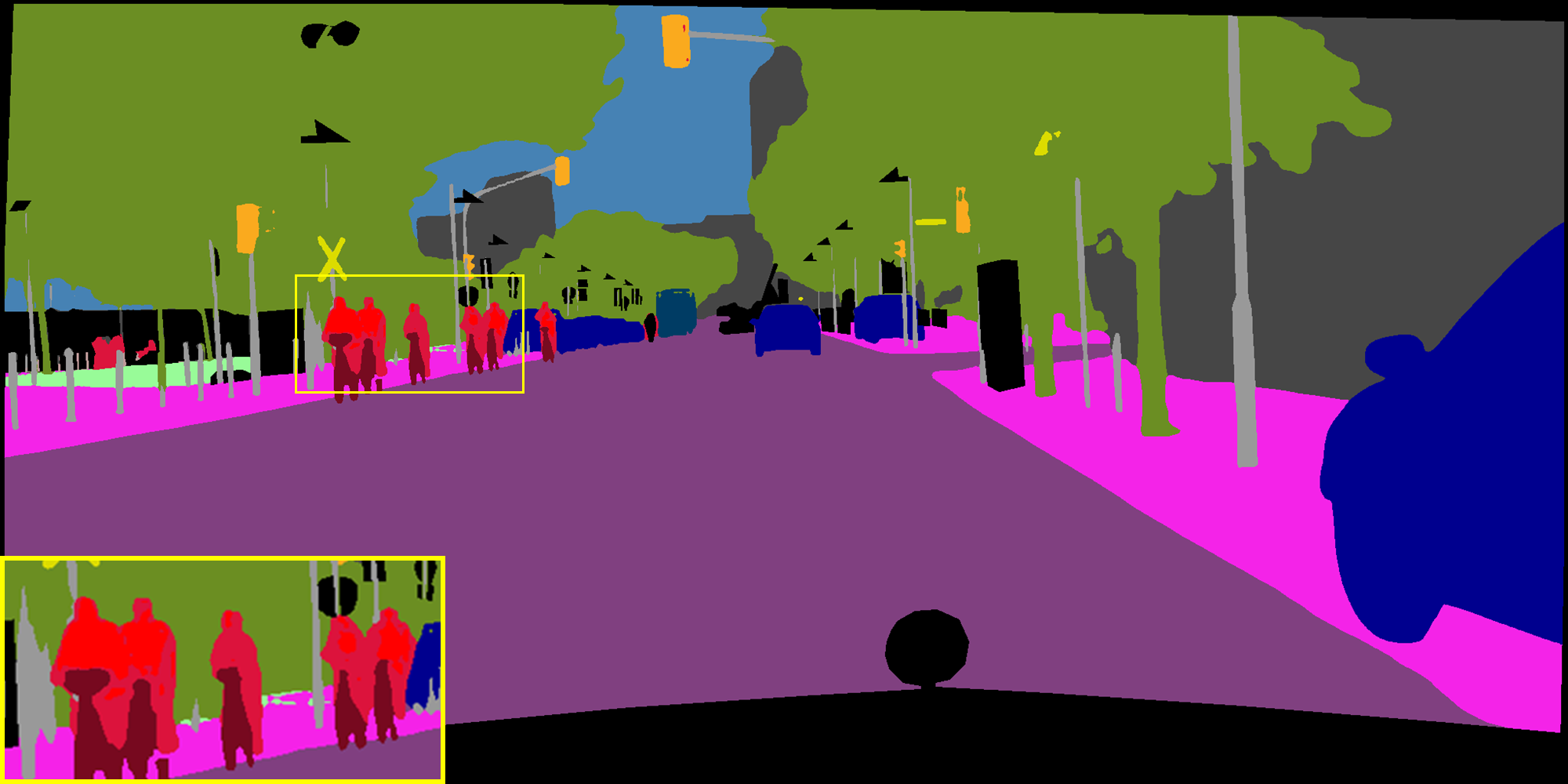}
  \end{subfigure}
  \begin{subfigure}[t]{0.24\linewidth}
    \includegraphics[width=\linewidth]{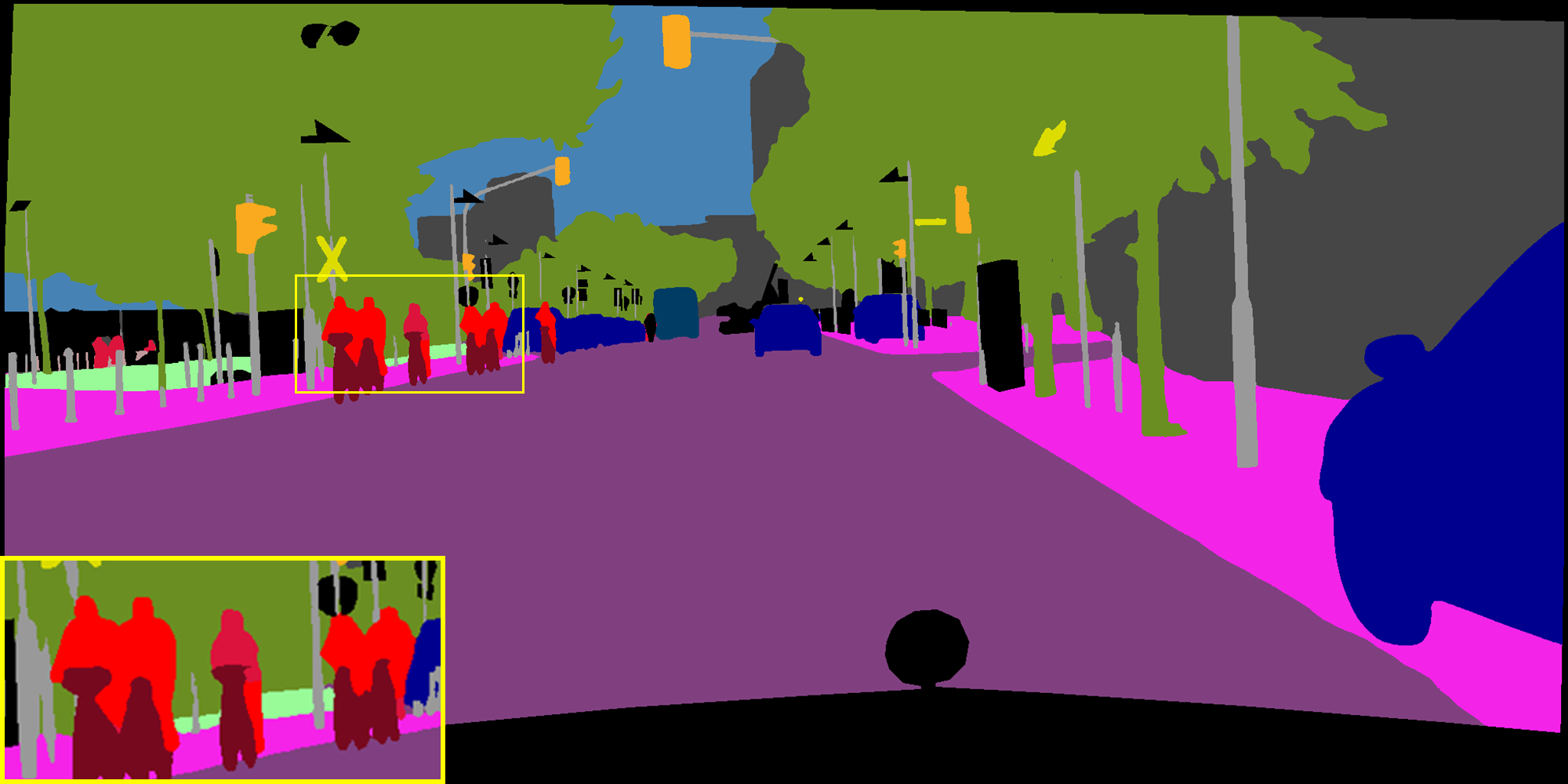}
  \end{subfigure}
  \vspace{1mm}
  \begin{subfigure}[t]{0.24\linewidth}
    \includegraphics[width=\linewidth]{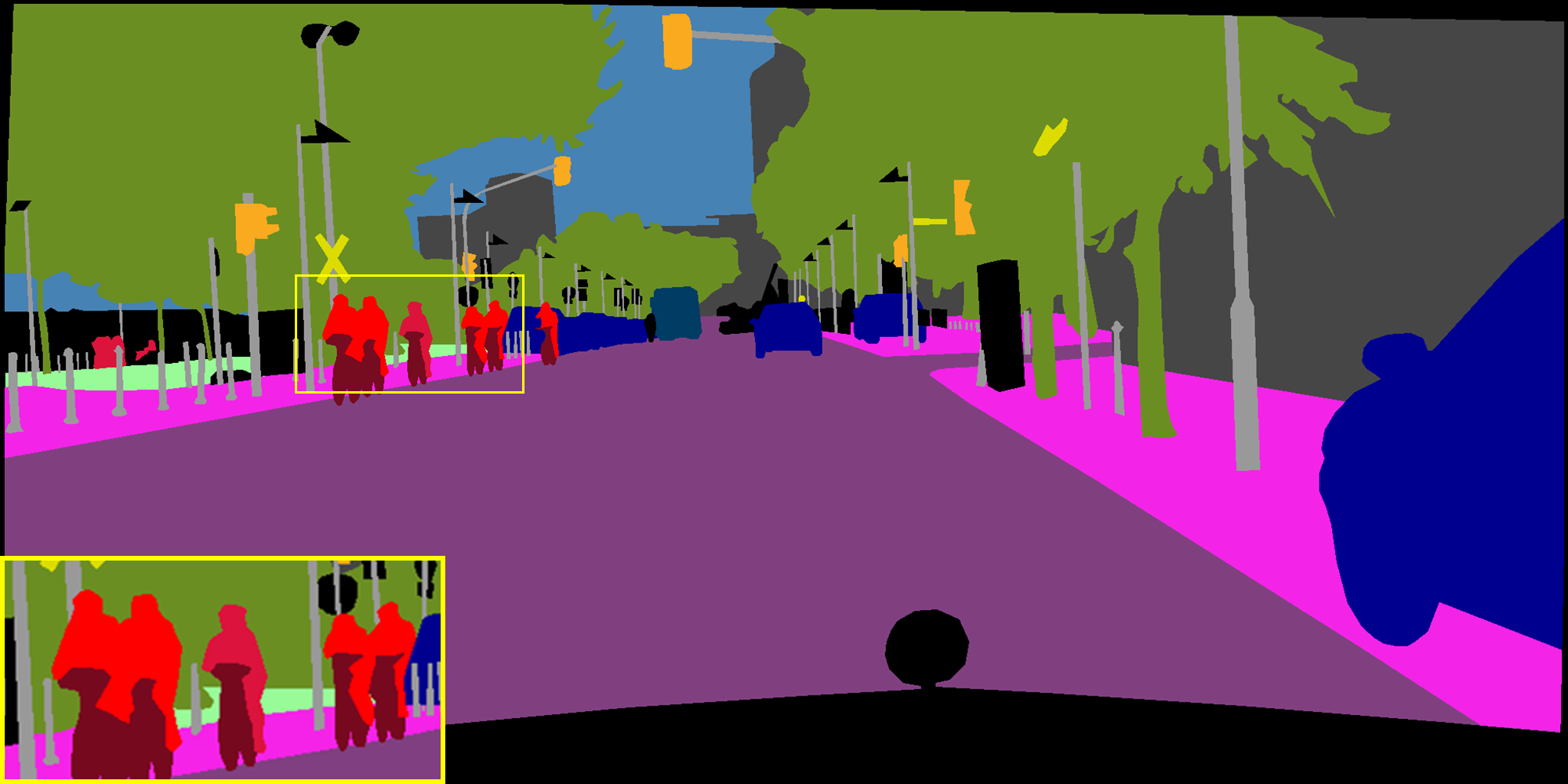}
  \end{subfigure}
   \begin{subfigure}[t]{0.24\linewidth}
    \includegraphics[width=\linewidth]{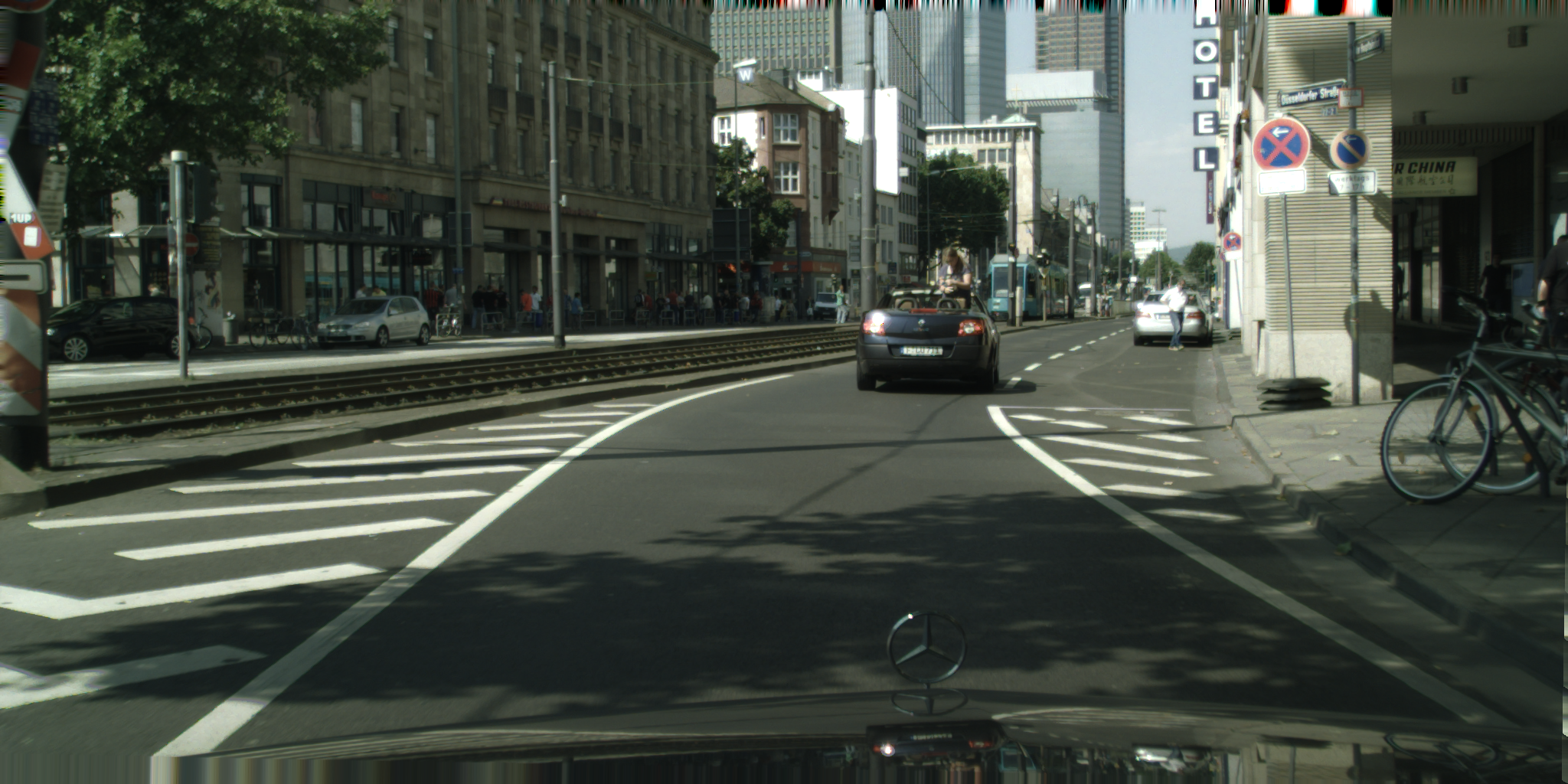}
  \end{subfigure}
  \begin{subfigure}[t]{0.24\linewidth}
    \includegraphics[width=\linewidth]{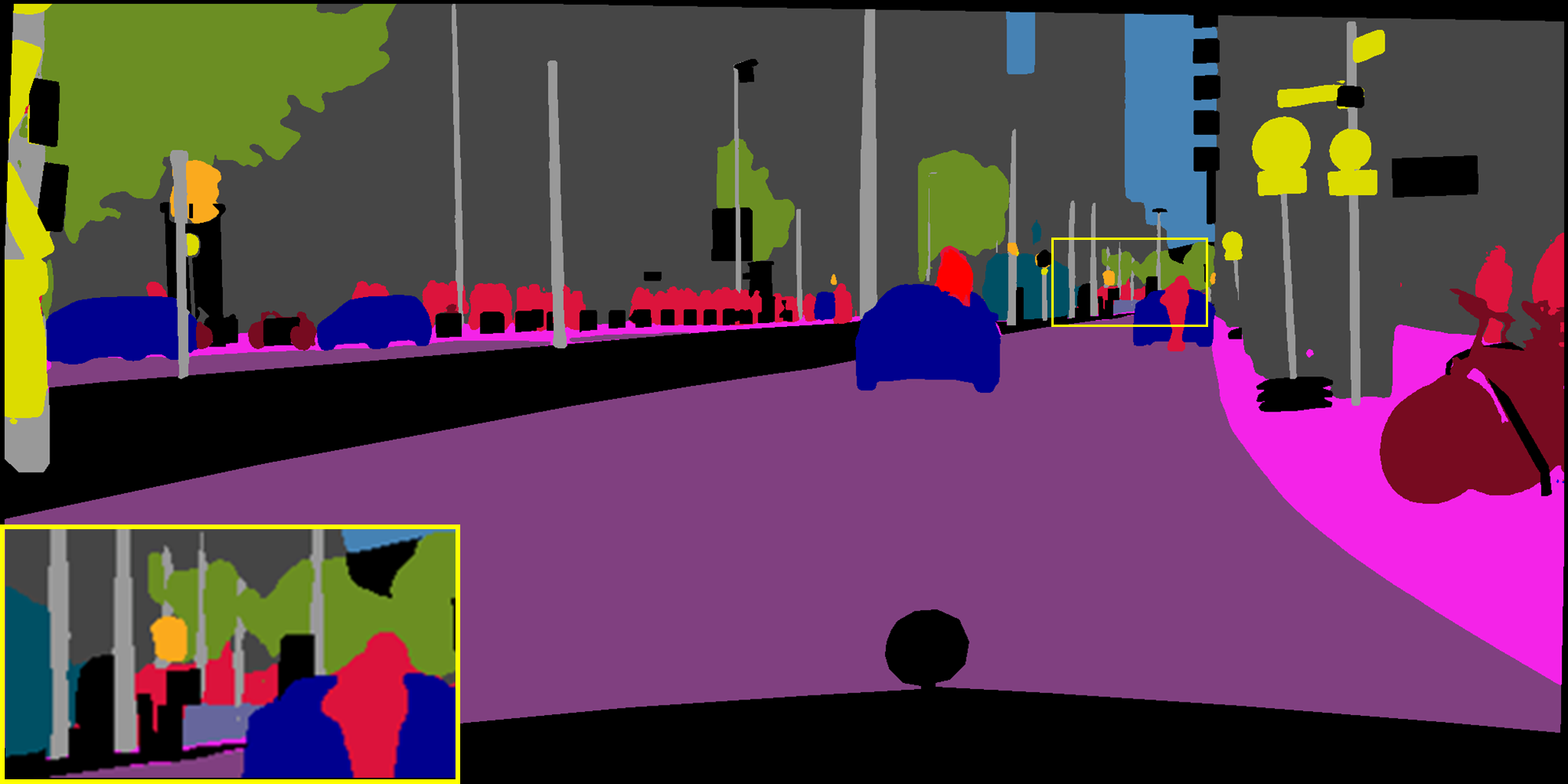}
  \end{subfigure}
  \begin{subfigure}[t]{0.24\linewidth}
    \includegraphics[width=\linewidth]{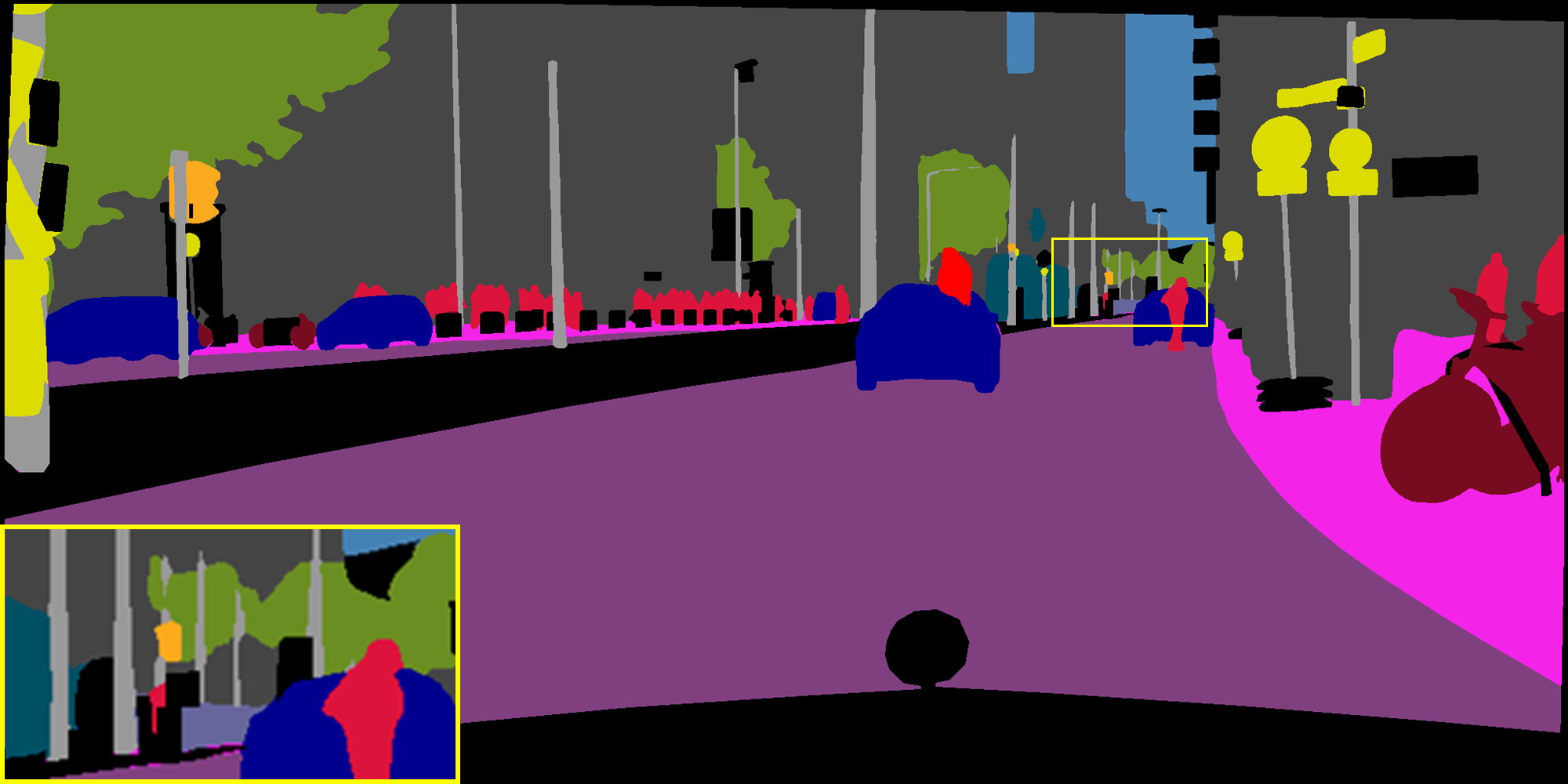}
  \end{subfigure}
  \vspace{1mm}
  \begin{subfigure}[t]{0.24\linewidth}
    \includegraphics[width=\linewidth]{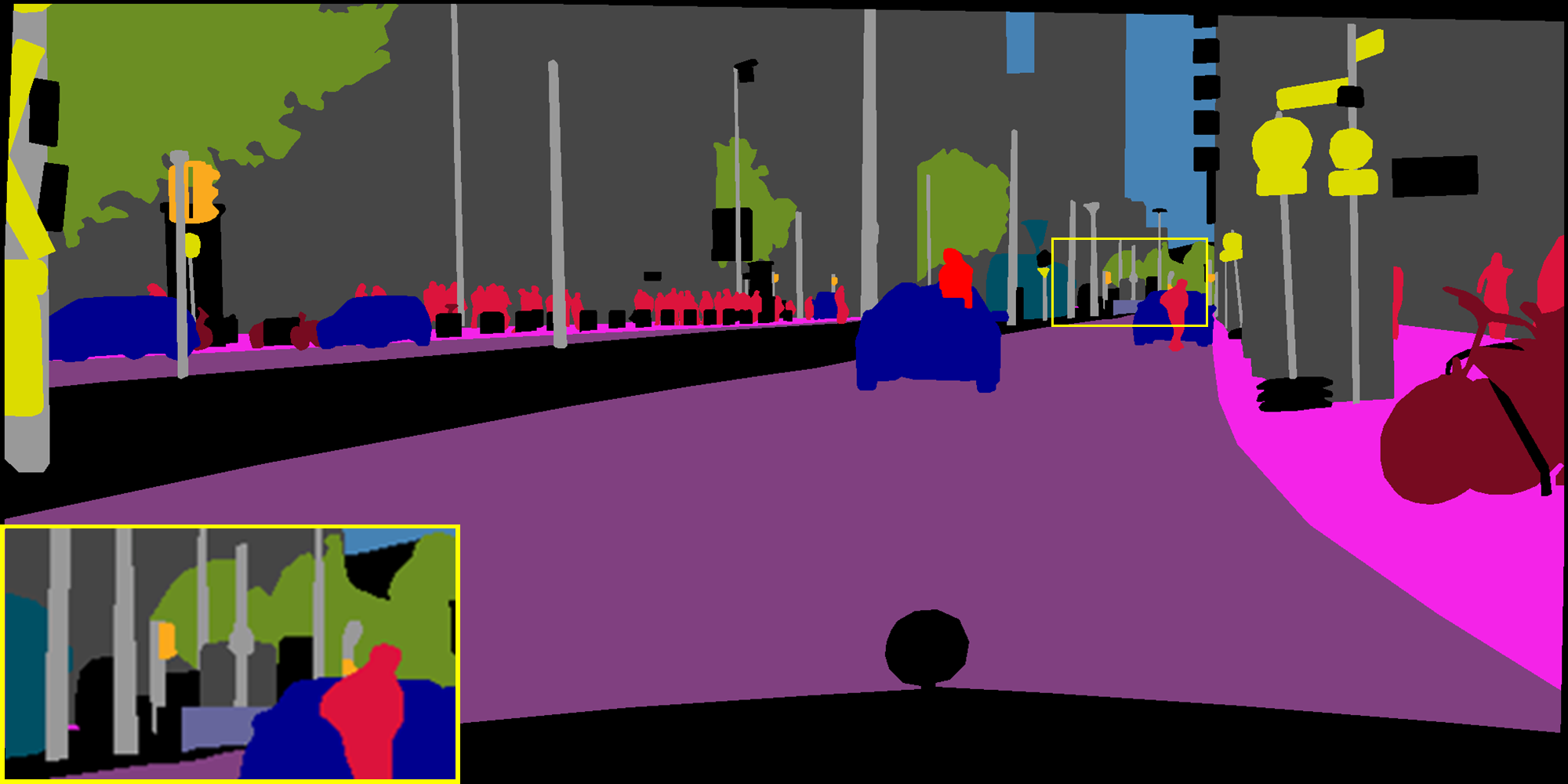}
  \end{subfigure}
   \begin{subfigure}[t]{0.24\linewidth}
    \includegraphics[width=\linewidth]{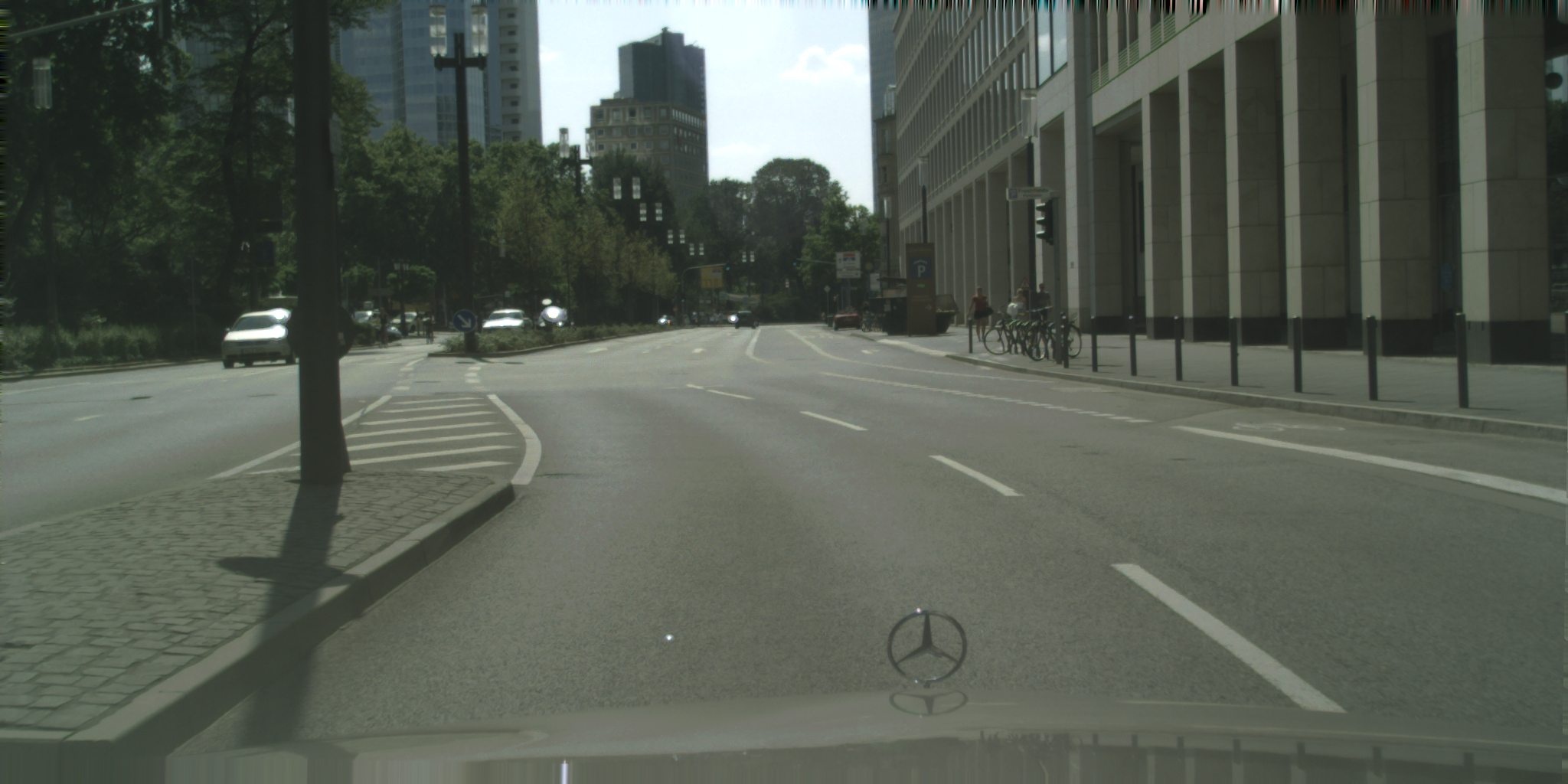}
  \end{subfigure}
  \begin{subfigure}[t]{0.24\linewidth}
    \includegraphics[width=\linewidth]{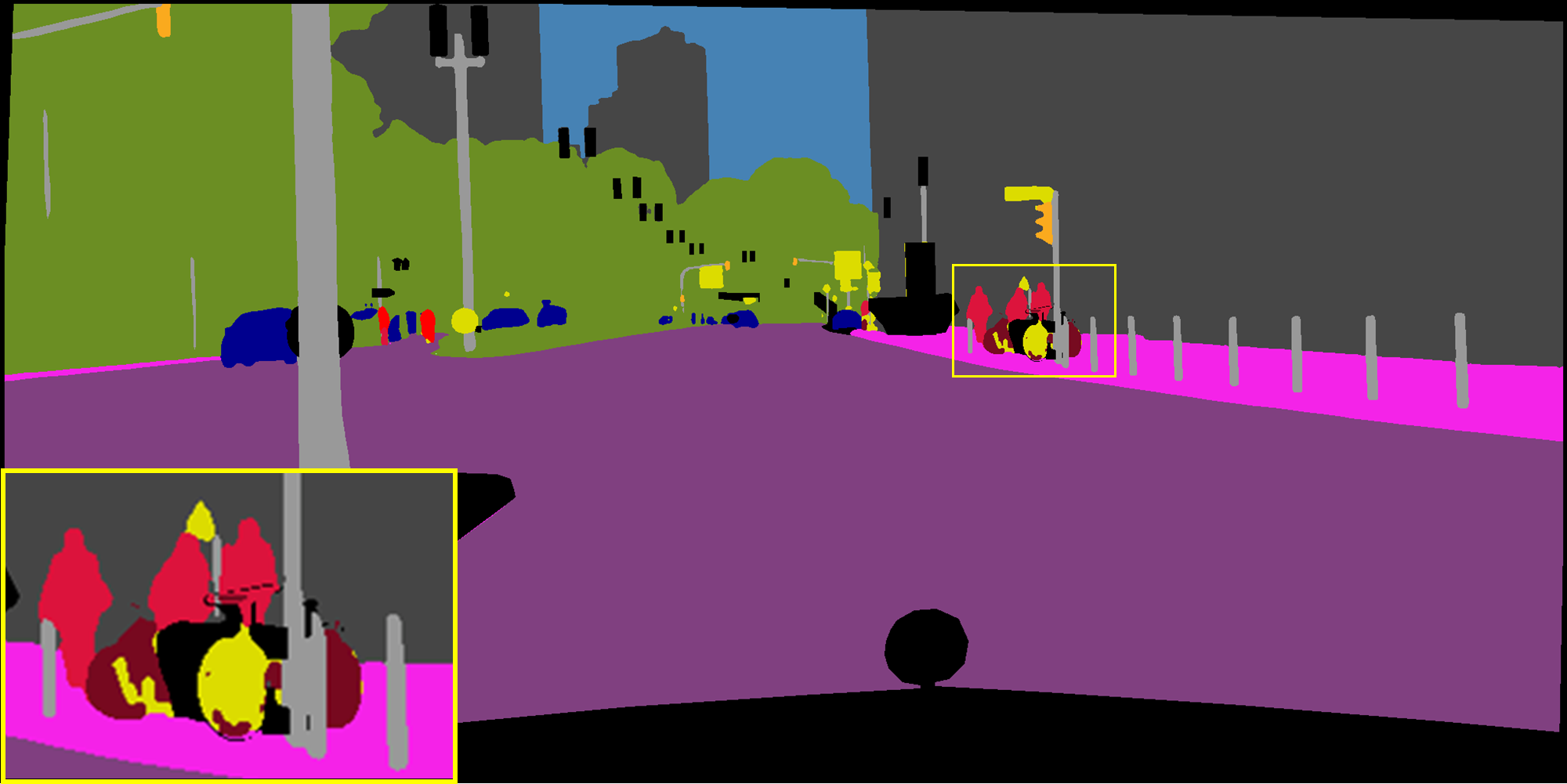}
  \end{subfigure}
  \begin{subfigure}[t]{0.24\linewidth}
    \includegraphics[width=\linewidth]{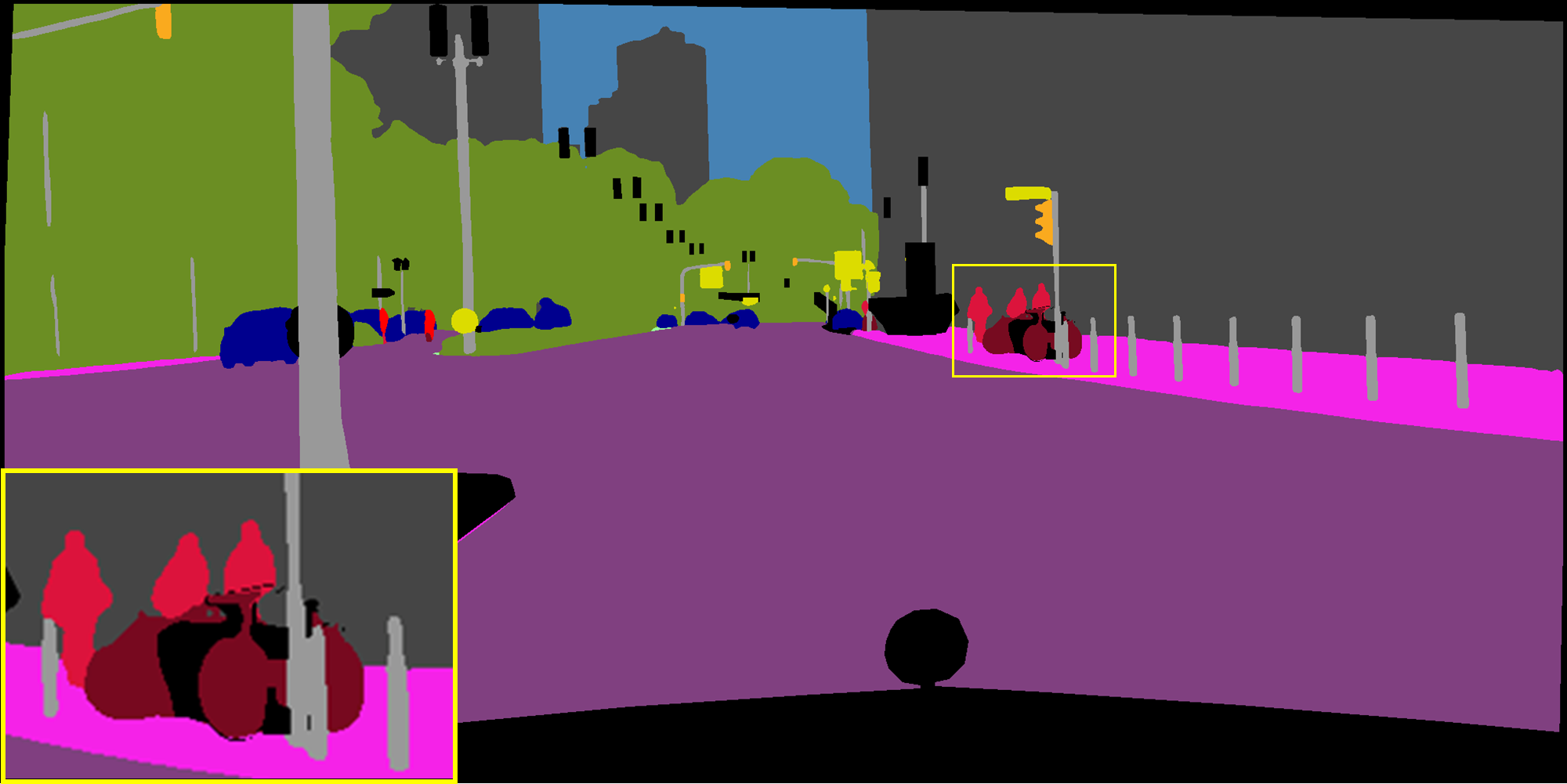}
  \end{subfigure}
  \vspace{1mm}
  \begin{subfigure}[t]{0.24\linewidth}
    \includegraphics[width=\linewidth]{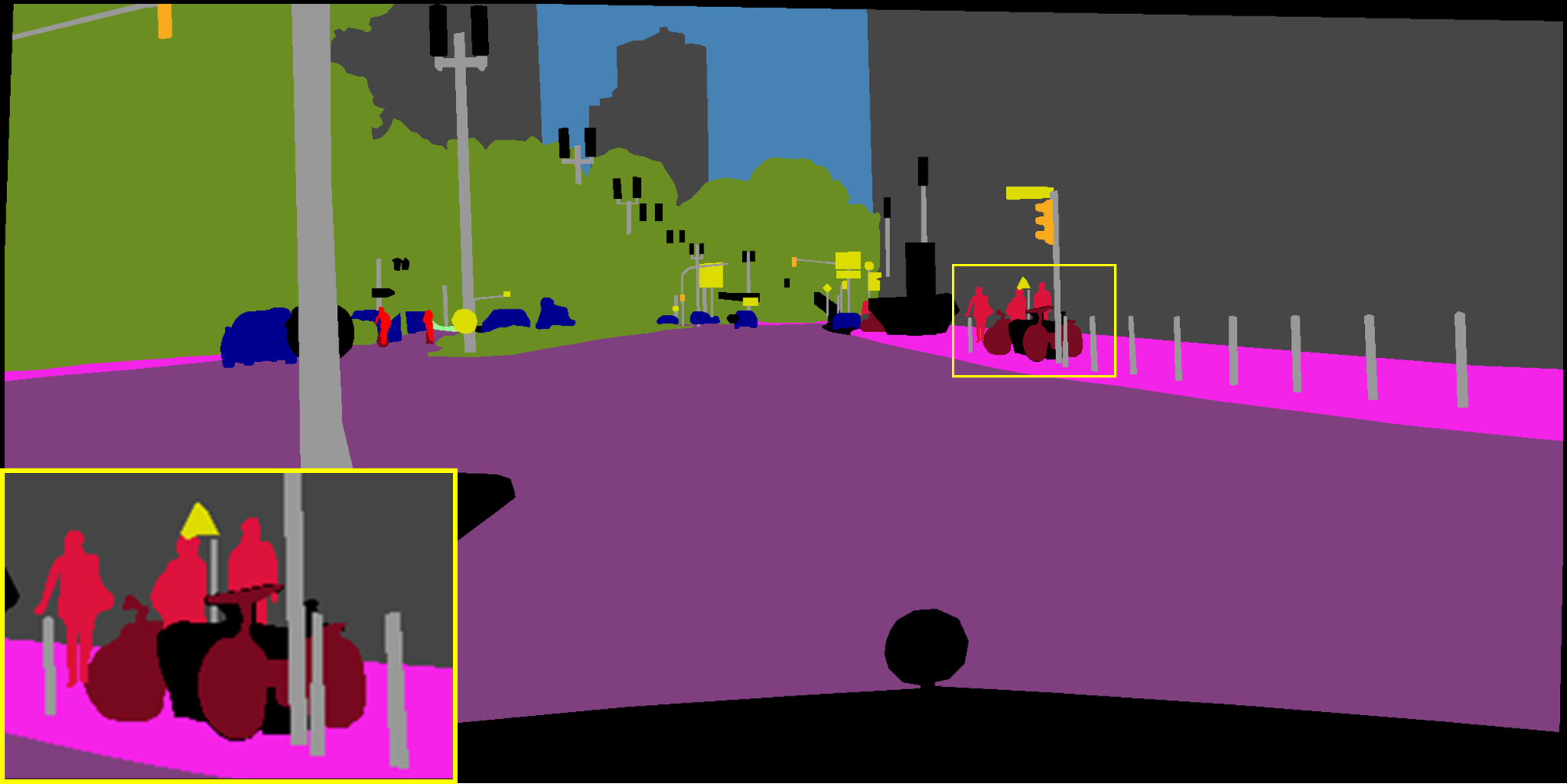}
  \end{subfigure}
   \begin{subfigure}[t]{0.24\linewidth}
    \includegraphics[width=\linewidth]{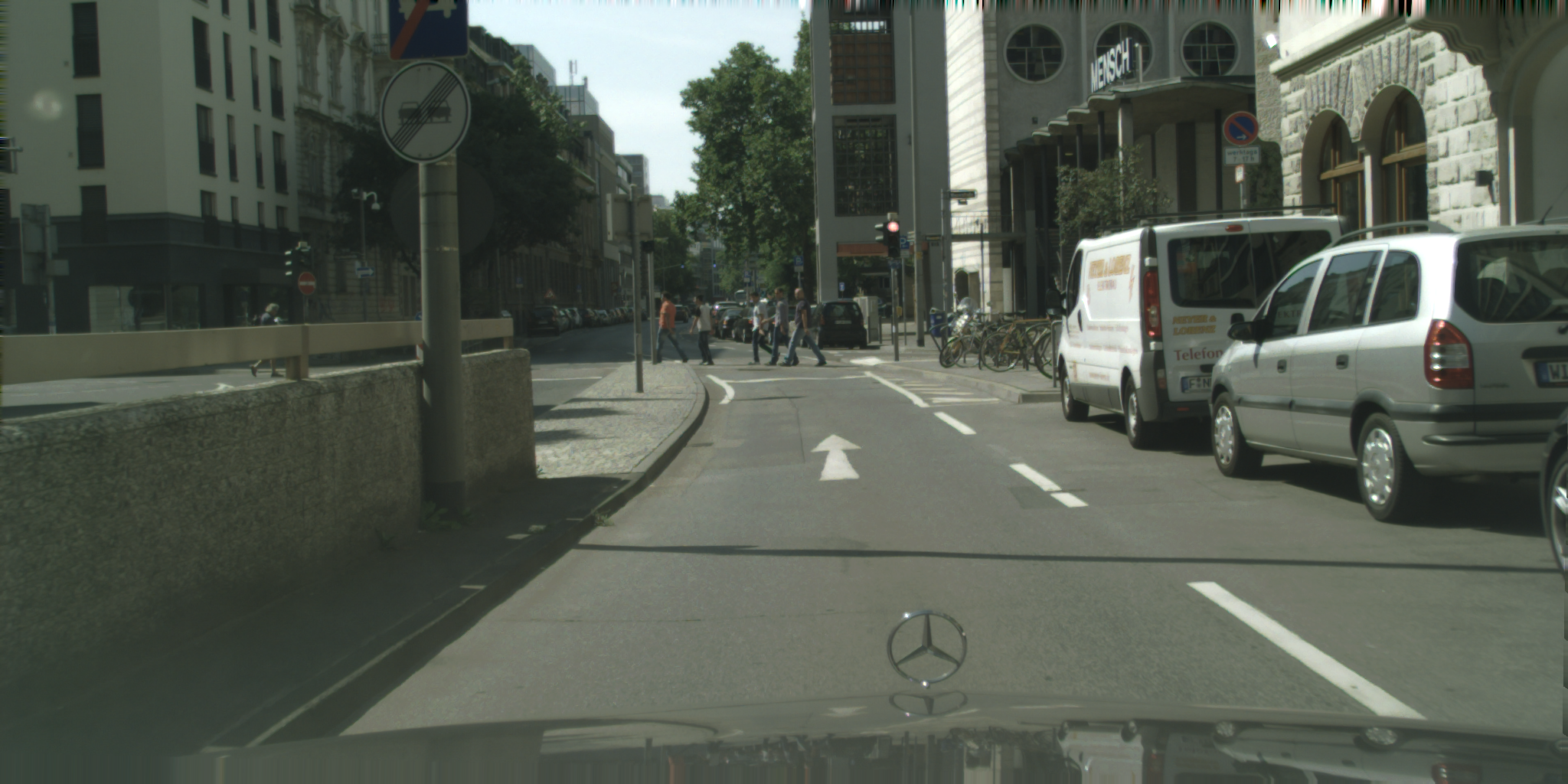}
    \caption*{Input image}
  \end{subfigure}
  \begin{subfigure}[t]{0.24\linewidth}
    \includegraphics[width=\linewidth]{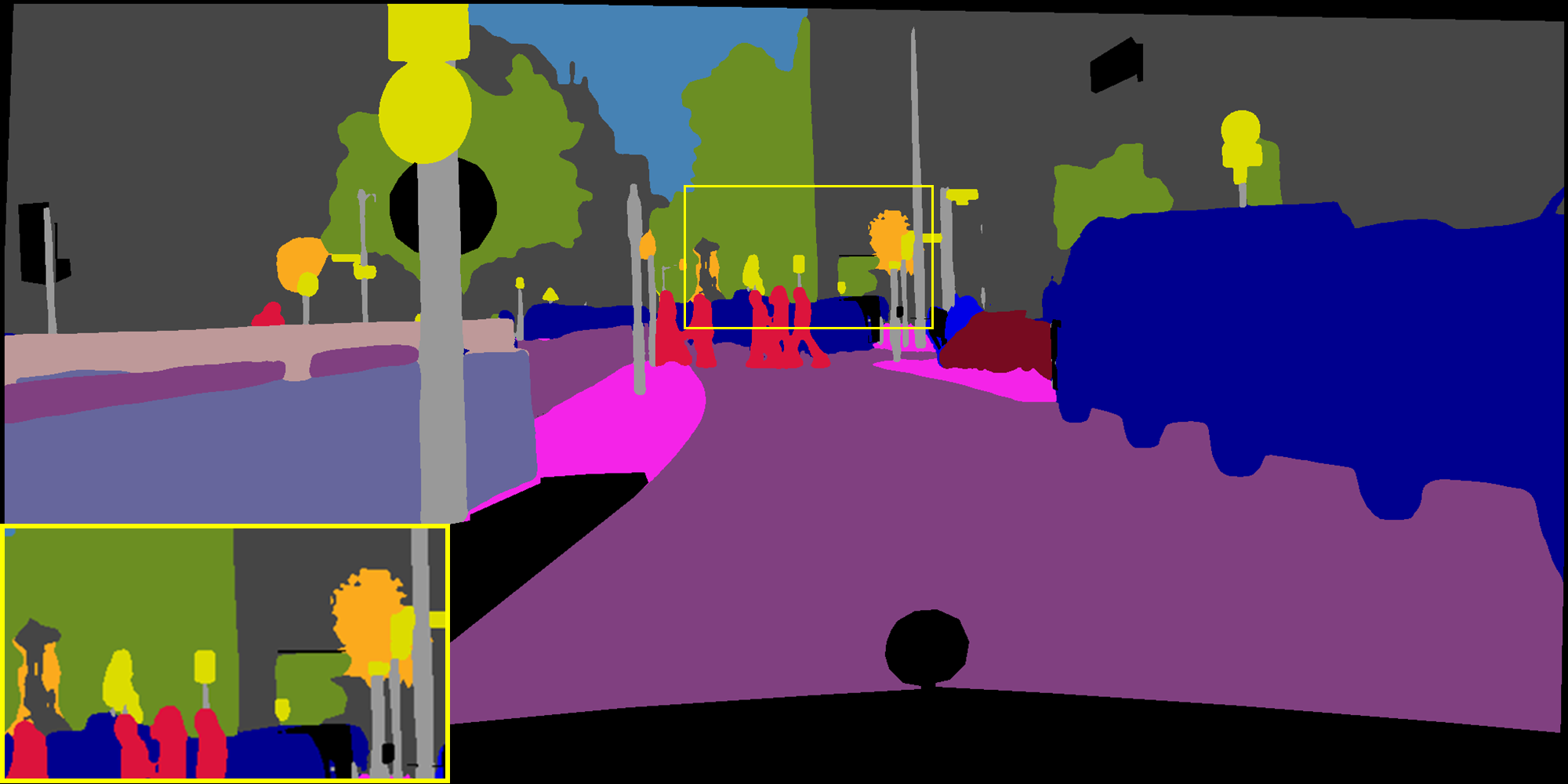}
    \caption*{G-BRS-sb}
  \end{subfigure}
  \begin{subfigure}[t]{0.24\linewidth}
    \includegraphics[width=\linewidth]{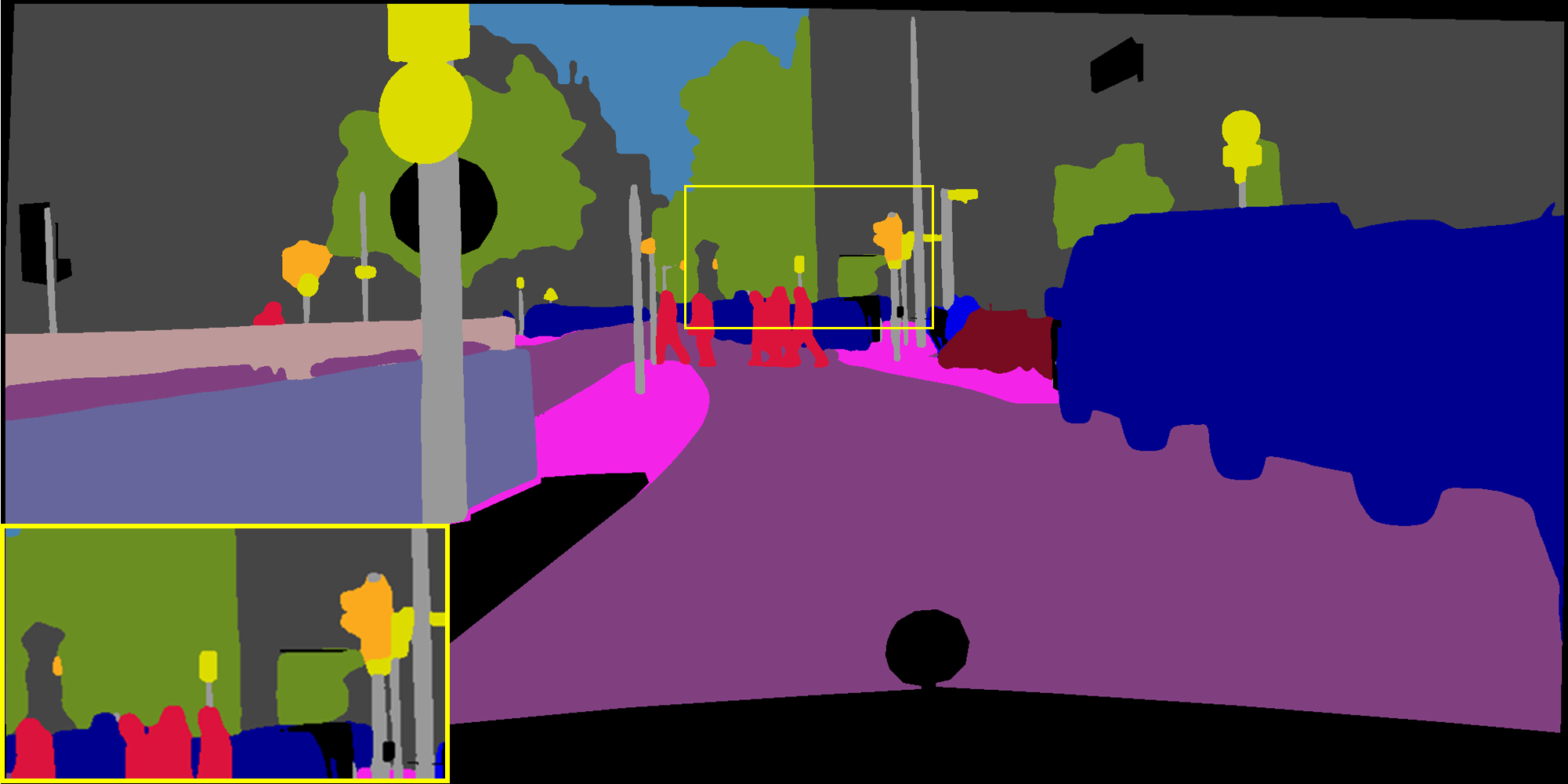}
    \caption*{G-BRS-bmconv}
  \end{subfigure}
  \begin{subfigure}[t]{0.24\linewidth}
    \includegraphics[width=\linewidth]{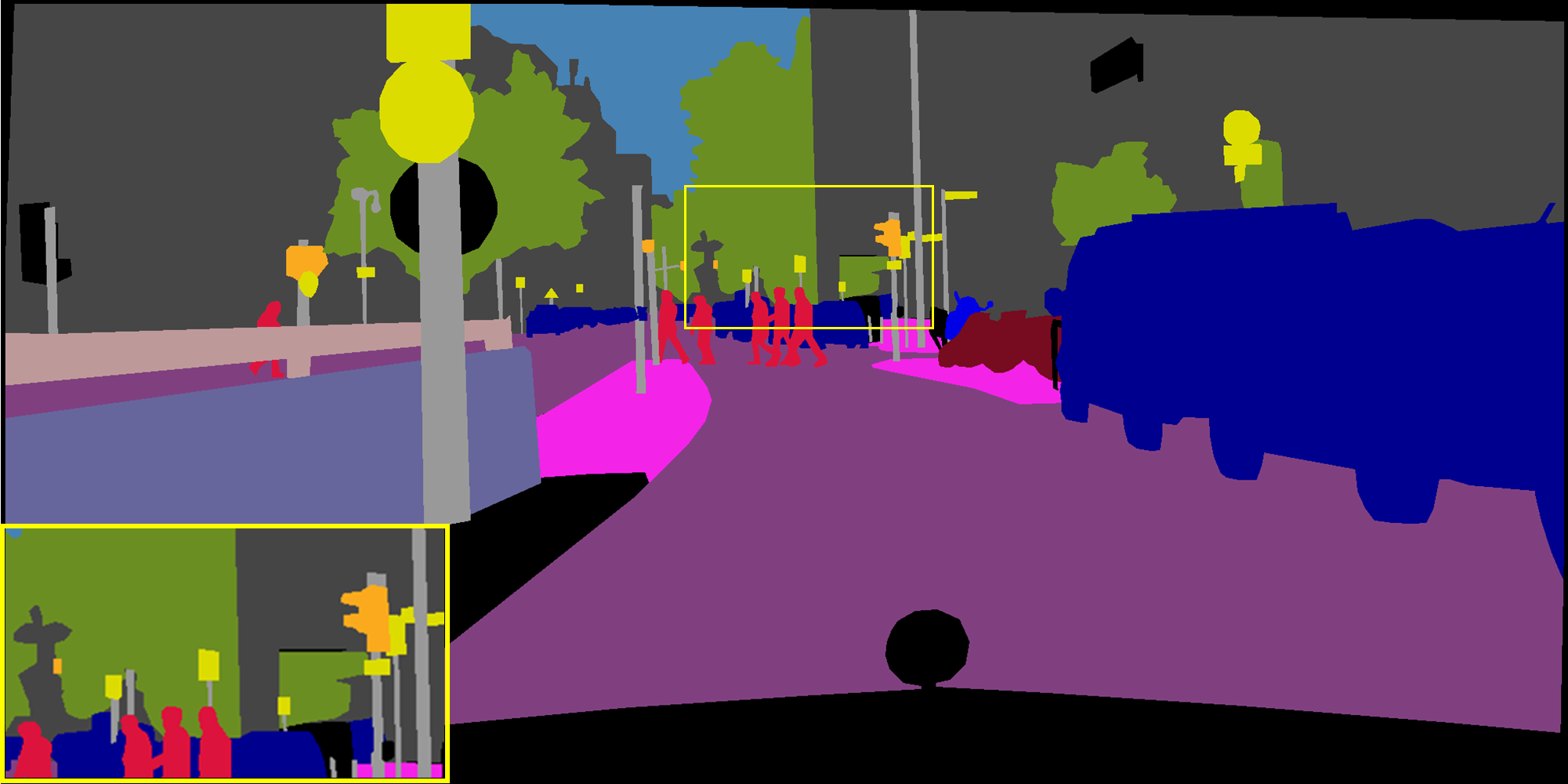}
    \caption*{Ground truth}
  \end{subfigure}
  \caption{Qualitative comparison between performance of the G-BRS-sb layer and our G-BRS-bmconv layer for the task of semantic segmentation on \textbf{Cityscapes}. Since both methods are highly accurate with G-BRS-bmconv providing more refined details, we highlight the differences in the yellow zoomed window. Invalid labels are shown in black. Best viewed in magnification.}
  \label{fig:qualitative_compare_cityscapes}
  \vspace{-4mm}
\end{figure*}

\begin{figure*}[h]
  \centering
   \begin{subfigure}[t]{0.24\linewidth}
    \includegraphics[width=\linewidth]{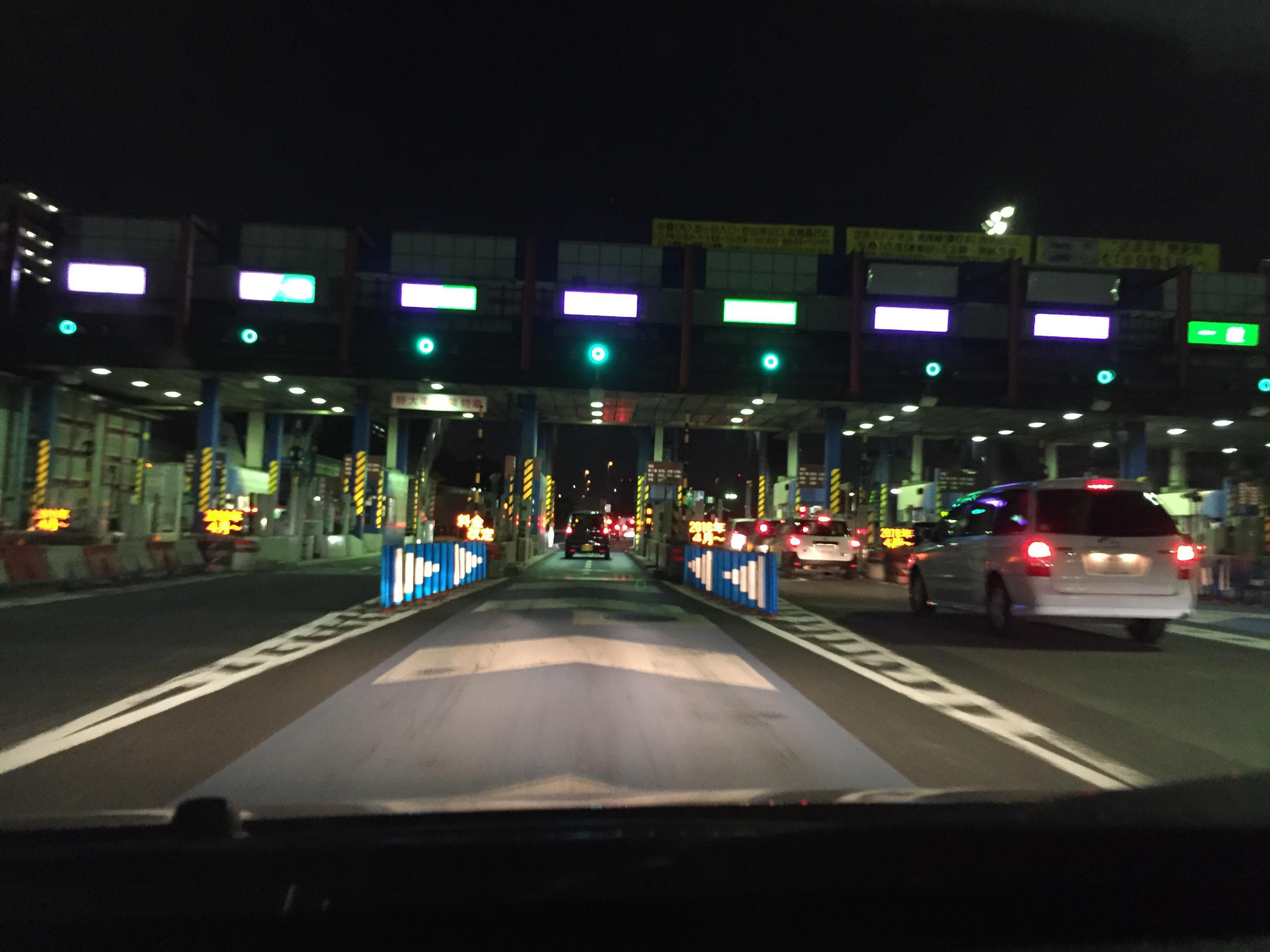}
  \end{subfigure}
  \begin{subfigure}[t]{0.24\linewidth}
    \includegraphics[width=\linewidth]{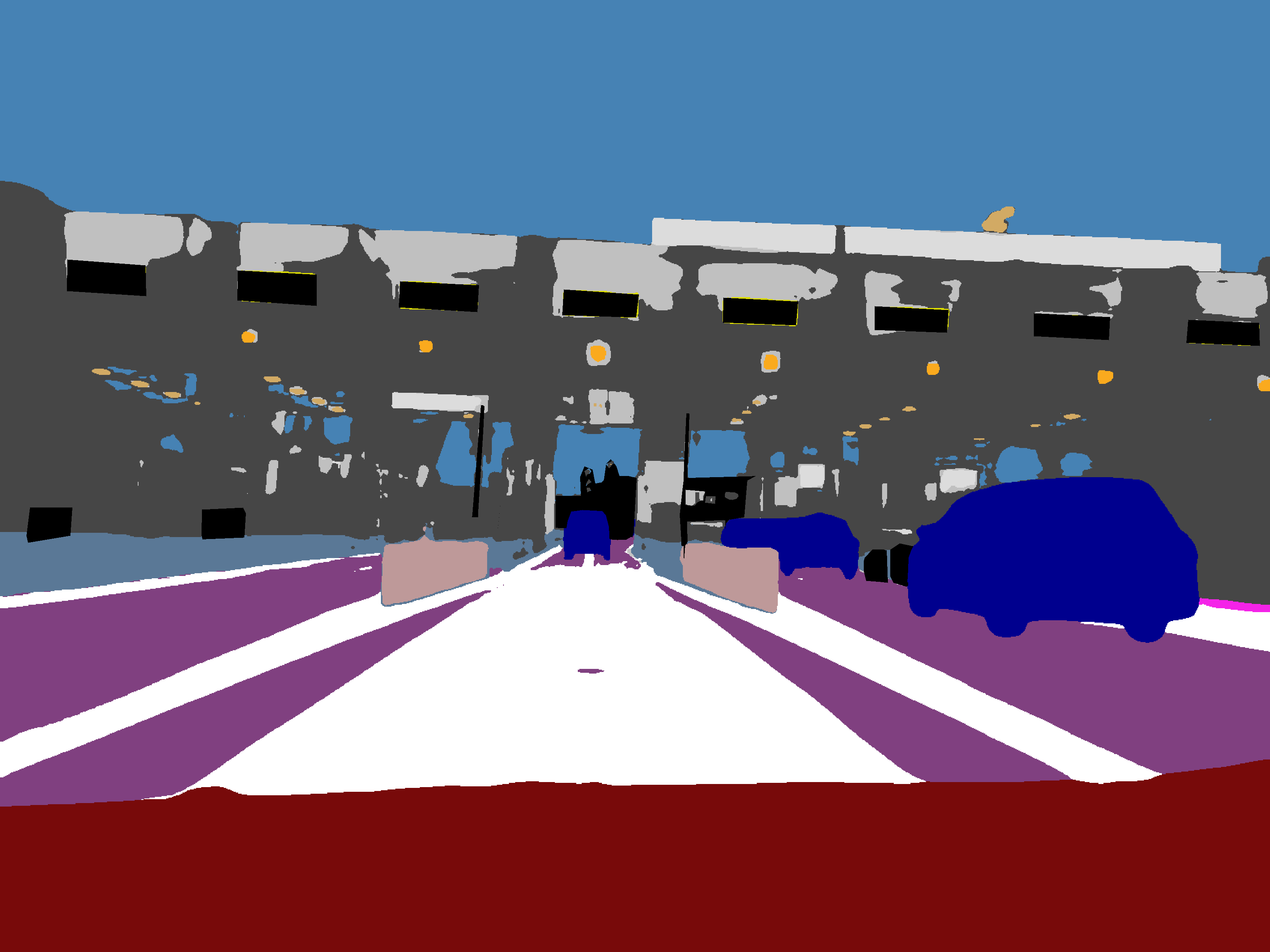}
  \end{subfigure}
  \begin{subfigure}[t]{0.24\linewidth}
    \includegraphics[width=\linewidth]{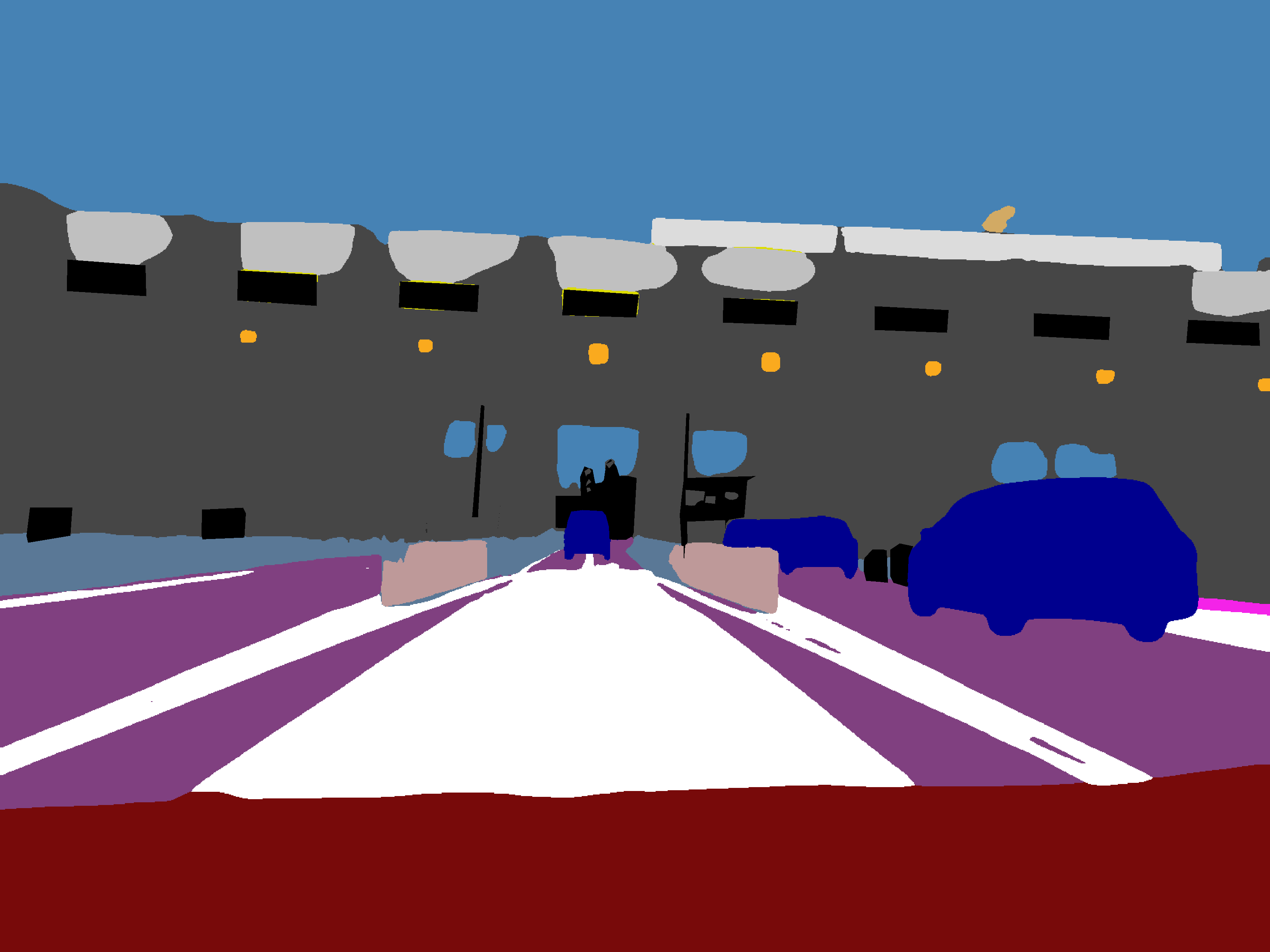}
  \end{subfigure}
  \vspace{1mm}
  \begin{subfigure}[t]{0.24\linewidth}
    \includegraphics[width=\linewidth]{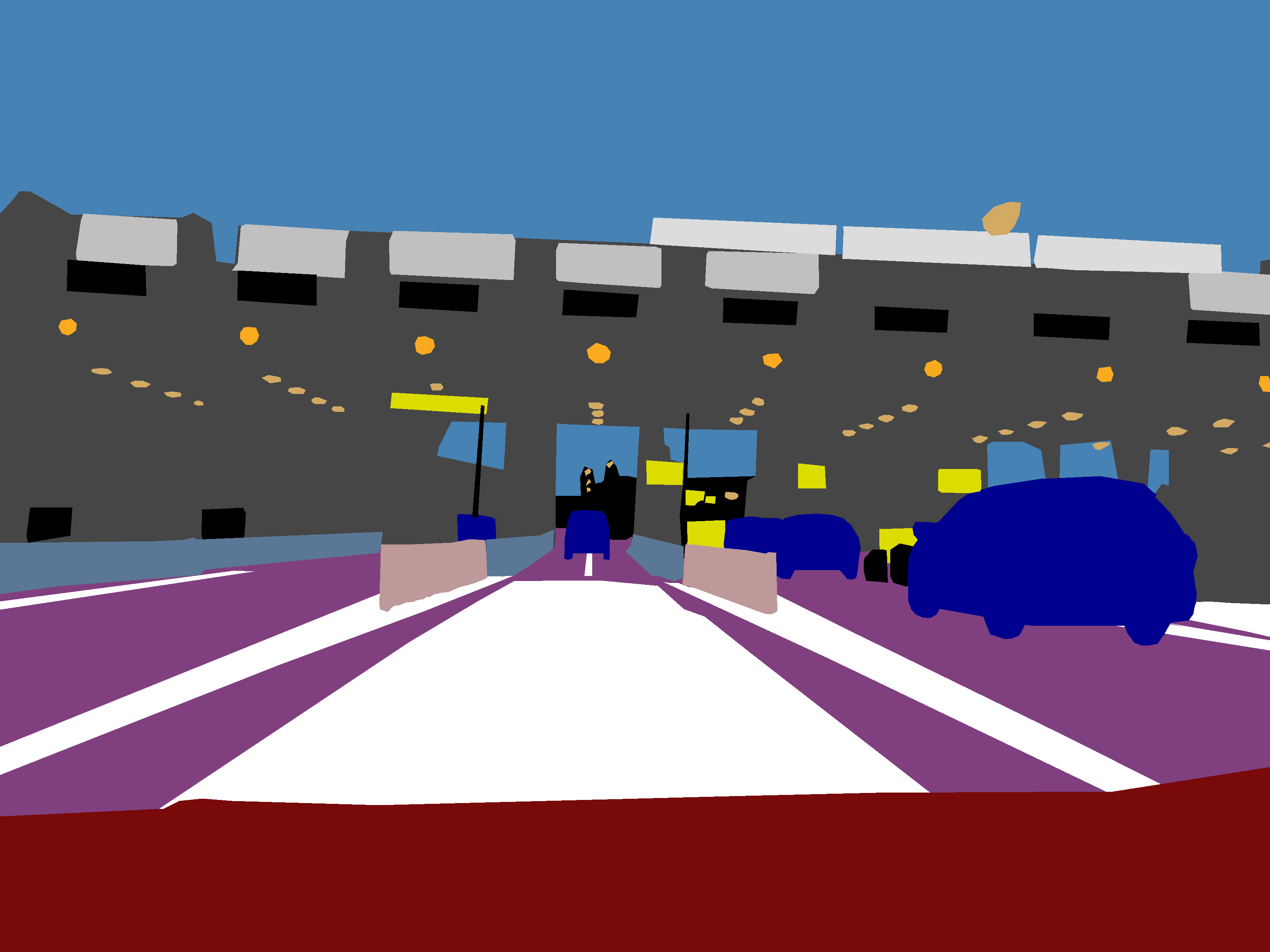}
  \end{subfigure}
   \begin{subfigure}[t]{0.24\linewidth}
    \includegraphics[width=\linewidth]{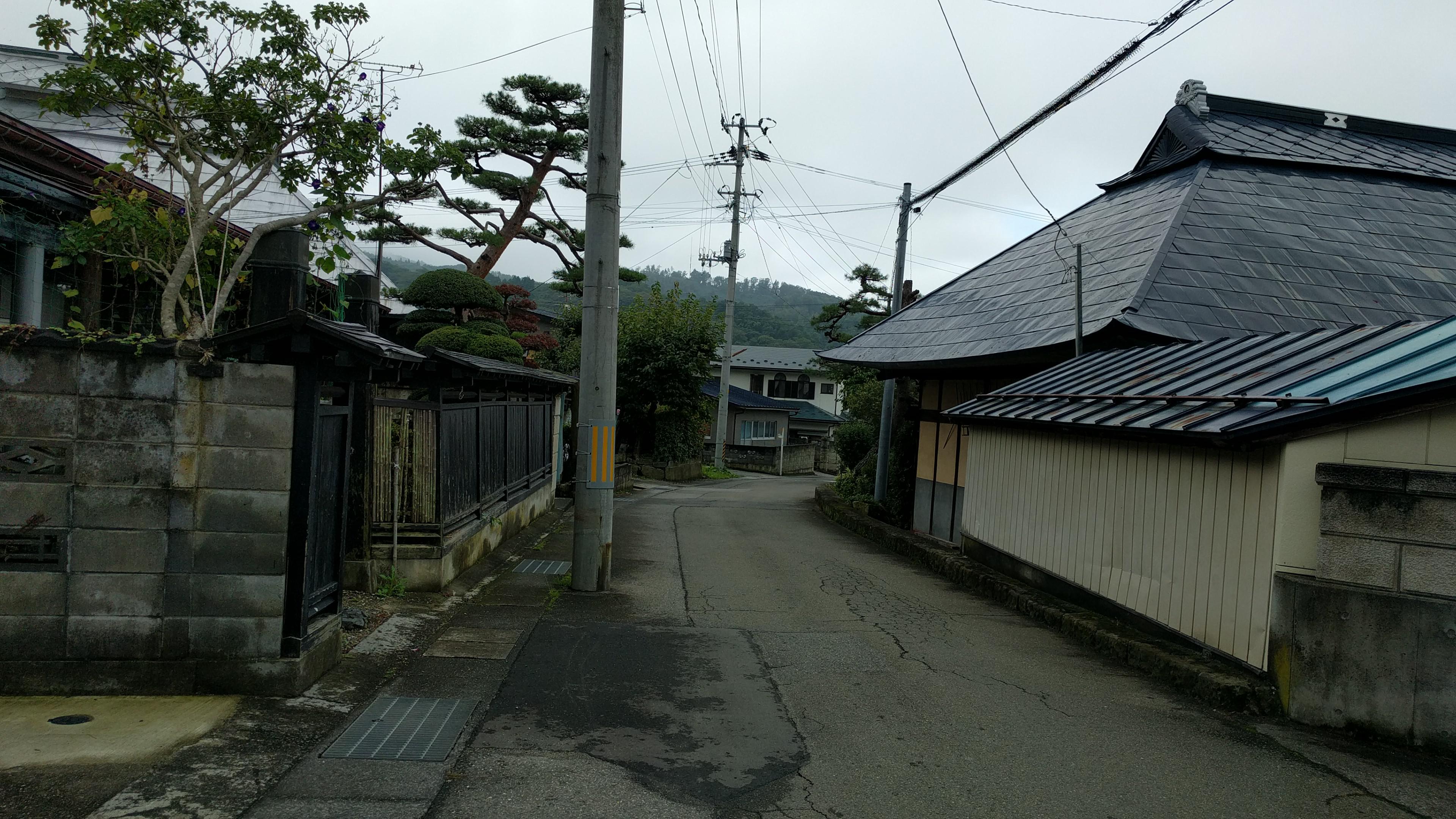}
  \end{subfigure}
  \begin{subfigure}[t]{0.24\linewidth}
    \includegraphics[width=\linewidth]{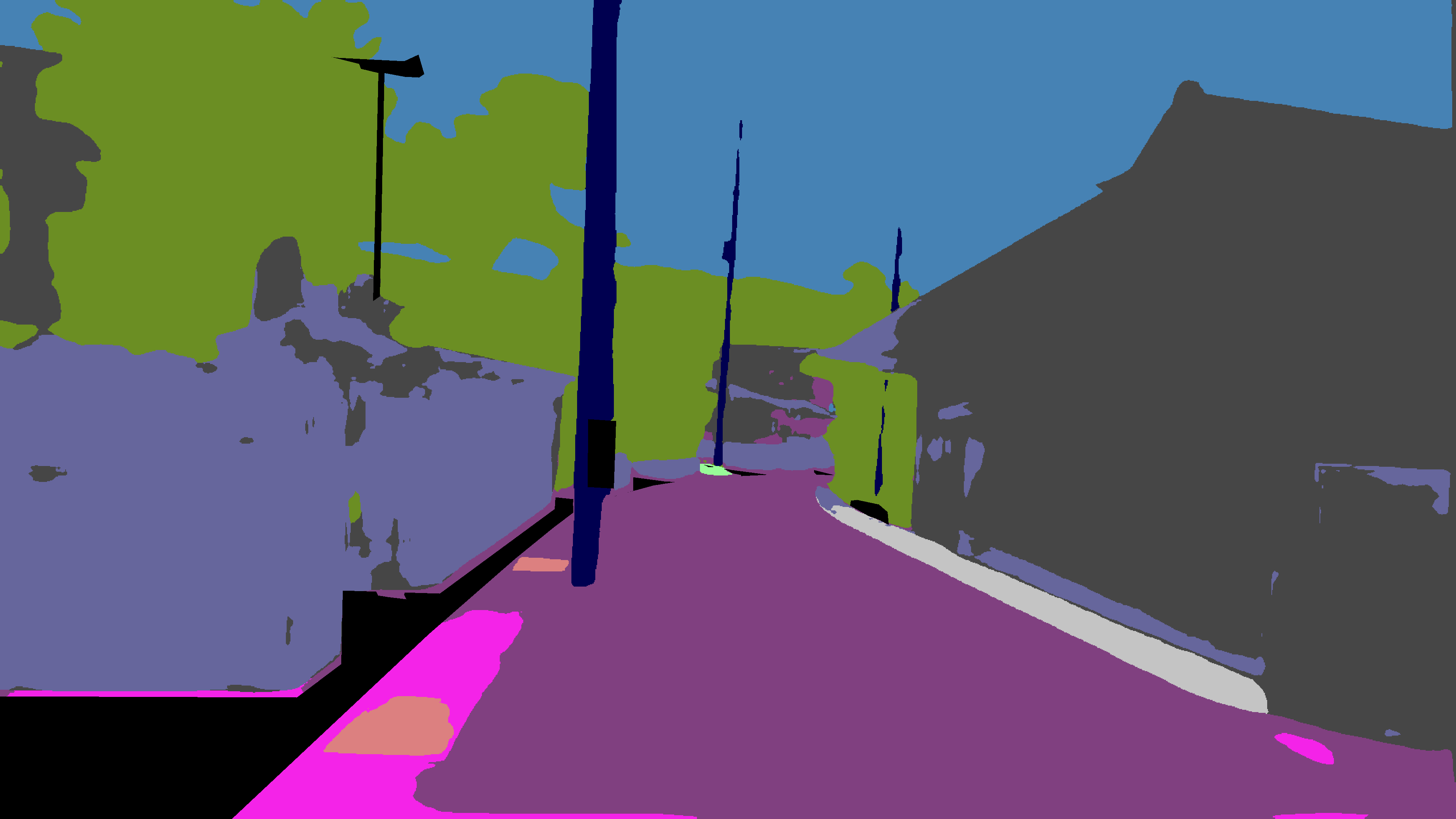}
  \end{subfigure}
  \begin{subfigure}[t]{0.24\linewidth}
    \includegraphics[width=\linewidth]{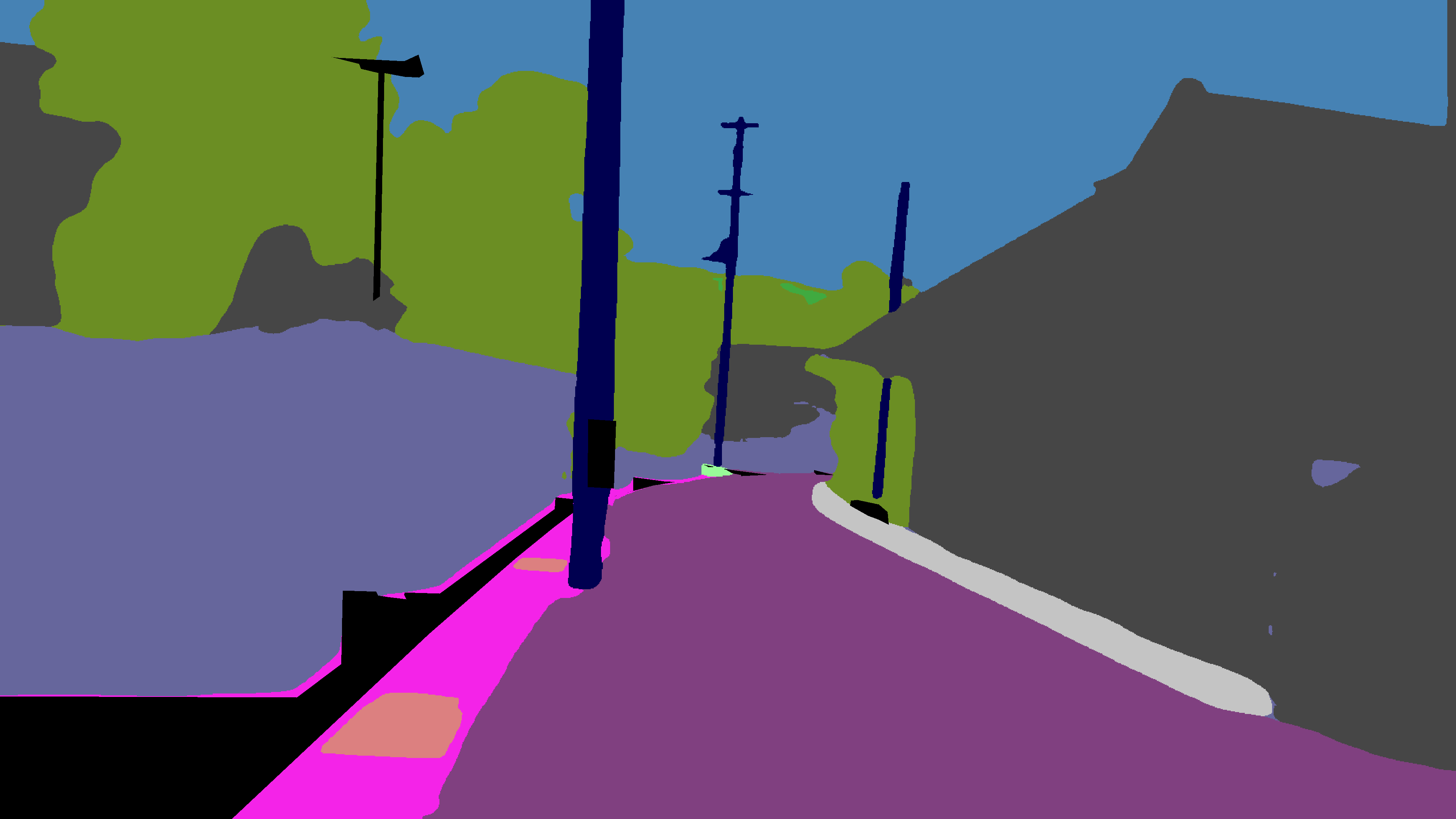}
  \end{subfigure}
  \vspace{1mm}
  \begin{subfigure}[t]{0.24\linewidth}
    \includegraphics[width=\linewidth]{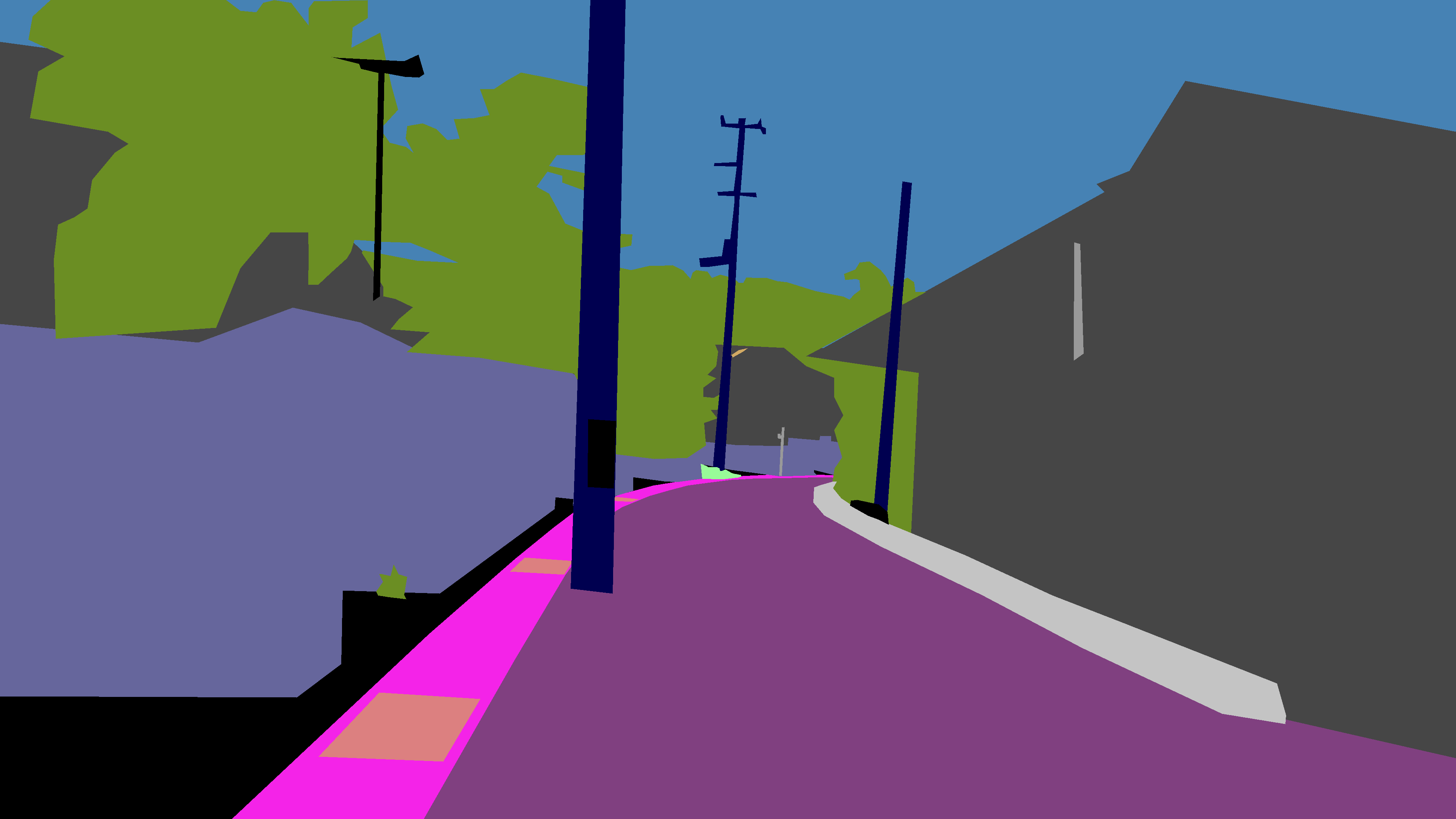}
  \end{subfigure}
   \begin{subfigure}[t]{0.24\linewidth}
    \includegraphics[width=\linewidth]{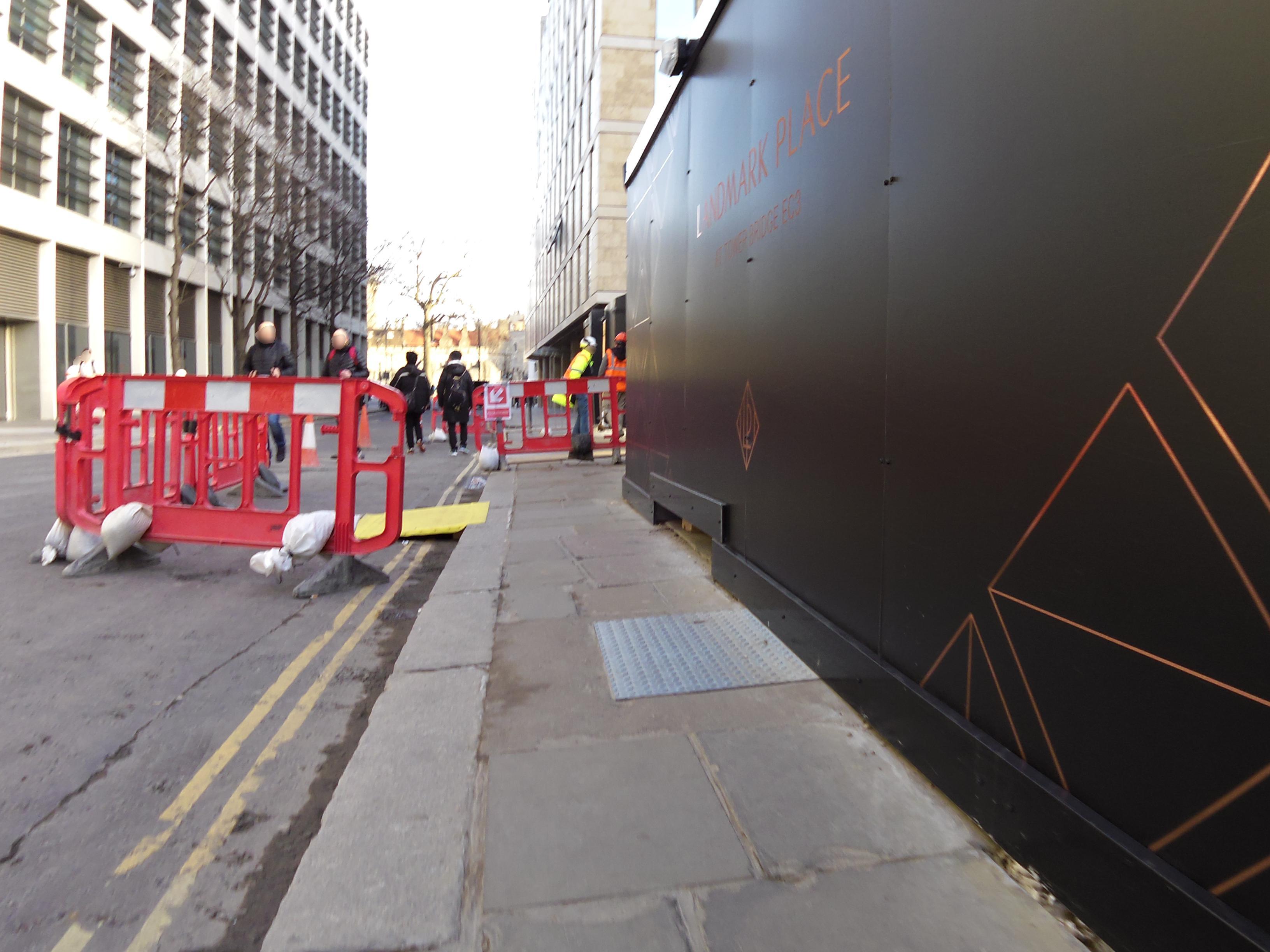}
    \caption*{Input image}
  \end{subfigure}
  \begin{subfigure}[t]{0.24\linewidth}
    \includegraphics[width=\linewidth]{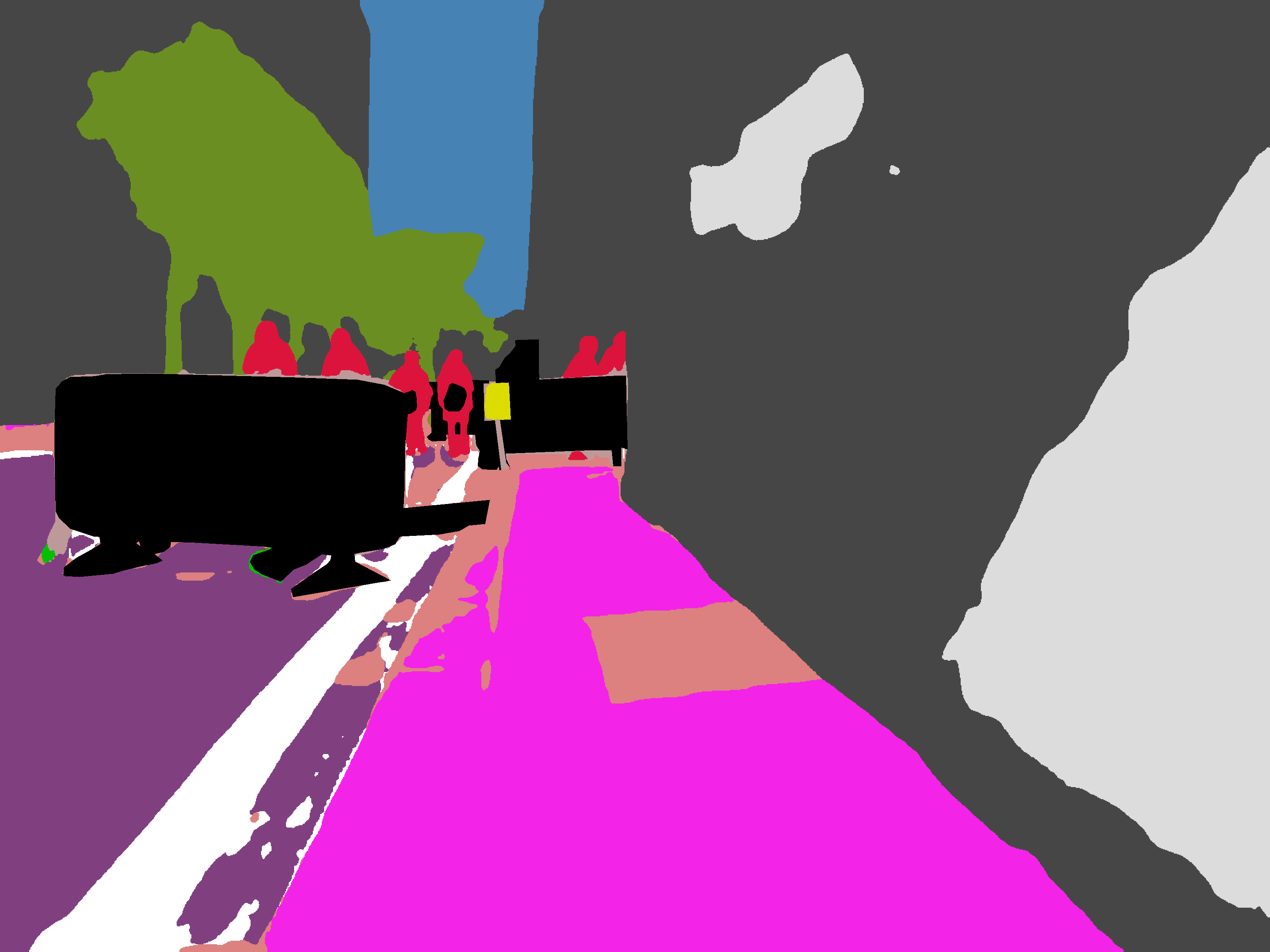}
    \caption*{G-BRS-sb}
  \end{subfigure}
  \begin{subfigure}[t]{0.24\linewidth}
    \includegraphics[width=\linewidth]{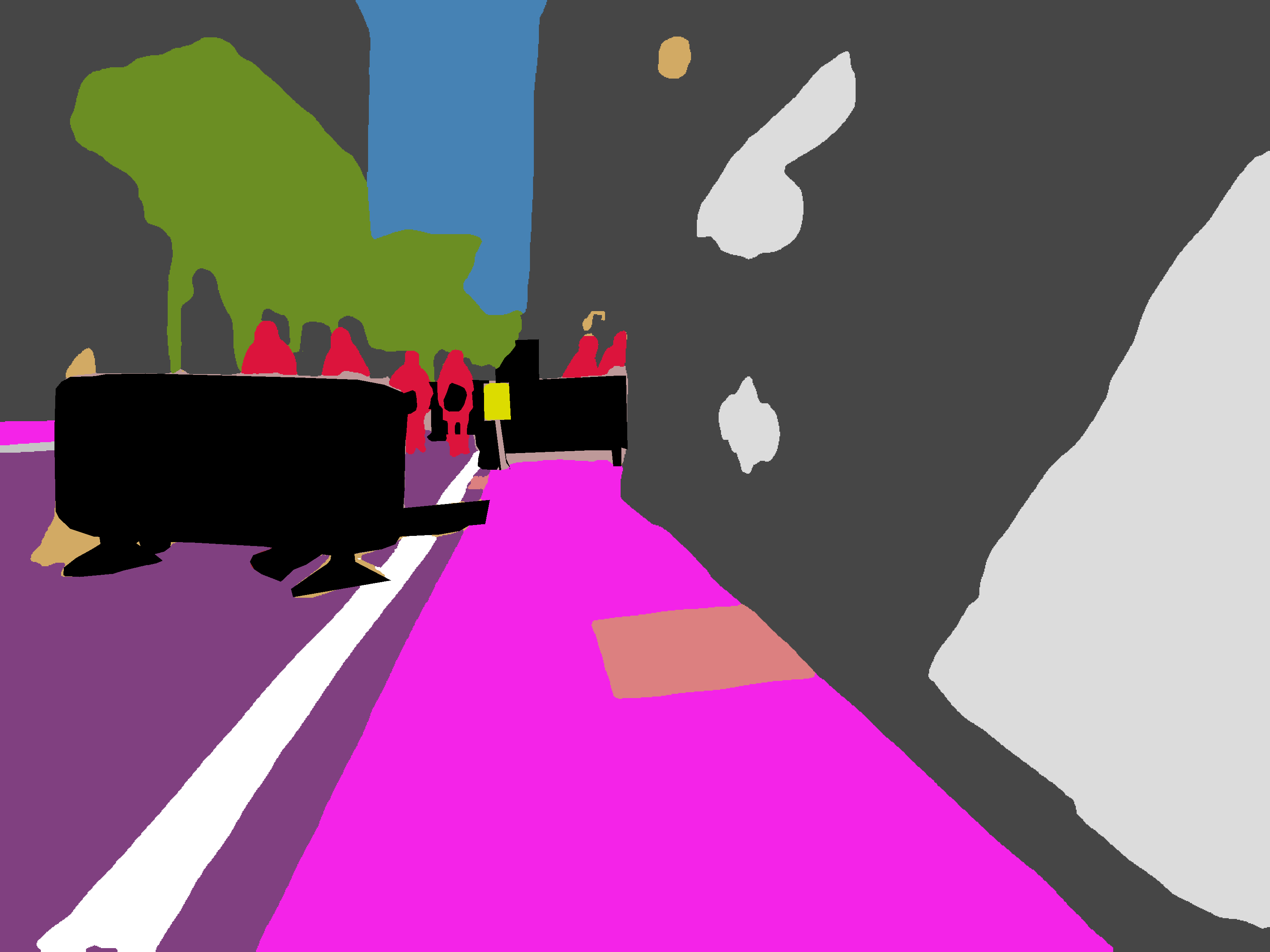}
    \caption*{G-BRS-bmconv}
  \end{subfigure}
  \begin{subfigure}[t]{0.24\linewidth}
    \includegraphics[width=\linewidth]{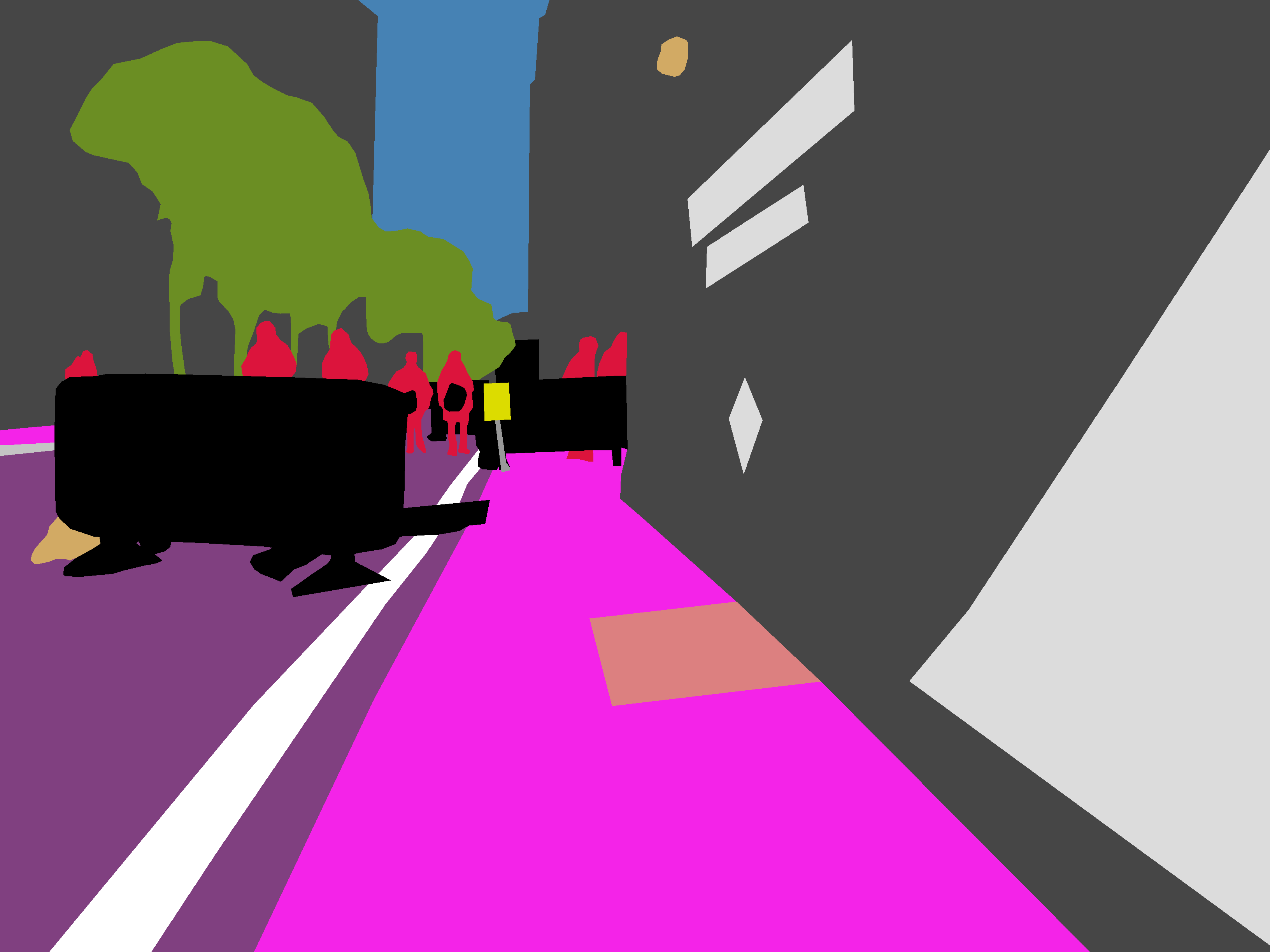}
    \caption*{Ground truth}
  \end{subfigure}
  \caption{Qualitative comparison between performance of the G-BRS-sb layer and our G-BRS-bmconv layer for the task of semantic segmentation on \textbf{Mapillary Vista}. Invalid labels are shown in black.}
  \label{fig:qualitative_compare_mapillary}
  \vspace{-4mm}
\end{figure*}

\begin{figure*}[h]
  \centering
   \begin{subfigure}[t]{0.24\linewidth}
    \includegraphics[width=\linewidth]{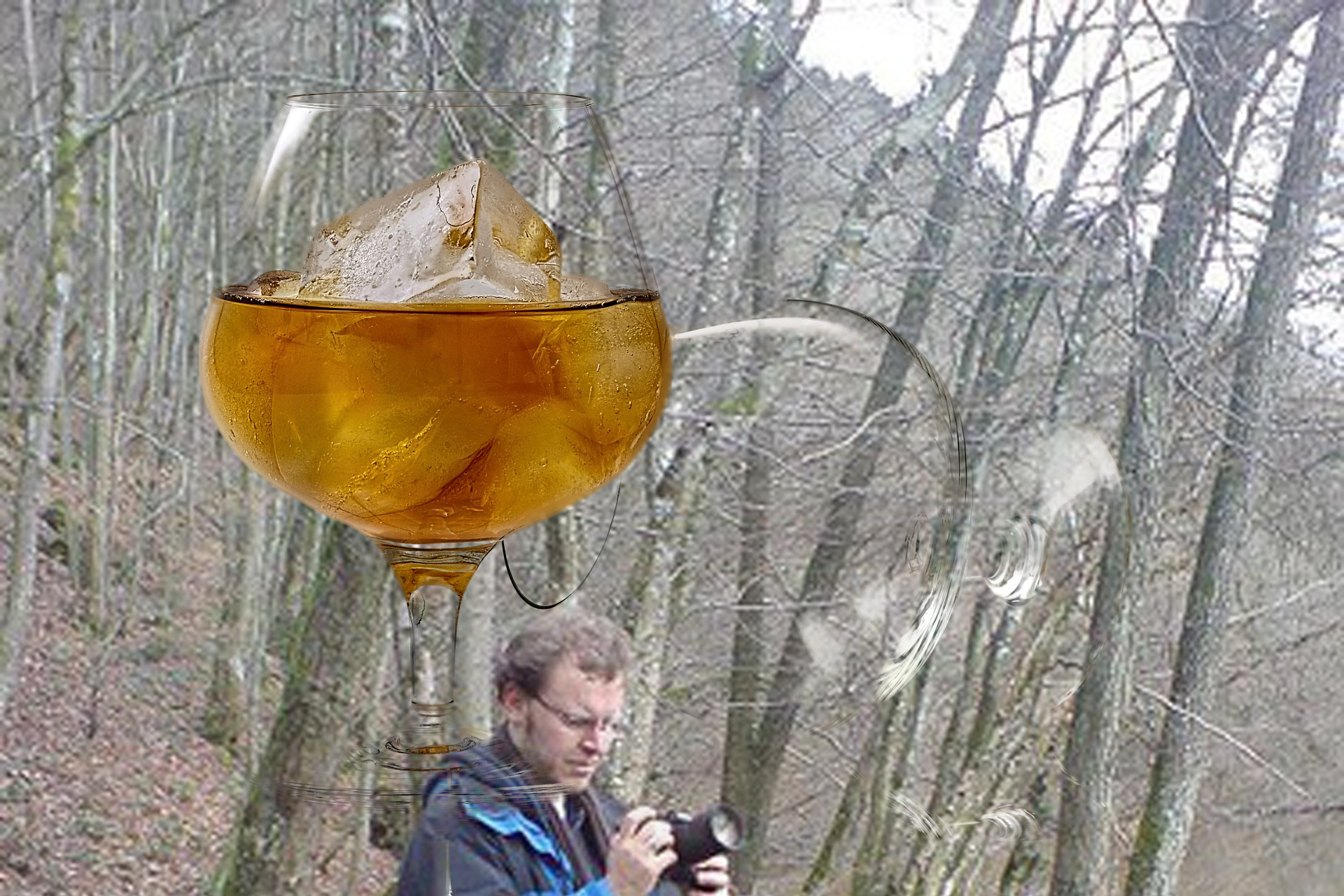}
  \end{subfigure}
  \begin{subfigure}[t]{0.24\linewidth}
    \includegraphics[width=\linewidth]{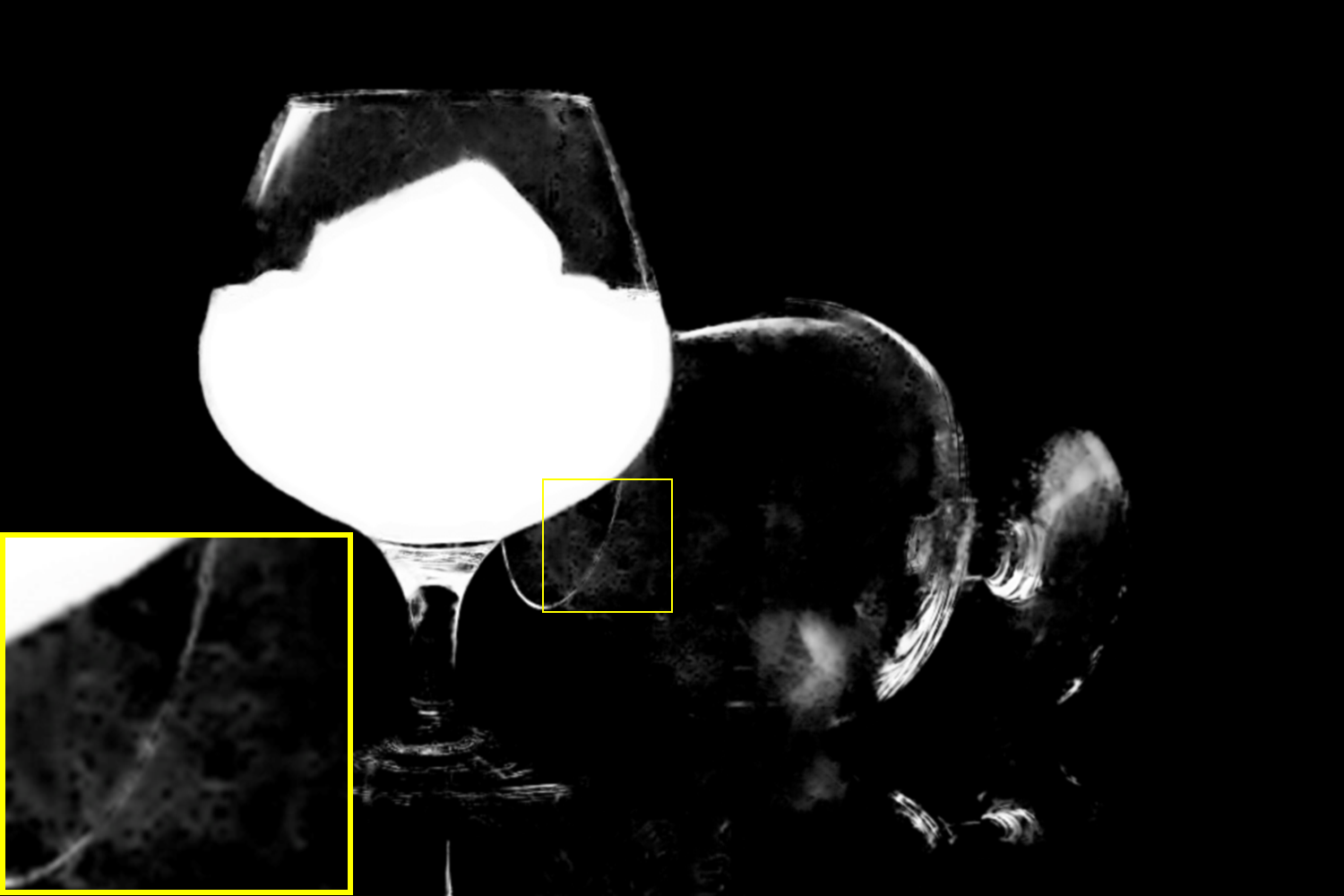}
  \end{subfigure}
  \begin{subfigure}[t]{0.24\linewidth}
    \includegraphics[width=\linewidth]{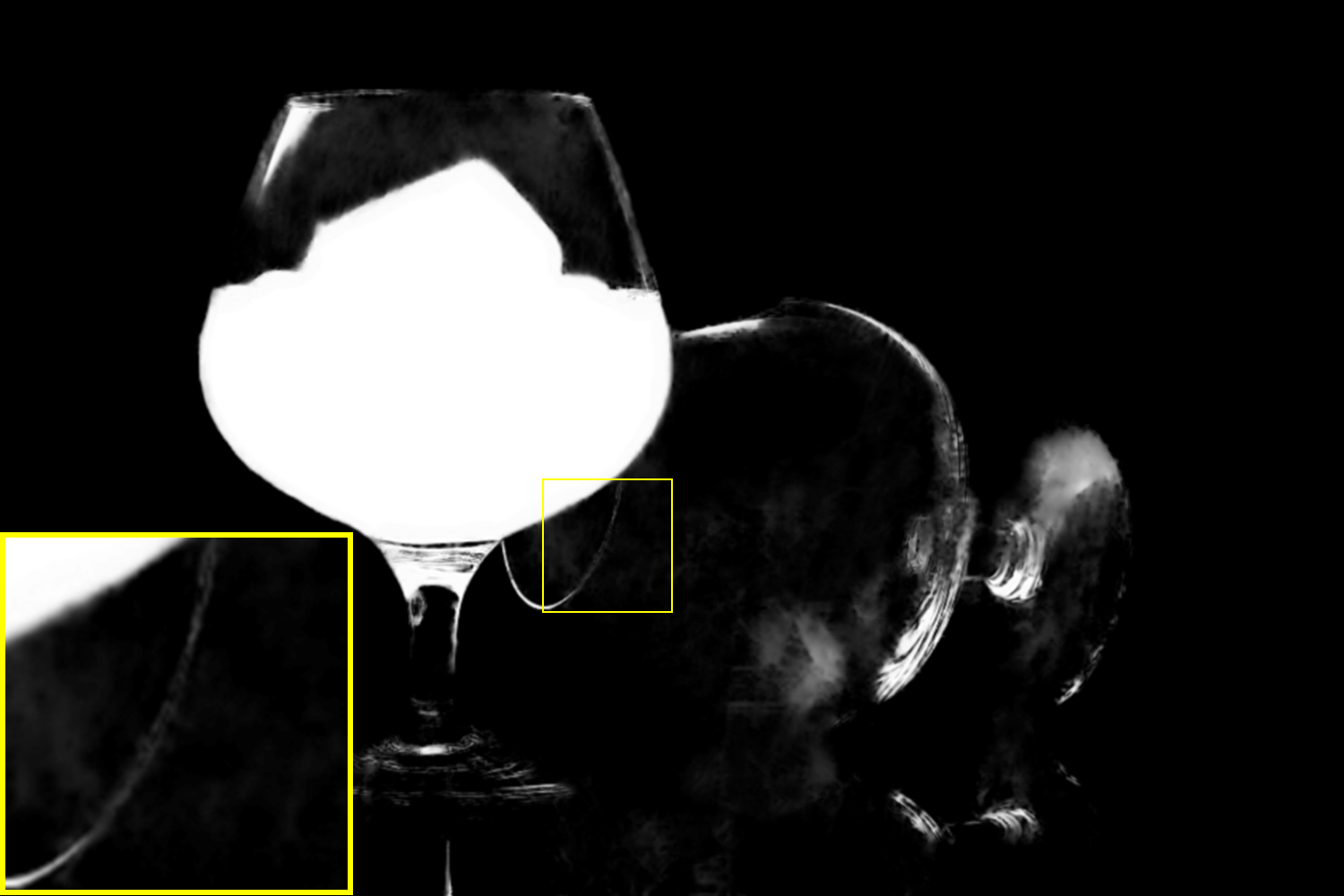}
  \end{subfigure}
  \vspace{1mm}
  \begin{subfigure}[t]{0.24\linewidth}
    \includegraphics[width=\linewidth]{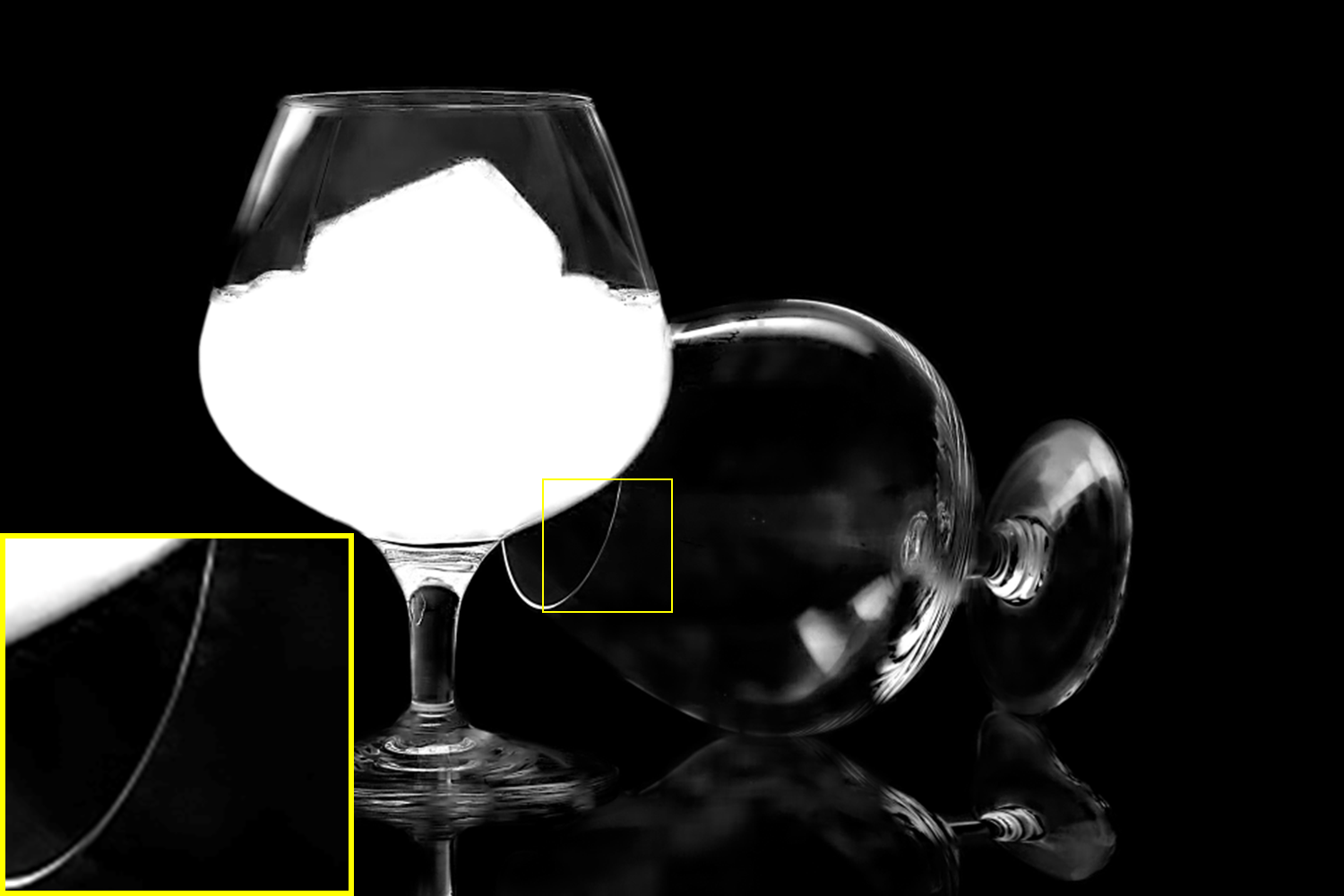}
  \end{subfigure}
   \begin{subfigure}[t]{0.24\linewidth}
    \includegraphics[width=\linewidth]{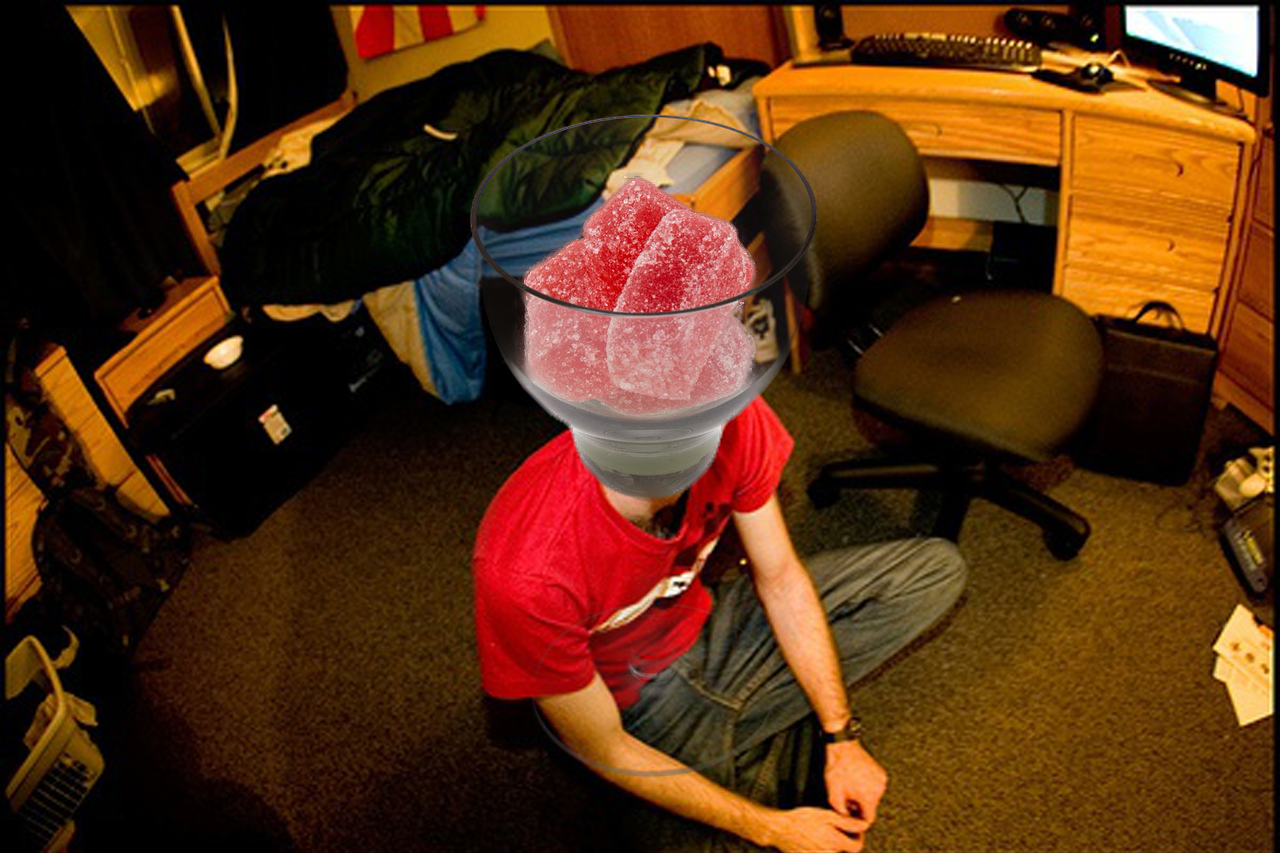}
  \end{subfigure}
  \begin{subfigure}[t]{0.24\linewidth}
    \includegraphics[width=\linewidth]{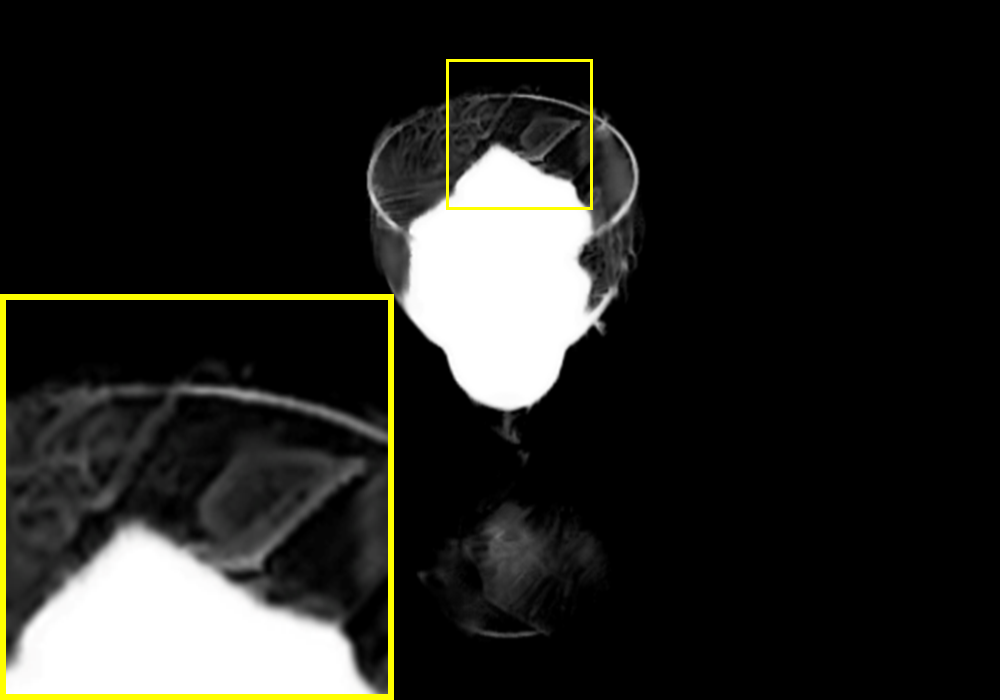}
  \end{subfigure}
  \begin{subfigure}[t]{0.24\linewidth}
    \includegraphics[width=\linewidth]{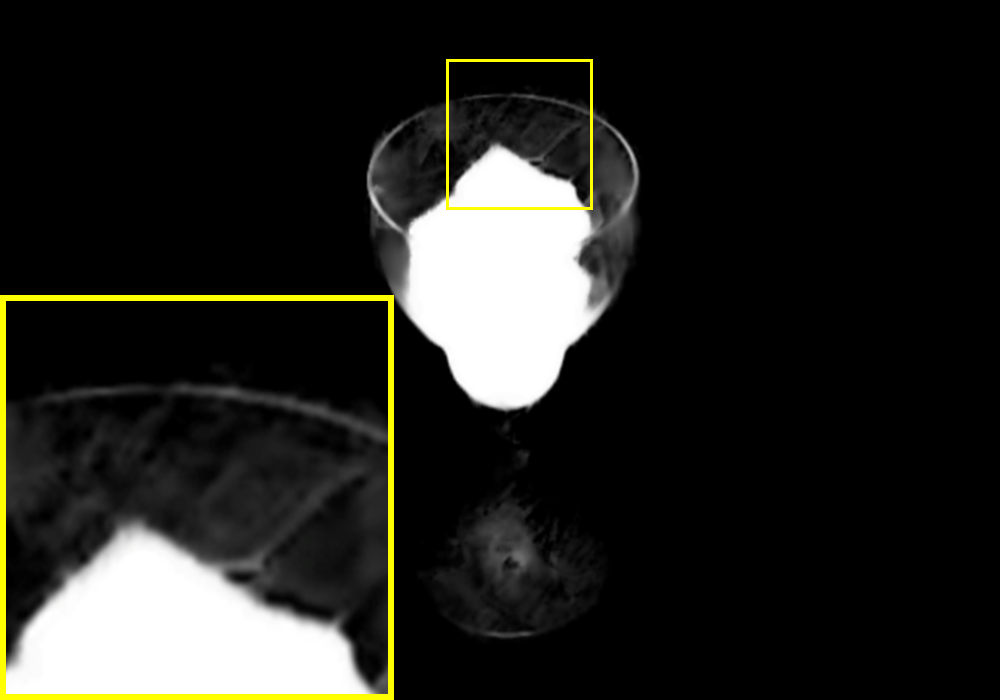}
  \end{subfigure}
  \vspace{1mm}
  \begin{subfigure}[t]{0.24\linewidth}
    \includegraphics[width=\linewidth]{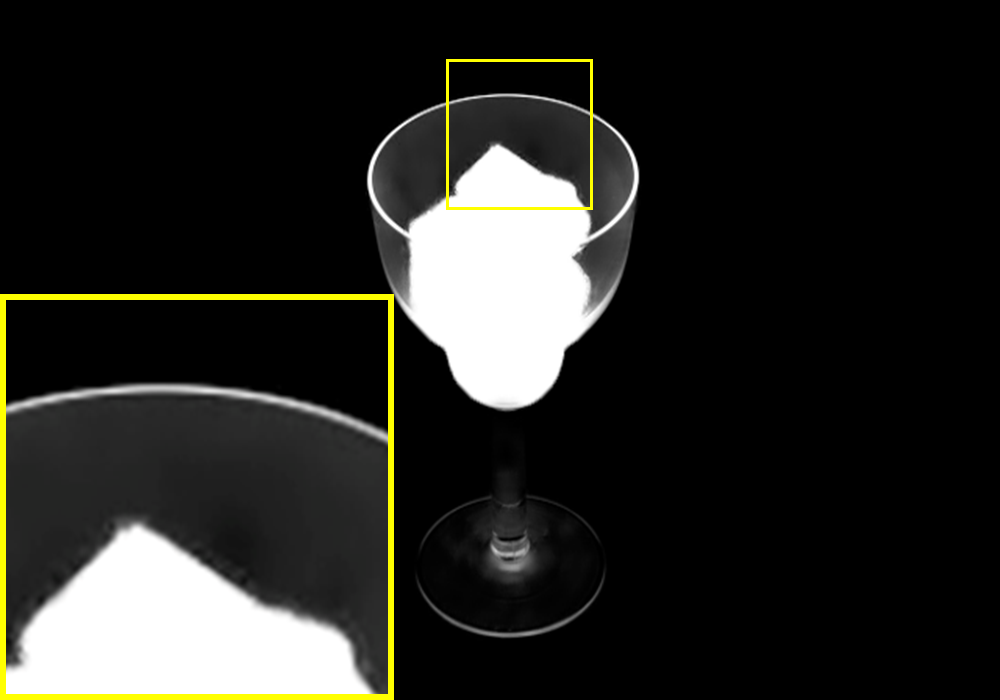}
  \end{subfigure}
   \begin{subfigure}[t]{0.24\linewidth}
    \includegraphics[width=\linewidth]{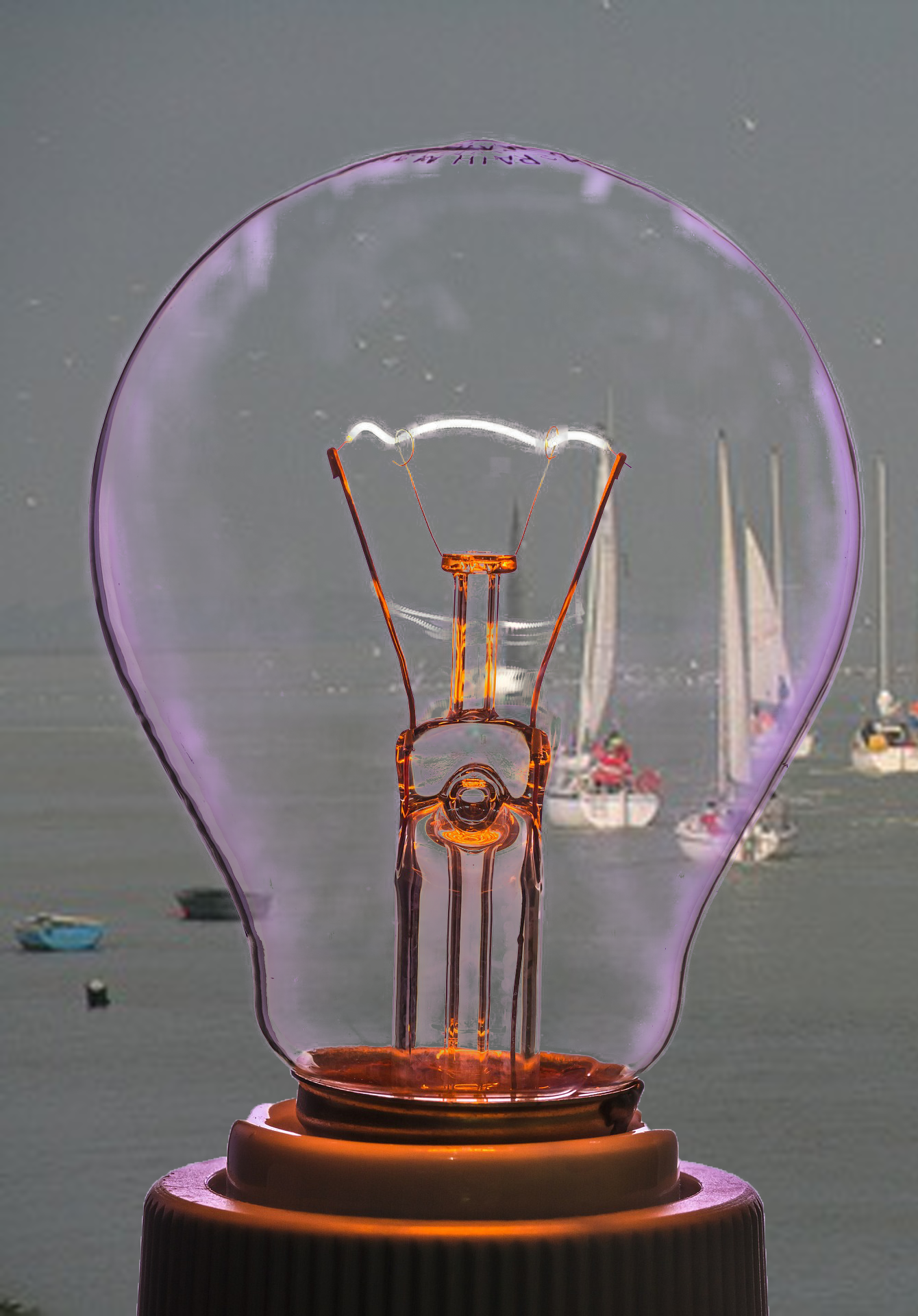}
  \end{subfigure}
  \begin{subfigure}[t]{0.24\linewidth}
    \includegraphics[width=\linewidth]{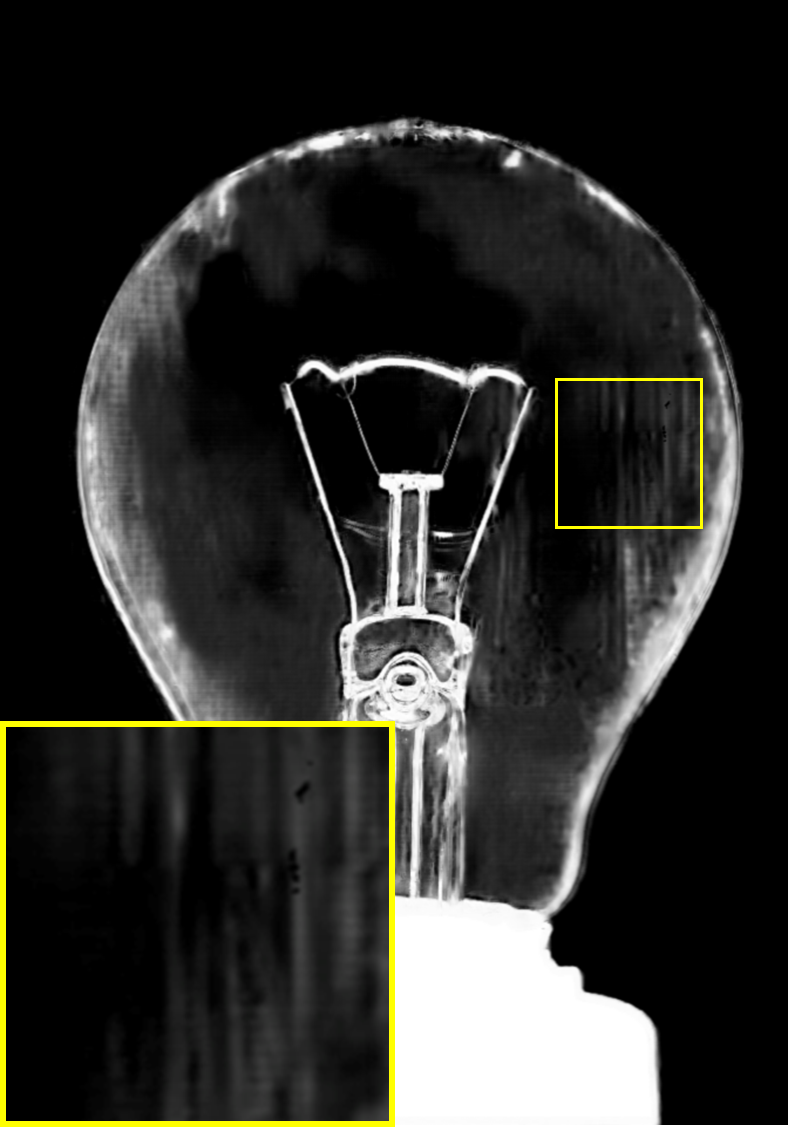}
  \end{subfigure}
  \begin{subfigure}[t]{0.24\linewidth}
    \includegraphics[width=\linewidth]{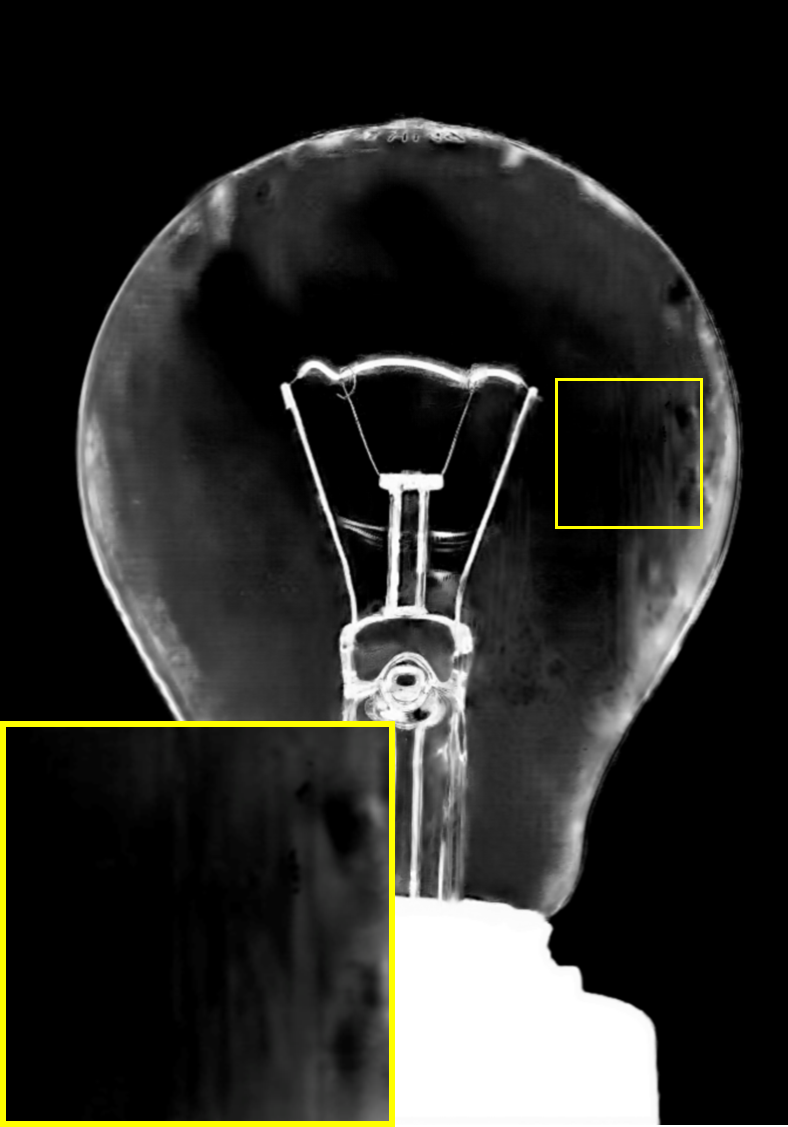}
  \end{subfigure}
  \vspace{1mm}
  \begin{subfigure}[t]{0.24\linewidth}
    \includegraphics[width=\linewidth]{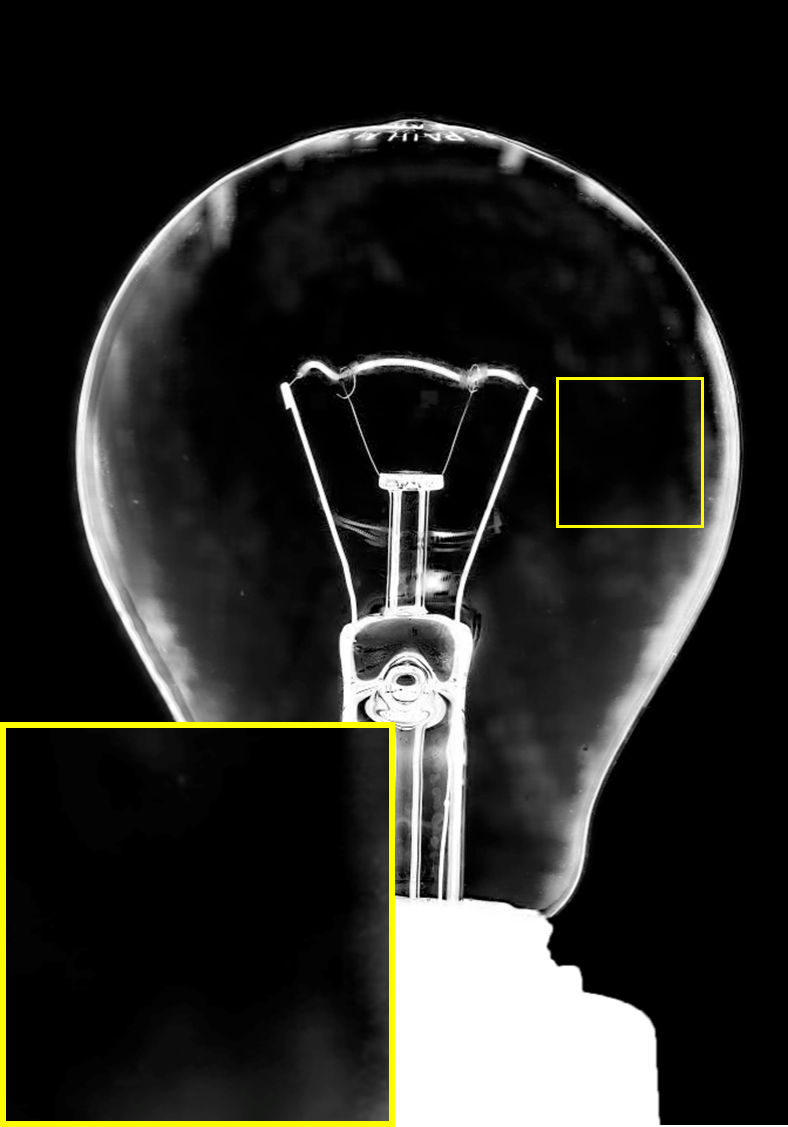}
  \end{subfigure}
   \begin{subfigure}[t]{0.24\linewidth}
    \includegraphics[width=\linewidth]{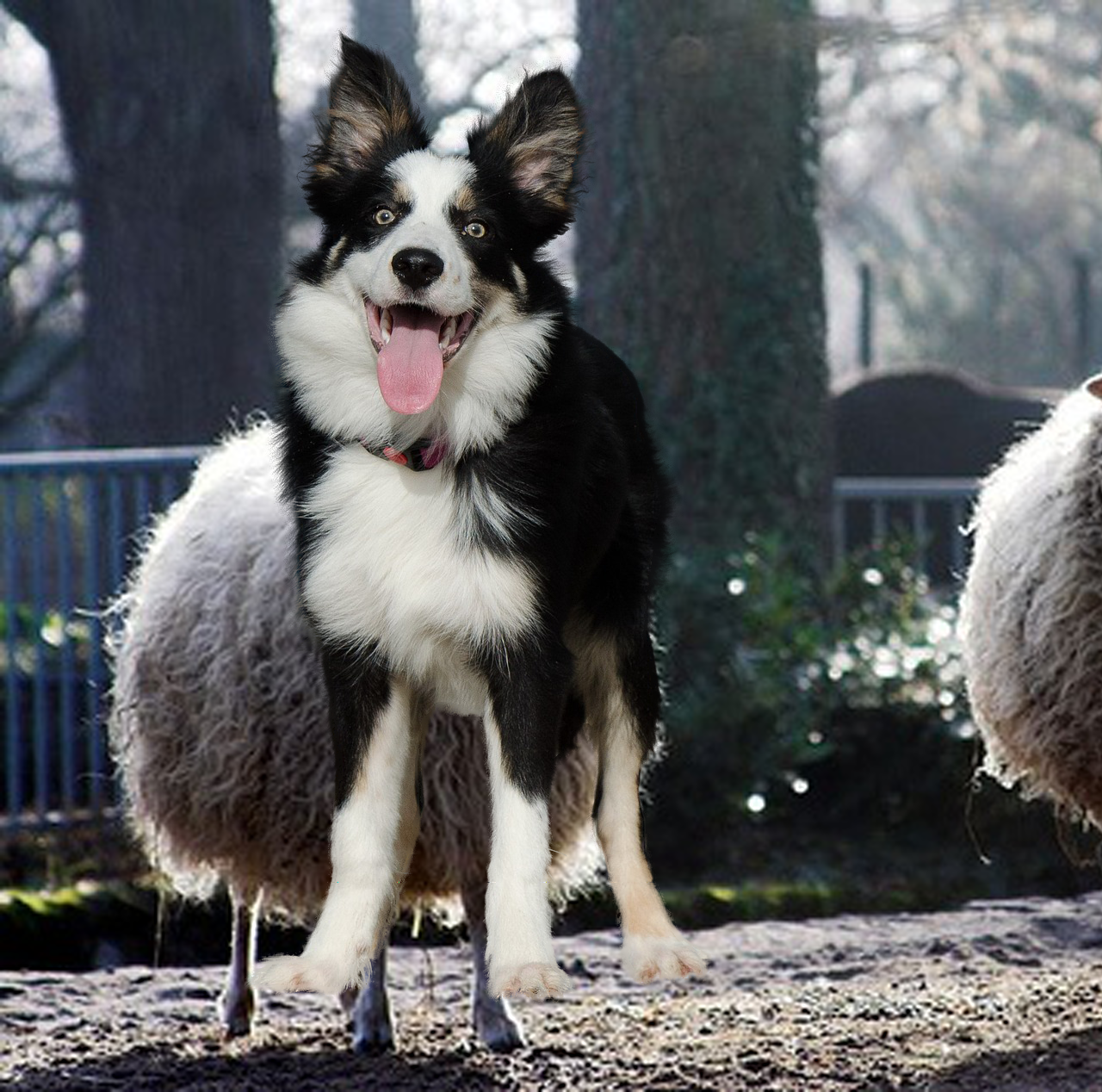}
    \caption*{Input image}
  \end{subfigure}
  \begin{subfigure}[t]{0.24\linewidth}
    \includegraphics[width=\linewidth]{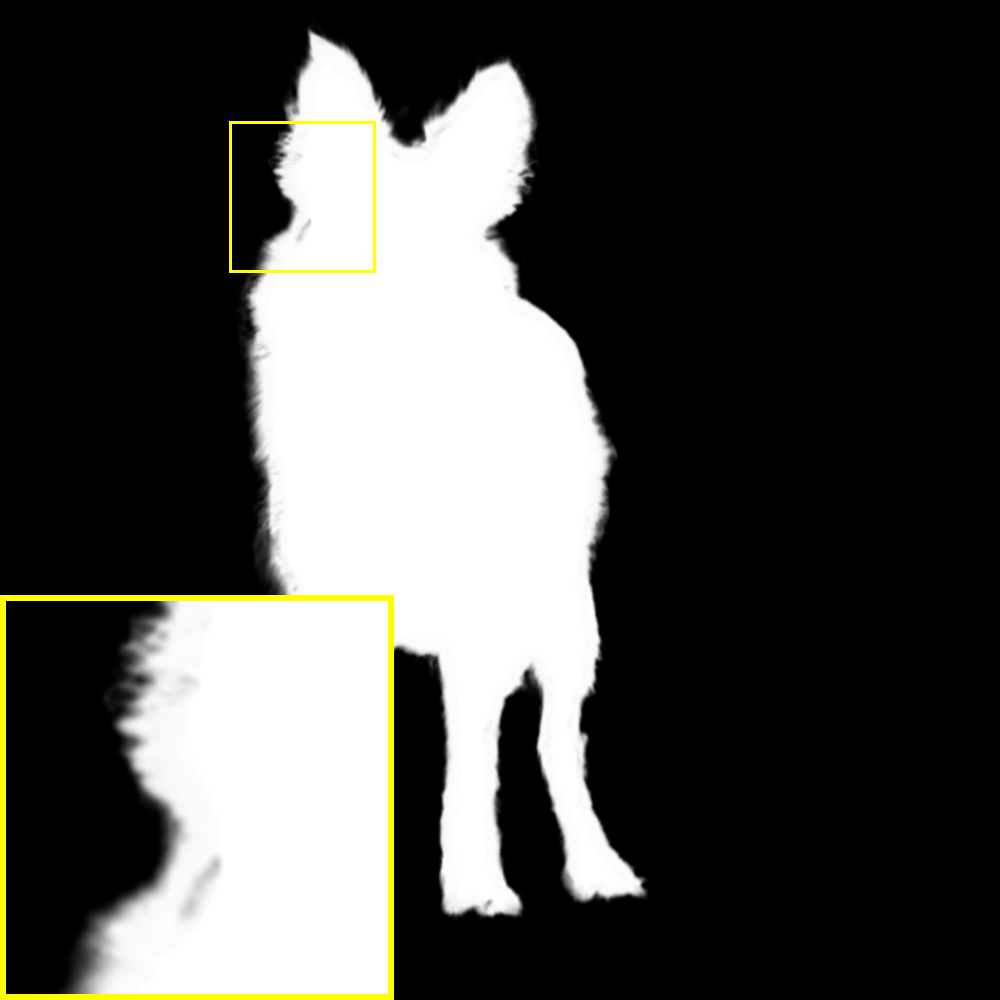}
    \caption*{G-BRS-sb}
  \end{subfigure}
  \begin{subfigure}[t]{0.24\linewidth}
    \includegraphics[width=\linewidth]{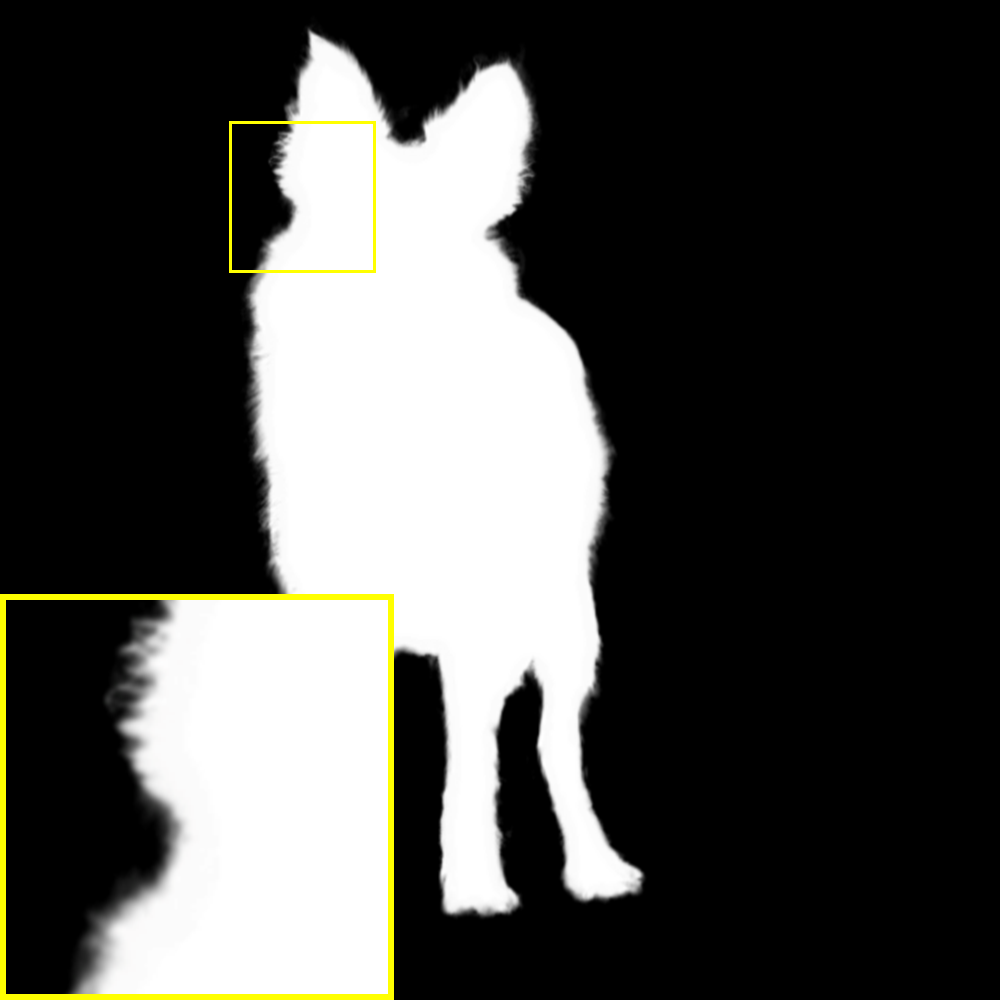}
    \caption*{G-BRS-bmconv}
  \end{subfigure}
  \begin{subfigure}[t]{0.24\linewidth}
    \includegraphics[width=\linewidth]{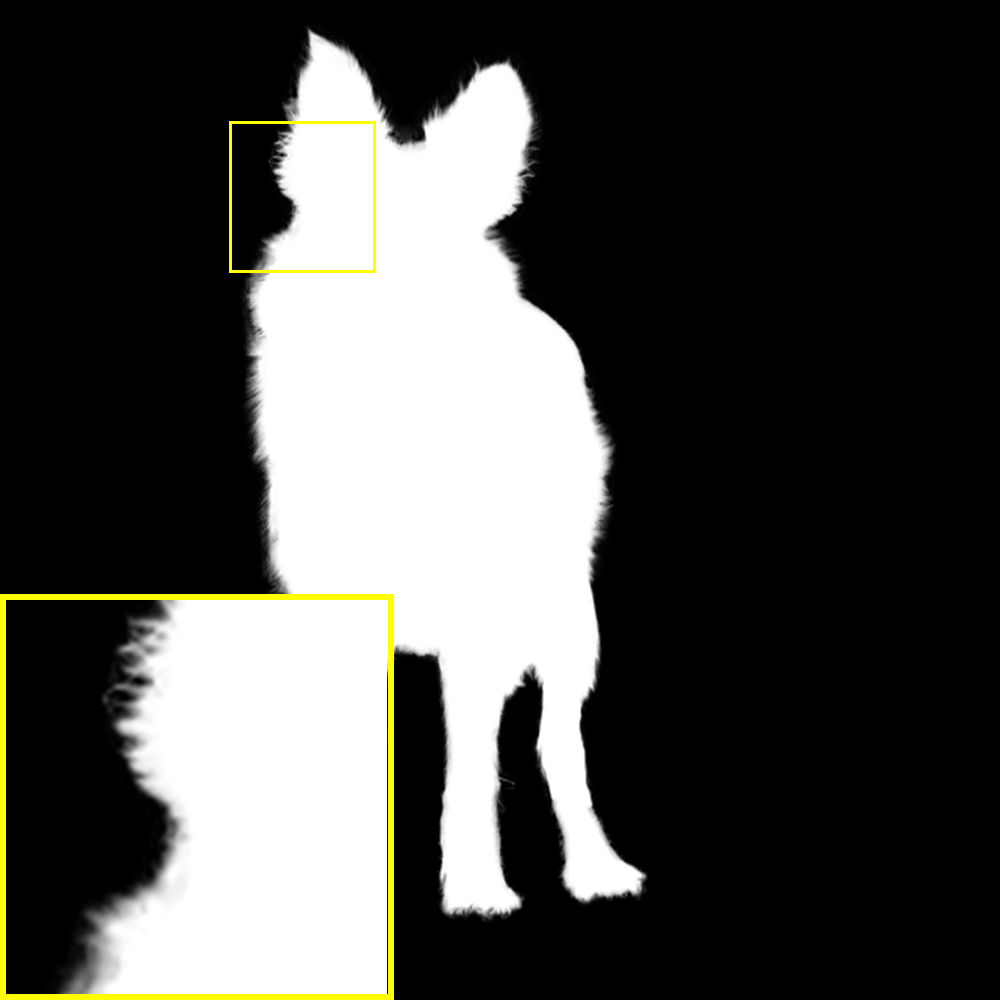}
    \caption*{Ground truth}
  \end{subfigure}
  \caption{Qualitative comparison between performance of the G-BRS-sb layer and our G-BRS-bmconv layer for the task of image matting on \textbf{Composition-1k}. }
  \label{fig:qualitative_compare_comp1k}
  \vspace{-4mm}
\end{figure*}

\begin{figure*}[h]
  \centering
   \begin{subfigure}[t]{0.24\linewidth}
    \includegraphics[width=\linewidth]{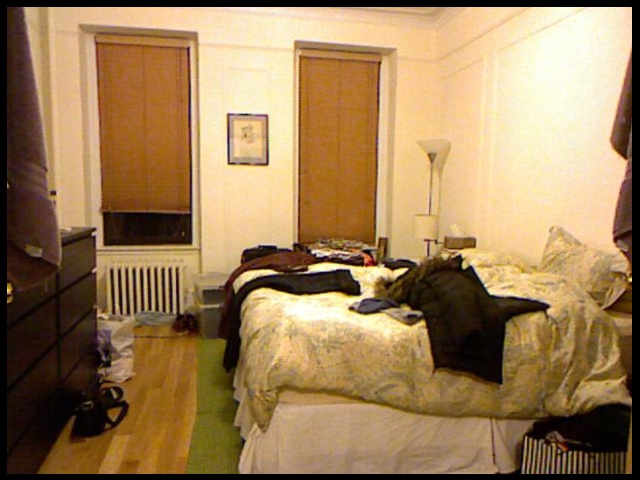}
  \end{subfigure}
  \begin{subfigure}[t]{0.24\linewidth}
    \includegraphics[width=\linewidth]{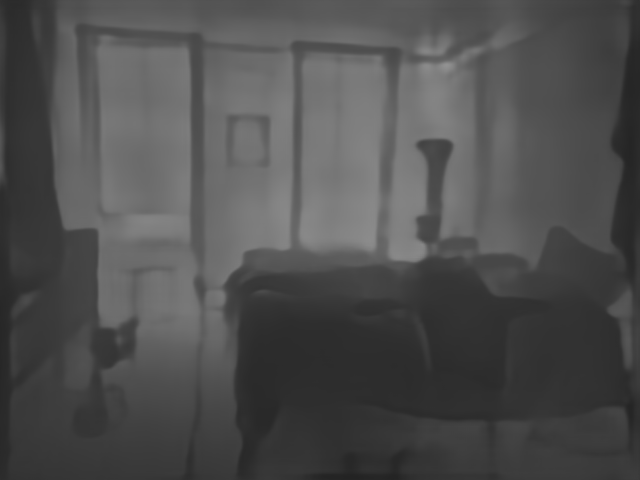}
  \end{subfigure}
  \begin{subfigure}[t]{0.24\linewidth}
    \includegraphics[width=\linewidth]{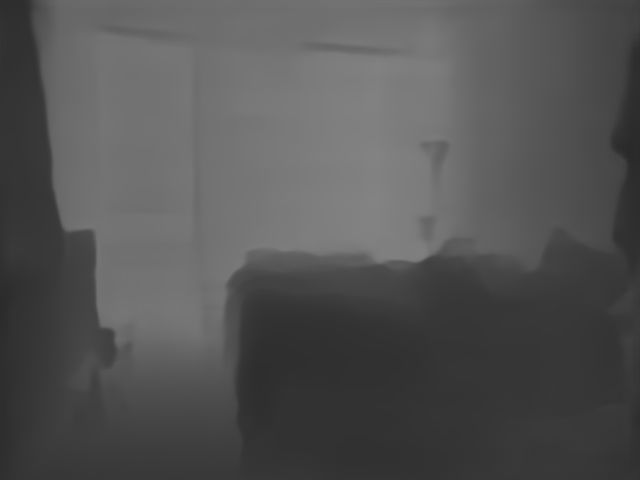}
  \end{subfigure}
  \vspace{1mm}
  \begin{subfigure}[t]{0.24\linewidth}
    \includegraphics[width=\linewidth]{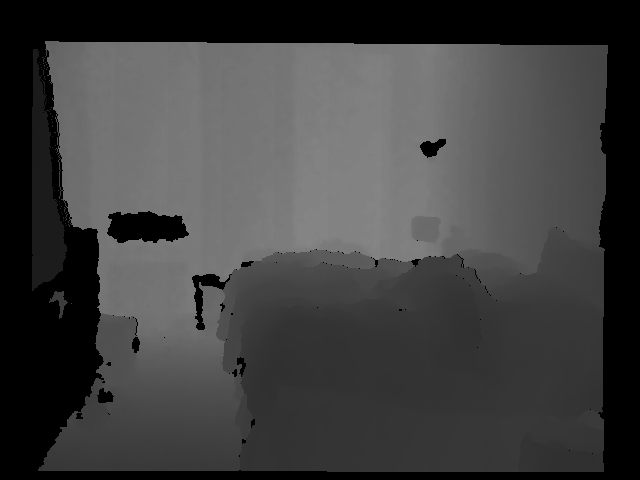}
  \end{subfigure}
   \begin{subfigure}[t]{0.24\linewidth}
    \includegraphics[width=\linewidth]{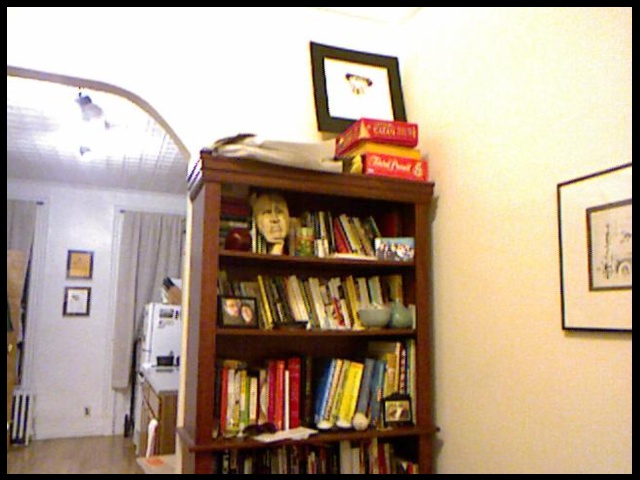}
  \end{subfigure}
  \begin{subfigure}[t]{0.24\linewidth}
    \includegraphics[width=\linewidth]{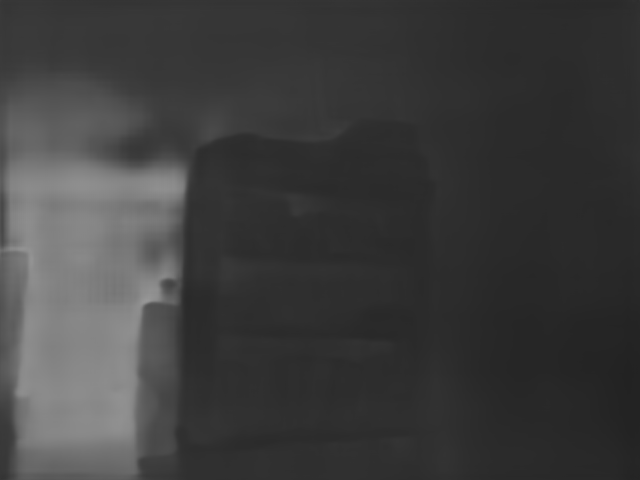}
  \end{subfigure}
  \begin{subfigure}[t]{0.24\linewidth}
    \includegraphics[width=\linewidth]{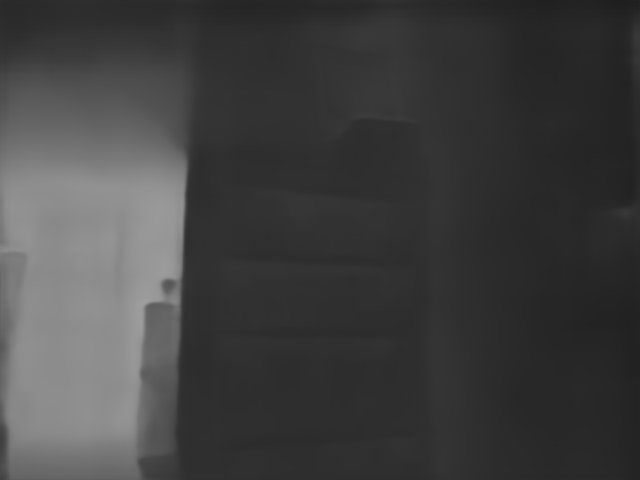}
  \end{subfigure}
  \vspace{1mm}
  \begin{subfigure}[t]{0.24\linewidth}
    \includegraphics[width=\linewidth]{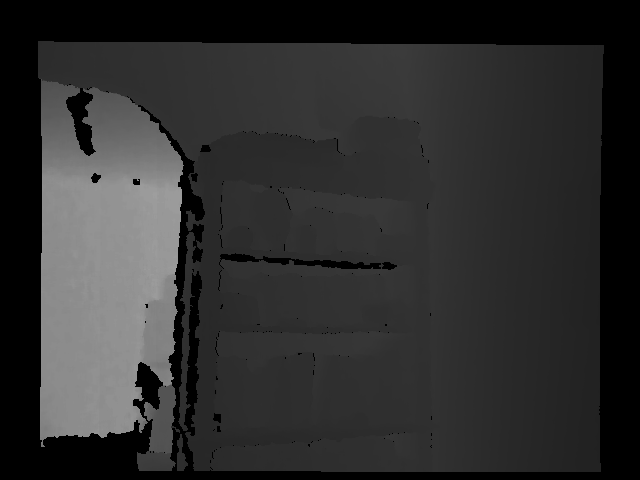}
  \end{subfigure}
   \begin{subfigure}[t]{0.24\linewidth}
    \includegraphics[width=\linewidth]{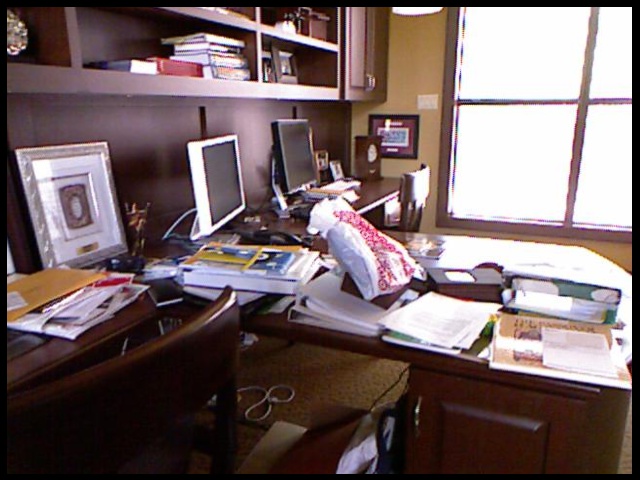}
  \end{subfigure}
  \begin{subfigure}[t]{0.24\linewidth}
    \includegraphics[width=\linewidth]{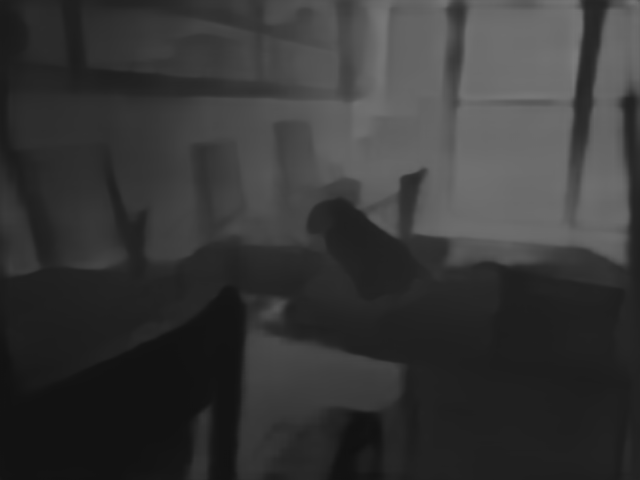}
  \end{subfigure}
  \begin{subfigure}[t]{0.24\linewidth}
    \includegraphics[width=\linewidth]{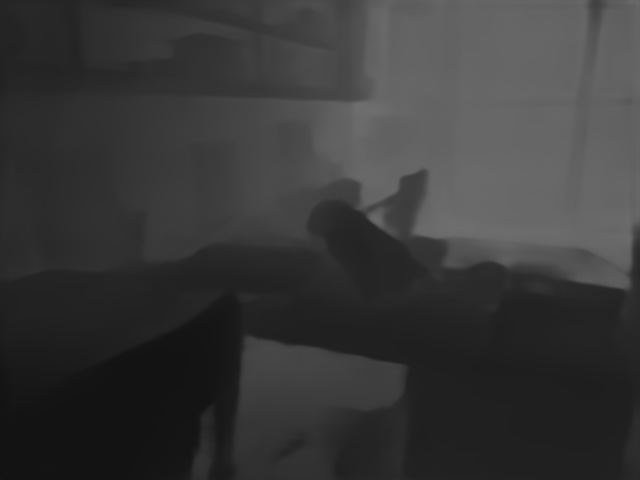}
  \end{subfigure}
  \vspace{1mm}
  \begin{subfigure}[t]{0.24\linewidth}
    \includegraphics[width=\linewidth]{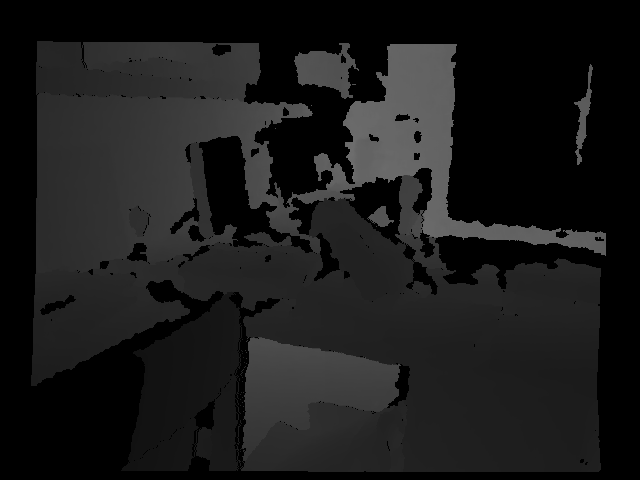}
  \end{subfigure}
   \begin{subfigure}[t]{0.24\linewidth}
    \includegraphics[width=\linewidth]{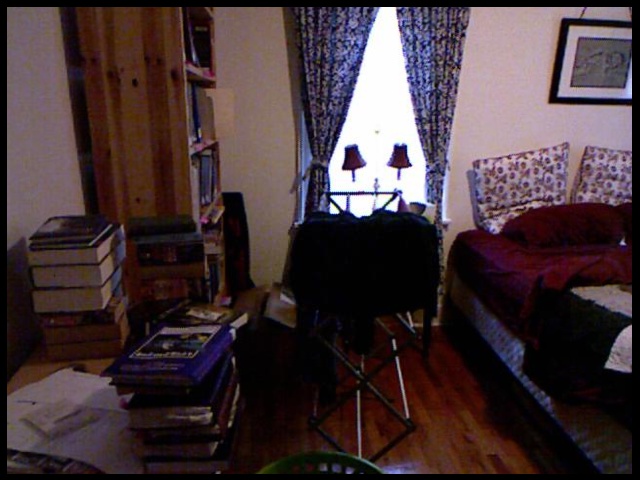}
  \end{subfigure}
  \begin{subfigure}[t]{0.24\linewidth}
    \includegraphics[width=\linewidth]{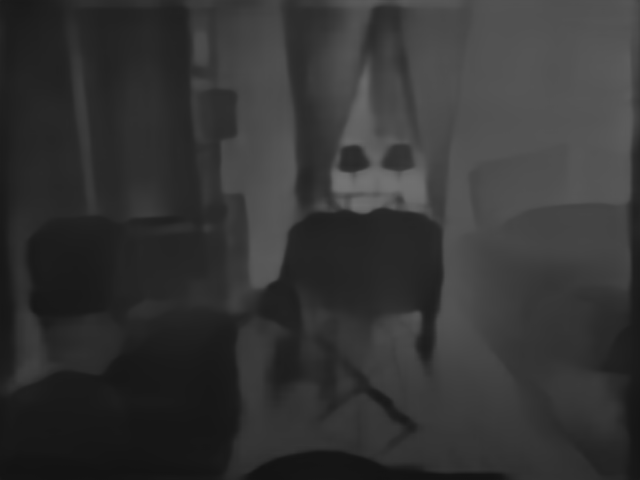}
  \end{subfigure}
  \begin{subfigure}[t]{0.24\linewidth}
    \includegraphics[width=\linewidth]{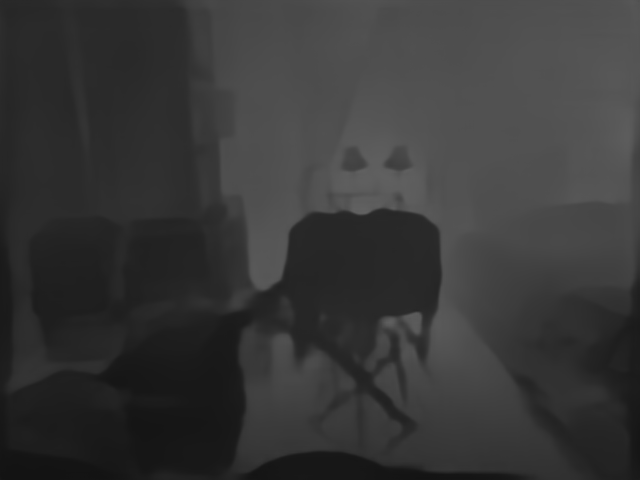}
  \end{subfigure}
  \vspace{1mm}
  \begin{subfigure}[t]{0.24\linewidth}
    \includegraphics[width=\linewidth]{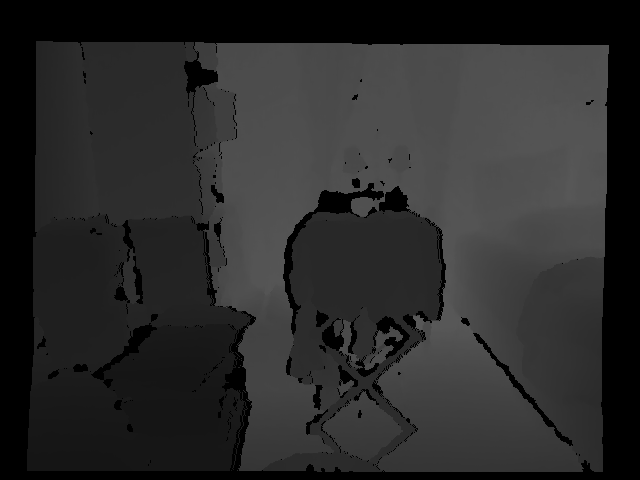}
  \end{subfigure}
   \begin{subfigure}[t]{0.24\linewidth}
    \includegraphics[width=\linewidth]{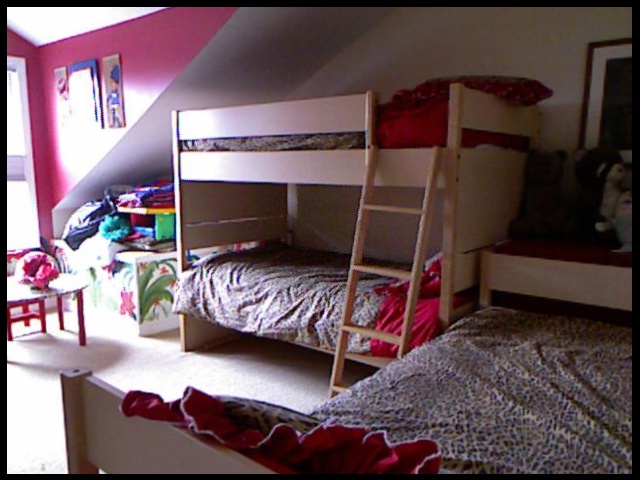}
  \end{subfigure}
  \begin{subfigure}[t]{0.24\linewidth}
    \includegraphics[width=\linewidth]{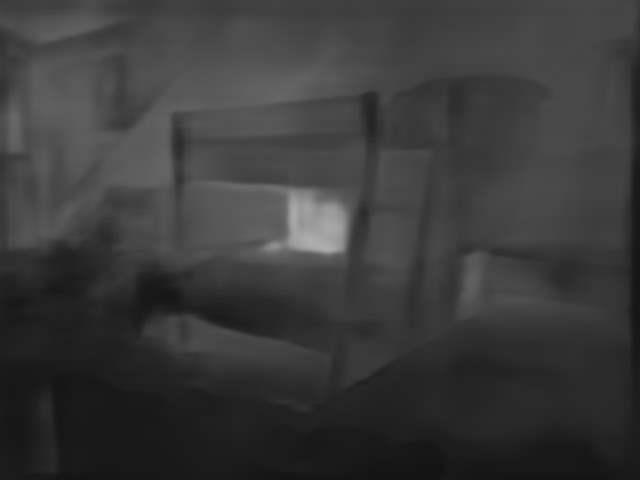}
  \end{subfigure}
  \begin{subfigure}[t]{0.24\linewidth}
    \includegraphics[width=\linewidth]{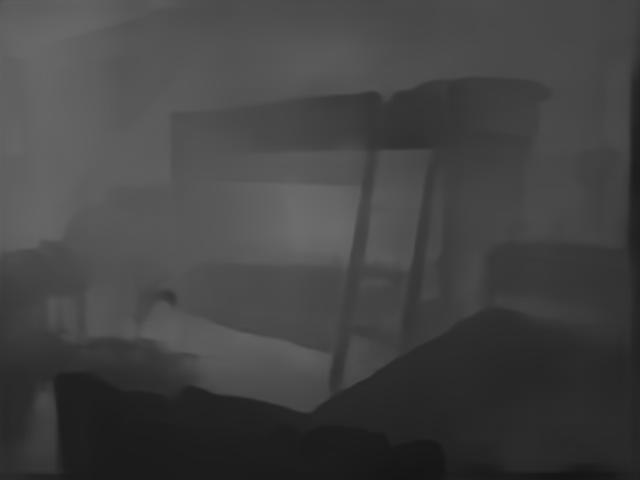}
  \end{subfigure}
  \vspace{1mm}
  \begin{subfigure}[t]{0.24\linewidth}
    \includegraphics[width=\linewidth]{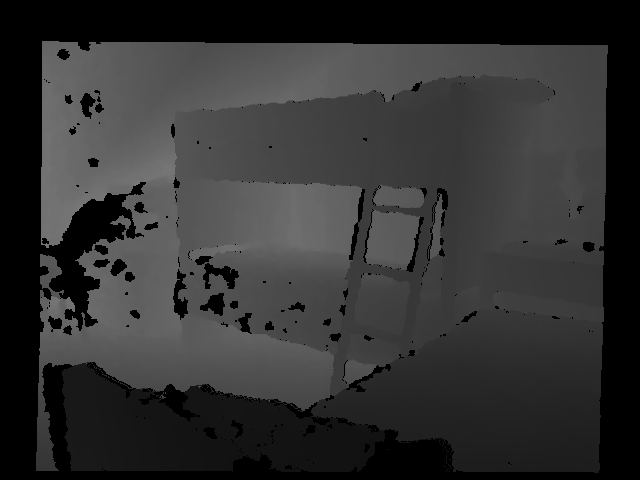}
  \end{subfigure}
   \begin{subfigure}[t]{0.24\linewidth}
    \includegraphics[width=\linewidth]{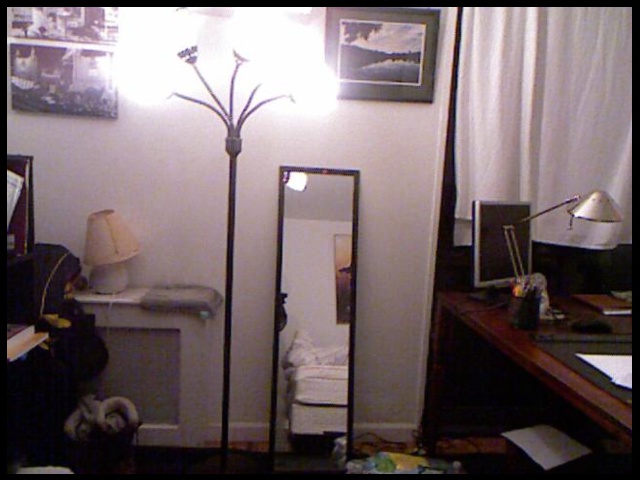}
    \caption*{Input image}
  \end{subfigure}
  \begin{subfigure}[t]{0.24\linewidth}
    \includegraphics[width=\linewidth]{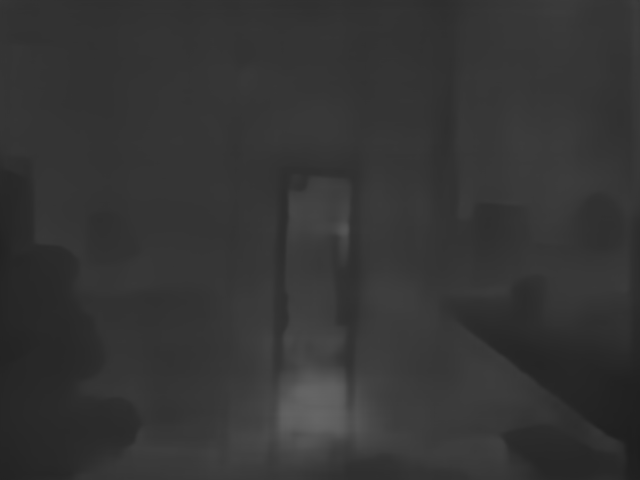}
    \caption*{G-BRS-sb}
  \end{subfigure}
  \begin{subfigure}[t]{0.24\linewidth}
    \includegraphics[width=\linewidth]{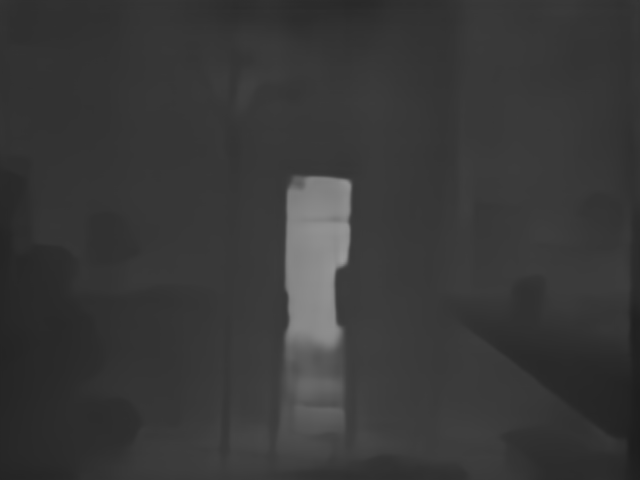}
    \caption*{G-BRS-bmconv}
  \end{subfigure}
  \begin{subfigure}[t]{0.24\linewidth}
    \includegraphics[width=\linewidth]{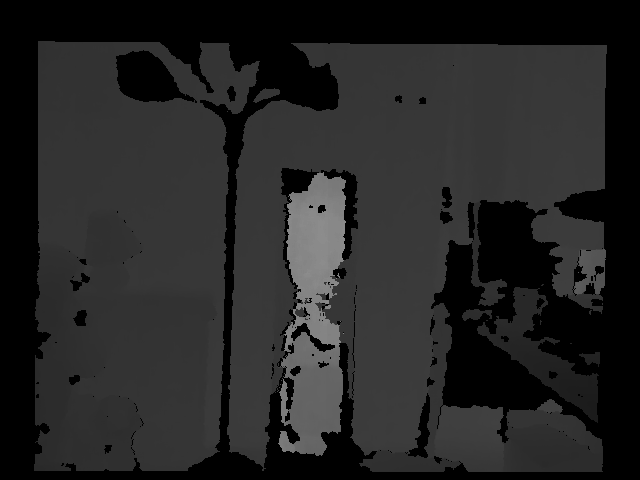}
    \caption*{Ground truth}
  \end{subfigure}
  \caption{Qualitative comparison between performance of the G-BRS-sb layer and our G-BRS-bmconv layer for the task of depth estimation on \textbf{NYU-Depth-V2}. Invalid regions are shown in black in the ground truth.}
  \label{fig:qualitative_compare_nyu}
  \vspace{-4mm}
\end{figure*}

To compare the performance of backpropagating refinement using our proposed G-BRS layers with the previously proposed channel-wise scale and bias as auxiliary variables by \textit{f}-BRS, we select the G-BRS-bmconv layer and the G-BRS-sb layer (channel-wise $s$ and $b$) for qualitative comparison. \\
\indent For interactive segmentation, we compare with the prior approach \textit{f}-BRS directly as the G-BRS-sb layer for this task is equivalent to the solution proposed by \textit{f}-BRS. Figure \ref{fig:qualitative_compare_sbd} shows that the G-BRS-bmconv layer can make more detailed refinement. Since the original \textit{f}-BRS was not implemented for the other applications, we make the best comparison possible by comparing the proposed G-BRS-bmconv layer with the G-BRS-sb layer. We emphasize that the specific G-BRS configurations for various architectures as well as the utilized consistency loss are part of our contribution. \\
\indent For semantic segmentation, we use 3 G-BRS layers for both G-BRS-sb and G-BRS-bmconv for the best performance as shown in Tables \ref{tab:table_scores_cityscapes}-\ref{tab:table_scores_mapillary}. Figure \ref{fig:qualitative_compare_cityscapes} shows that both G-BRS-sb and G-BRS-bmconv achieve high accuracy on Cityscapes while the G-BRS-bmconv layer is capable of making more detailed refinement. Figure \ref{fig:qualitative_compare_mapillary} shows that in a more challenging semantic segmentation dataset (Mapillary Vista) with 65 classes, the G-BRS-bmconv layer outperforms the G-BRS-sb layer with a larger margin. For image matting, we use 3 G-BRS layers as well for the best performance (Figure \ref{fig:mt_supp}). Figure \ref{fig:qualitative_compare_comp1k} shows that the G-BRS-bmconv layer is capable of refining alpha matte at a more detailed level. For depth estimation, Figure \ref{fig:qualitative_compare_nyu} shows that the G-BRS-sb layer is more susceptible to undesired global error while the G-BRS-bmconv layer produces better depth maps with both global and localized accuracy.

\renewcommand{\arraystretch}{1.25}
\begin{table*}[t]
  \small
  \centering
  \begin{tabular}{c|c|c}
  \toprule
\#Layers& 
\begin{tabular}{@{}c@{}}w $\mathcal{L}_{c}$\\\hline
	\begin{tabular}{P{1.4cm}|P{1cm}|P{1cm}|P{1.2cm}|P{1cm}}
		RGB-BRS&sb&bmsb&bmsb-m&bmconv
	\end{tabular}
\end{tabular}& 
\begin{tabular}{@{}c@{}}w/o $\mathcal{L}_{c}$\\\hline
	\begin{tabular}{P{1.4cm}|P{1cm}|P{1cm}|P{1.2cm}|P{1cm}}
		RGB-BRS&sb&bmsb&bmsb-m&bmconv
	\end{tabular}
\end{tabular}\\\hline
\midrule
L0 & \begin{tabular}{P{1.4cm}|P{1cm}|P{1cm}|P{1.2cm}|P{1cm}}$0.1\cdot{\frac{1}{2}^9}$&-&-&-&-\end{tabular} & \begin{tabular}{P{1.4cm}|P{1cm}|P{1cm}|P{1.2cm}|P{1cm}}$0.1\cdot{\frac{1}{2}^9}$&-&-&-&-\end{tabular} \\\hline
L1 & \begin{tabular}{P{1.4cm}|P{1cm}|P{1cm}|P{1.2cm}|P{1cm}}-&$0.1\cdot{\frac{1}{2}^1}$&$0.1\cdot{\frac{1}{2}^4}$&$0.1\cdot{\frac{1}{2}^2}$&$0.1\cdot{\frac{1}{2}^9}$\end{tabular} & \begin{tabular}{P{1.4cm}|P{1cm}|P{1cm}|P{1.2cm}|P{1cm}}-&$0.1\cdot{\frac{1}{2}^6}$&$0.1\cdot{\frac{1}{2}^5}$&$0.1\cdot{\frac{1}{2}^4}$&$0.1\cdot{\frac{1}{2}^9}$\end{tabular} \\\hline
  \end{tabular}
\caption{Learning rate settings for the backpropagating refinement layouts on \textbf{SBD}.}
\label{tab:table_lr_sbd}
\end{table*}
\begin{table*}[t]
  \small
  \centering
  \begin{tabular}{c|c|c}
  \toprule
\#Layers& 
\begin{tabular}{@{}c@{}}w $\mathcal{L}_{c}$\\\hline
	\begin{tabular}{P{1.4cm}|P{1cm}|P{1cm}|P{1.2cm}|P{1cm}}
		RGB-BRS&sb&bmsb&bmsb-m&bmconv
	\end{tabular}
\end{tabular}& 
\begin{tabular}{@{}c@{}}w/o $\mathcal{L}_{c}$\\\hline
	\begin{tabular}{P{1.4cm}|P{1cm}|P{1cm}|P{1.2cm}|P{1cm}}
		RGB-BRS&sb&bmsb&bmsb-m&bmconv
	\end{tabular}
\end{tabular}\\\hline
\midrule
L0 & \begin{tabular}{P{1.4cm}|P{1cm}|P{1cm}|P{1.2cm}|P{1cm}}$0.1\cdot{\frac{1}{2}^8}$&-&-&-&-\end{tabular} & \begin{tabular}{P{1.4cm}|P{1cm}|P{1cm}|P{1.2cm}|P{1cm}}$0.1\cdot{\frac{1}{2}^9}$&-&-&-&-\end{tabular} \\\hline
L1 & \begin{tabular}{P{1.4cm}|P{1cm}|P{1cm}|P{1.2cm}|P{1cm}}-&$0.1\cdot{\frac{1}{2}^3}$&$0.1\cdot{\frac{1}{2}^4}$&$0.1\cdot{\frac{1}{2}^3}$&$0.1\cdot{\frac{1}{2}^7}$\end{tabular} & \begin{tabular}{P{1.4cm}|P{1cm}|P{1cm}|P{1.2cm}|P{1cm}}-&$0.1\cdot{\frac{1}{2}^4}$&$0.1\cdot{\frac{1}{2}^5}$&$0.1\cdot{\frac{1}{2}^4}$&$0.1\cdot{\frac{1}{2}^9}$\end{tabular} \\\hline
L2 & \begin{tabular}{P{1.4cm}|P{1cm}|P{1cm}|P{1.2cm}|P{1cm}}-&$0.1\cdot{\frac{1}{2}^4}$&$0.1\cdot{\frac{1}{2}^4}$&$0.1\cdot{\frac{1}{2}^2}$&$0.1\cdot{\frac{1}{2}^7}$\end{tabular} & \begin{tabular}{P{1.4cm}|P{1cm}|P{1cm}|P{1.2cm}|P{1cm}}-&$0.1\cdot{\frac{1}{2}^6}$&$0.1\cdot{\frac{1}{2}^6}$&$0.1\cdot{\frac{1}{2}^5}$&$0.1\cdot{\frac{1}{2}^9}$\end{tabular} \\\hline
L3 & \begin{tabular}{P{1.4cm}|P{1cm}|P{1cm}|P{1.2cm}|P{1cm}}-&$0.1\cdot{\frac{1}{2}^4}$&$0.1\cdot{\frac{1}{2}^4}$&$0.1\cdot{\frac{1}{2}^3}$&$0.1\cdot{\frac{1}{2}^8}$\end{tabular} & \begin{tabular}{P{1.4cm}|P{1cm}|P{1cm}|P{1.2cm}|P{1cm}}-&$0.1\cdot{\frac{1}{2}^6}$&$0.1\cdot{\frac{1}{2}^7}$&$0.1\cdot{\frac{1}{2}^5}$&$0.1\cdot{\frac{1}{2}^9}$\end{tabular} \\\hline
  \end{tabular}
\caption{Learning rate settings for the backpropagating refinement layouts on \textbf{Cityscapes}.}
\label{tab:table_lr_cityscapes}
\end{table*}
\begin{table*}[t]
  \small
  \centering
  \begin{tabular}{c|c|c}
  \toprule
\#Layers& 
\begin{tabular}{@{}c@{}}w $\mathcal{L}_{c}$\\\hline
	\begin{tabular}{P{1.4cm}|P{1cm}|P{1cm}|P{1.2cm}|P{1cm}}
		RGB-BRS&sb&bmsb&bmsb-m&bmconv
	\end{tabular}
\end{tabular}& 
\begin{tabular}{@{}c@{}}w/o $\mathcal{L}_{c}$\\\hline
	\begin{tabular}{P{1.4cm}|P{1cm}|P{1cm}|P{1.2cm}|P{1cm}}
		RGB-BRS&sb&bmsb&bmsb-m&bmconv
	\end{tabular}
\end{tabular}\\\hline
\midrule
L0 & \begin{tabular}{P{1.4cm}|P{1cm}|P{1cm}|P{1.2cm}|P{1cm}}$0.1\cdot{\frac{1}{2}^4}$&-&-&-&-\end{tabular} & \begin{tabular}{P{1.4cm}|P{1cm}|P{1cm}|P{1.2cm}|P{1cm}}$0.1\cdot{\frac{1}{2}^9}$&-&-&-&-\end{tabular} \\\hline
L1 & \begin{tabular}{P{1.4cm}|P{1cm}|P{1cm}|P{1.2cm}|P{1cm}}-&$0.1\cdot{\frac{1}{2}^2}$&$0.1\cdot{\frac{1}{2}^3}$&$0.1\cdot{\frac{1}{2}^3}$&$0.1\cdot{\frac{1}{2}^7}$\end{tabular} & \begin{tabular}{P{1.4cm}|P{1cm}|P{1cm}|P{1.2cm}|P{1cm}}-&$0.1\cdot{\frac{1}{2}^3}$&$0.1\cdot{\frac{1}{2}^5}$&$0.1\cdot{\frac{1}{2}^3}$&$0.1\cdot{\frac{1}{2}^9}$\end{tabular} \\\hline
L2 & \begin{tabular}{P{1.4cm}|P{1cm}|P{1cm}|P{1.2cm}|P{1cm}}-&$0.1\cdot{\frac{1}{2}^2}$&$0.1\cdot{\frac{1}{2}^3}$&$0.1\cdot{\frac{1}{2}^2}$&$0.1\cdot{\frac{1}{2}^7}$\end{tabular} & \begin{tabular}{P{1.4cm}|P{1cm}|P{1cm}|P{1.2cm}|P{1cm}}-&$0.1\cdot{\frac{1}{2}^4}$&$0.1\cdot{\frac{1}{2}^6}$&$0.1\cdot{\frac{1}{2}^5}$&$0.1\cdot{\frac{1}{2}^9}$\end{tabular} \\\hline
L3 & \begin{tabular}{P{1.4cm}|P{1cm}|P{1cm}|P{1.2cm}|P{1cm}}-&$0.1\cdot{\frac{1}{2}^3}$&$0.1\cdot{\frac{1}{2}^4}$&$0.1\cdot{\frac{1}{2}^3}$&$0.1\cdot{\frac{1}{2}^8}$\end{tabular} & \begin{tabular}{P{1.4cm}|P{1cm}|P{1cm}|P{1.2cm}|P{1cm}}-&$0.1\cdot{\frac{1}{2}^5}$&$0.1\cdot{\frac{1}{2}^6}$&$0.1\cdot{\frac{1}{2}^5}$&$0.1\cdot{\frac{1}{2}^9}$\end{tabular} \\\hline
  \end{tabular}
\caption{Learning rate settings for the backpropagating refinement layouts on \textbf{Mapillary Vista}.}
\label{tab:table_lr_mapillary}
\end{table*}
\begin{table*}[t]
  \small
  \centering
  \begin{tabular}{c|c|c}
  \toprule
\#Layers& 
\begin{tabular}{@{}c@{}}w $\mathcal{L}_{c}$\\\hline
	\begin{tabular}{P{1.4cm}|P{1cm}|P{1cm}|P{1.2cm}|P{1cm}}
		RGB-BRS&sb&bmsb&bmsb-m&bmconv
	\end{tabular}
\end{tabular}& 
\begin{tabular}{@{}c@{}}w/o $\mathcal{L}_{c}$\\\hline
	\begin{tabular}{P{1.4cm}|P{1cm}|P{1cm}|P{1.2cm}|P{1cm}}
		RGB-BRS&sb&bmsb&bmsb-m&bmconv
	\end{tabular}
\end{tabular}\\\hline
\midrule
L0 & \begin{tabular}{P{1.4cm}|P{1cm}|P{1cm}|P{1.2cm}|P{1cm}}$0.1\cdot{\frac{1}{2}^9}$&-&-&-&-\end{tabular} & \begin{tabular}{P{1.4cm}|P{1cm}|P{1cm}|P{1.2cm}|P{1cm}}$0.1\cdot{\frac{1}{2}^9}$&-&-&-&-\end{tabular} \\\hline
L1 & \begin{tabular}{P{1.4cm}|P{1cm}|P{1cm}|P{1.2cm}|P{1cm}}-&$0.1\cdot{\frac{1}{2}^1}$&$0.1\cdot{\frac{1}{2}^4}$&$0.1\cdot{\frac{1}{2}^4}$&$0.1\cdot{\frac{1}{2}^7}$\end{tabular} & \begin{tabular}{P{1.4cm}|P{1cm}|P{1cm}|P{1.2cm}|P{1cm}}-&$0.1\cdot{\frac{1}{2}^7}$&$0.1\cdot{\frac{1}{2}^5}$&$0.1\cdot{\frac{1}{2}^5}$&$0.1\cdot{\frac{1}{2}^9}$\end{tabular} \\\hline
L2 & \begin{tabular}{P{1.4cm}|P{1cm}|P{1cm}|P{1.2cm}|P{1cm}}-&$0.1\cdot{\frac{1}{2}^2}$&$0.1\cdot{\frac{1}{2}^5}$&$0.1\cdot{\frac{1}{2}^4}$&$0.1\cdot{\frac{1}{2}^8}$\end{tabular} & \begin{tabular}{P{1.4cm}|P{1cm}|P{1cm}|P{1.2cm}|P{1cm}}-&$0.1\cdot{\frac{1}{2}^8}$&$0.1\cdot{\frac{1}{2}^6}$&$0.1\cdot{\frac{1}{2}^6}$&$0.1\cdot{\frac{1}{2}^8}$\end{tabular} \\\hline
L3 & \begin{tabular}{P{1.4cm}|P{1cm}|P{1cm}|P{1.2cm}|P{1cm}}-&$0.1\cdot{\frac{1}{2}^3}$&$0.1\cdot{\frac{1}{2}^6}$&$0.1\cdot{\frac{1}{2}^5}$&$0.1\cdot{\frac{1}{2}^9}$\end{tabular} & \begin{tabular}{P{1.4cm}|P{1cm}|P{1cm}|P{1.2cm}|P{1cm}}-&$0.1\cdot{\frac{1}{2}^9}$&$0.1\cdot{\frac{1}{2}^6}$&$0.1\cdot{\frac{1}{2}^6}$&$0.1\cdot{\frac{1}{2}^9}$\end{tabular} \\\hline
  \end{tabular}
\caption{Learning rate settings for the backpropagating refinement layouts on \textbf{Composition-1k}.}
\label{tab:table_lr_comp1k}
\end{table*}
\begin{table*}[t]
  \small
  \centering
  \begin{tabular}{c|c|c}
  \toprule
\#Layers& 
\begin{tabular}{@{}c@{}}w $\mathcal{L}_{c}$\\\hline
	\begin{tabular}{P{1.4cm}|P{1cm}|P{1cm}|P{1.2cm}|P{1cm}}
		RGB-BRS&sb&bmsb&bmsb-m&bmconv
	\end{tabular}
\end{tabular}& 
\begin{tabular}{@{}c@{}}w/o $\mathcal{L}_{c}$\\\hline
	\begin{tabular}{P{1.4cm}|P{1cm}|P{1cm}|P{1.2cm}|P{1cm}}
		RGB-BRS&sb&bmsb&bmsb-m&bmconv
	\end{tabular}
\end{tabular}\\\hline
\midrule
L0 & \begin{tabular}{P{1.4cm}|P{1cm}|P{1cm}|P{1.2cm}|P{1cm}}$0.1\cdot{\frac{1}{2}^7}$&-&-&-&-\end{tabular} & \begin{tabular}{P{1.4cm}|P{1cm}|P{1cm}|P{1.2cm}|P{1cm}}$0.1\cdot{\frac{1}{2}^8}$&-&-&-&-\end{tabular} \\\hline
L1 & \begin{tabular}{P{1.4cm}|P{1cm}|P{1cm}|P{1.2cm}|P{1cm}}-&$0.1\cdot{\frac{1}{2}^2}$&$0.1\cdot{\frac{1}{2}^3}$&$0.1\cdot{\frac{1}{2}^2}$&$0.1\cdot{\frac{1}{2}^7}$\end{tabular} & \begin{tabular}{P{1.4cm}|P{1cm}|P{1cm}|P{1.2cm}|P{1cm}}-&$0.1\cdot{\frac{1}{2}^2}$&$0.1\cdot{\frac{1}{2}^3}$&$0.1\cdot{\frac{1}{2}^2}$&$0.1\cdot{\frac{1}{2}^7}$\end{tabular} \\\hline
  \end{tabular}
\caption{Learning rate settings for the backpropagating refinement layouts on \textbf{NYU-Depth-V2}.}
\label{tab:table_lr_nyu}
\end{table*}
\begin{table*}[t]
  \small
  \centering
  \begin{tabular}{c|c|c}
  \toprule
\#Layers& 
\begin{tabular}{@{}c@{}}w $\mathcal{L}_{c}$\\\hline
	\begin{tabular}{P{1.4cm}|P{0.9cm}|P{1cm}|P{1.2cm}|P{0.9cm}}
		RGB-BRS&sb&bmsb&bmsb-m&bmconv
	\end{tabular}
\end{tabular}& 
\begin{tabular}{@{}c@{}}w/o $\mathcal{L}_{c}$\\\hline
	\begin{tabular}{P{1.4cm}|P{0.9cm}|P{1cm}|P{1.2cm}|P{0.9cm}}
		RGB-BRS&sb&bmsb&bmsb-m&bmconv
	\end{tabular}
\end{tabular}\\\hline
\midrule
L0\ (AUC) & \begin{tabular}{P{1.4cm}|P{0.9cm}|P{1cm}|P{1.2cm}|P{0.9cm}}0.8506&-&-&-&-\end{tabular} & \begin{tabular}{P{1.4cm}|P{0.9cm}|P{1cm}|P{1.2cm}|P{0.9cm}}0.8251&-&-&-&-\end{tabular} \\\hline
L1\ (AUC) & \begin{tabular}{P{1.4cm}|P{0.9cm}|P{1cm}|P{1.2cm}|P{0.9cm}}-&0.8521&0.8591&0.8589&0.8594\end{tabular} & \begin{tabular}{P{1.4cm}|P{0.9cm}|P{1cm}|P{1.2cm}|P{0.9cm}}-&0.8427&0.8529&0.8457&0.8322\end{tabular} \\\hline
\midrule
L0\ ($\text{mIoU}^*$) & \begin{tabular}{P{1.4cm}|P{0.9cm}|P{1cm}|P{1.2cm}|P{0.9cm}}0.9051&-&-&-&-\end{tabular} & \begin{tabular}{P{1.4cm}|P{0.9cm}|P{1cm}|P{1.2cm}|P{0.9cm}}0.8605&-&-&-&-\end{tabular} \\\hline
L1\ ($\text{mIoU}^*$) & \begin{tabular}{P{1.4cm}|P{0.9cm}|P{1cm}|P{1.2cm}|P{0.9cm}}-&0.9083&0.9162&0.9169&0.9181\end{tabular} & \begin{tabular}{P{1.4cm}|P{0.9cm}|P{1cm}|P{1.2cm}|P{0.9cm}}-&0.8931&0.9117&0.9052&0.8867\end{tabular} \\\hline
  \end{tabular}
\caption{Area Under Curve (AUC) computed using mIoU and maximum mIoU achieved in total number of clicks ($\text{mIoU}^*$) on \textbf{SBD} for all backpropagating refinement settings.}
\label{tab:table_scores_sbd}
\end{table*}
\begin{table*}[t]
  \small
  \centering
  \begin{tabular}{c|c|c}
  \toprule
\#Layers& 
\begin{tabular}{@{}c@{}}w $\mathcal{L}_{c}$\\\hline
	\begin{tabular}{P{1.4cm}|P{0.9cm}|P{1cm}|P{1.2cm}|P{0.9cm}}
		RGB-BRS&sb&bmsb&bmsb-m&bmconv
	\end{tabular}
\end{tabular}& 
\begin{tabular}{@{}c@{}}w/o $\mathcal{L}_{c}$\\\hline
	\begin{tabular}{P{1.4cm}|P{0.9cm}|P{1cm}|P{1.2cm}|P{0.9cm}}
		RGB-BRS&sb&bmsb&bmsb-m&bmconv
	\end{tabular}
\end{tabular}\\\hline
\midrule
L0\ (AUC) & \begin{tabular}{P{1.4cm}|P{0.9cm}|P{1cm}|P{1.2cm}|P{0.9cm}}0.8820&-&-&-&-\end{tabular} & \begin{tabular}{P{1.4cm}|P{0.9cm}|P{1cm}|P{1.2cm}|P{0.9cm}}0.8678&-&-&-&-\end{tabular} \\\hline
L1\ (AUC) & \begin{tabular}{P{1.4cm}|P{0.9cm}|P{1cm}|P{1.2cm}|P{0.9cm}}-&0.8847&0.8909&0.8893&0.8964\end{tabular} & \begin{tabular}{P{1.4cm}|P{0.9cm}|P{1cm}|P{1.2cm}|P{0.9cm}}-&0.8787&0.8849&0.8800&0.8889\end{tabular} \\\hline
L2\ (AUC) & \begin{tabular}{P{1.4cm}|P{0.9cm}|P{1cm}|P{1.2cm}|P{0.9cm}}-&0.8870&0.8924&0.8912&0.8962\end{tabular} & \begin{tabular}{P{1.4cm}|P{0.9cm}|P{1cm}|P{1.2cm}|P{0.9cm}}-&0.8806&0.8859&0.8826&0.8856\end{tabular} \\\hline
L3\ (AUC) & \begin{tabular}{P{1.4cm}|P{0.9cm}|P{1cm}|P{1.2cm}|P{0.9cm}}-&0.8872&0.8932&0.8944&0.8966\end{tabular} & \begin{tabular}{P{1.4cm}|P{0.9cm}|P{1cm}|P{1.2cm}|P{0.9cm}}-&0.8795&0.8862&0.8826&0.8733\end{tabular} \\\hline
\midrule
L0\ ($\text{mIoU}^*$) & \begin{tabular}{P{1.4cm}|P{0.9cm}|P{1cm}|P{1.2cm}|P{0.9cm}}0.8965&-&-&-&-\end{tabular} & \begin{tabular}{P{1.4cm}|P{0.9cm}|P{1cm}|P{1.2cm}|P{0.9cm}}0.8873&-&-&-&-\end{tabular} \\\hline
L1\ ($\text{mIoU}^*$) & \begin{tabular}{P{1.4cm}|P{0.9cm}|P{1cm}|P{1.2cm}|P{0.9cm}}-&0.8943&0.9019&0.9003&0.9080\end{tabular} & \begin{tabular}{P{1.4cm}|P{0.9cm}|P{1cm}|P{1.2cm}|P{0.9cm}}-&0.8917&0.8966&0.8936&0.9028\end{tabular} \\\hline
L2\ ($\text{mIoU}^*$) & \begin{tabular}{P{1.4cm}|P{0.9cm}|P{1cm}|P{1.2cm}|P{0.9cm}}-&0.8986&0.9049&0.9005&0.9070\end{tabular} & \begin{tabular}{P{1.4cm}|P{0.9cm}|P{1cm}|P{1.2cm}|P{0.9cm}}-&0.8924&0.8996&0.9006&0.9055\end{tabular} \\\hline
L3\ ($\text{mIoU}^*$) & \begin{tabular}{P{1.4cm}|P{0.9cm}|P{1cm}|P{1.2cm}|P{0.9cm}}-&0.9002&0.9049&0.9055&0.9083\end{tabular} & \begin{tabular}{P{1.4cm}|P{0.9cm}|P{1cm}|P{1.2cm}|P{0.9cm}}-&0.8973&0.9006&0.9011&0.9000\end{tabular} \\\hline
  \end{tabular}
\caption{Area Under Curve (AUC) computed using mIoU and maximum mIoU achieved in total number of clicks ($\text{mIoU}^*$) on \textbf{Cityscapes} for all backpropagating refinement settings.}
\label{tab:table_scores_cityscapes}
\end{table*}
\begin{table*}[t]
  \small
  \centering
  \begin{tabular}{c|c|c}
  \toprule
\#Layers& 
\begin{tabular}{@{}c@{}}w $\mathcal{L}_{c}$\\\hline
	\begin{tabular}{P{1.4cm}|P{0.9cm}|P{1cm}|P{1.2cm}|P{0.9cm}}
		RGB-BRS&sb&bmsb&bmsb-m&bmconv
	\end{tabular}
\end{tabular}& 
\begin{tabular}{@{}c@{}}w/o $\mathcal{L}_{c}$\\\hline
	\begin{tabular}{P{1.4cm}|P{0.9cm}|P{1cm}|P{1.2cm}|P{0.9cm}}
		RGB-BRS&sb&bmsb&bmsb-m&bmconv
	\end{tabular}
\end{tabular}\\\hline
\midrule
L0\ (AUC) & \begin{tabular}{P{1.4cm}|P{0.9cm}|P{1cm}|P{1.2cm}|P{0.9cm}}0.6752&-&-&-&-\end{tabular} & \begin{tabular}{P{1.4cm}|P{0.9cm}|P{1cm}|P{1.2cm}|P{0.9cm}}0.6543&-&-&-&-\end{tabular} \\\hline
L1\ (AUC) & \begin{tabular}{P{1.4cm}|P{0.9cm}|P{1cm}|P{1.2cm}|P{0.9cm}}-&0.7517&0.7521&0.7481&0.7680\end{tabular} & \begin{tabular}{P{1.4cm}|P{0.9cm}|P{1cm}|P{1.2cm}|P{0.9cm}}-&0.7242&0.7154&0.7116&0.7270\end{tabular} \\\hline
L2\ (AUC) & \begin{tabular}{P{1.4cm}|P{0.9cm}|P{1cm}|P{1.2cm}|P{0.9cm}}-&0.7624&0.7659&0.7640&0.7774\end{tabular} & \begin{tabular}{P{1.4cm}|P{0.9cm}|P{1cm}|P{1.2cm}|P{0.9cm}}-&0.7360&0.7301&0.7308&0.7420\end{tabular} \\\hline
L3\ (AUC) & \begin{tabular}{P{1.4cm}|P{0.9cm}|P{1cm}|P{1.2cm}|P{0.9cm}}-&0.7676&0.7714&0.7708&0.7791\end{tabular} & \begin{tabular}{P{1.4cm}|P{0.9cm}|P{1cm}|P{1.2cm}|P{0.9cm}}-&0.7369&0.7391&0.7380&0.7109\end{tabular} \\\hline
\midrule
L0\ ($\text{mIoU}^*$) & \begin{tabular}{P{1.4cm}|P{0.9cm}|P{1cm}|P{1.2cm}|P{0.9cm}}0.7266&-&-&-&-\end{tabular} & \begin{tabular}{P{1.4cm}|P{0.9cm}|P{1cm}|P{1.2cm}|P{0.9cm}}0.7169&-&-&-&-\end{tabular} \\\hline
L1\ ($\text{mIoU}^*$) & \begin{tabular}{P{1.4cm}|P{0.9cm}|P{1cm}|P{1.2cm}|P{0.9cm}}-&0.7896&0.7918&0.7909&0.8148\end{tabular} & \begin{tabular}{P{1.4cm}|P{0.9cm}|P{1cm}|P{1.2cm}|P{0.9cm}}-&0.7802&0.7620&0.7563&0.7795\end{tabular} \\\hline
L2\ ($\text{mIoU}^*$) & \begin{tabular}{P{1.4cm}|P{0.9cm}|P{1cm}|P{1.2cm}|P{0.9cm}}-&0.8034&0.8092&0.8048&0.8213\end{tabular} & \begin{tabular}{P{1.4cm}|P{0.9cm}|P{1cm}|P{1.2cm}|P{0.9cm}}-&0.7978&0.7859&0.7851&0.8019\end{tabular} \\\hline
L3\ ($\text{mIoU}^*$) & \begin{tabular}{P{1.4cm}|P{0.9cm}|P{1cm}|P{1.2cm}|P{0.9cm}}-&0.8101&0.8153&0.8154&0.8215\end{tabular} & \begin{tabular}{P{1.4cm}|P{0.9cm}|P{1cm}|P{1.2cm}|P{0.9cm}}-&0.7967&0.7919&0.7936&0.7868\end{tabular} \\\hline
  \end{tabular}
\caption{Area Under Curve (AUC) computed using mIoU and maximum mIoU achieved in total number of clicks ($\text{mIoU}^*$) on \textbf{Mapillary Vista} for all backpropagating refinement settings.}
\label{tab:table_scores_mapillary}
\end{table*}
\renewcommand{\arraystretch}{1}
\renewcommand{\arraystretch}{1.25}
\begin{table*}[t]
  \small
  \centering
  \begin{tabular}{P{2.1cm}|c|c}
  \toprule
Metrics& 
\begin{tabular}{@{}c@{}}w $\mathcal{L}_{c}$\\\hline
	\begin{tabular}{P{1.4cm}|P{0.8cm}|P{0.9cm}|P{1.2cm}|P{0.9cm}}
		RGB-BRS&sb&bmsb&bmsb-m&bmconv
	\end{tabular}
\end{tabular}& 
\begin{tabular}{@{}c@{}}w/o $\mathcal{L}_{c}$\\\hline
	\begin{tabular}{P{1.4cm}|P{0.8cm}|P{0.9cm}|P{1.2cm}|P{0.9cm}}
		RGB-BRS&sb&bmsb&bmsb-m&bmconv
	\end{tabular}
\end{tabular}\\\hline
\midrule
$\text{AUC}_{\delta_1} \uparrow$& \begin{tabular}{P{1.4cm}|P{0.8cm}|P{0.9cm}|P{1.2cm}|P{0.9cm}}0.9610&0.9554&0.9560&0.9555&0.9628\end{tabular} & \begin{tabular}{P{1.4cm}|P{0.8cm}|P{0.9cm}|P{1.2cm}|P{0.9cm}}0.9592&0.9549&0.9556&0.9552&0.9623\end{tabular} \\\hline
$\text{AUC}_{\delta_2} \uparrow$& \begin{tabular}{P{1.4cm}|P{0.8cm}|P{0.9cm}|P{1.2cm}|P{0.9cm}}0.9914&0.9904&0.9906&0.9902&0.9917\end{tabular} & \begin{tabular}{P{1.4cm}|P{0.8cm}|P{0.9cm}|P{1.2cm}|P{0.9cm}}0.9908&0.9900&0.9904&0.9901&0.9912\end{tabular} \\\hline
$\text{AUC}_{\delta_3} \uparrow$& \begin{tabular}{P{1.4cm}|P{0.8cm}|P{0.9cm}|P{1.2cm}|P{0.9cm}}0.9974&0.9972&0.9974&0.9972&0.9976\end{tabular} & \begin{tabular}{P{1.4cm}|P{0.8cm}|P{0.9cm}|P{1.2cm}|P{0.9cm}}0.9974&0.9971&0.9974&0.9972&0.9975\end{tabular} \\\hline
$\text{AUC}_{\text{AbsRel}} \downarrow$& \begin{tabular}{P{1.4cm}|P{0.8cm}|P{0.9cm}|P{1.2cm}|P{0.9cm}}0.0565&0.0636&0.0628&0.0633&0.0555\end{tabular} & \begin{tabular}{P{1.4cm}|P{0.8cm}|P{0.9cm}|P{1.2cm}|P{0.9cm}}0.0578&0.0640&0.0630&0.0636&0.0559\end{tabular} \\\hline
$\text{AUC}_{\text{SqRel}} \downarrow$& \begin{tabular}{P{1.4cm}|P{0.8cm}|P{0.9cm}|P{1.2cm}|P{0.9cm}}0.0289&0.0326&0.0321&0.0332&0.0281\end{tabular} & \begin{tabular}{P{1.4cm}|P{0.8cm}|P{0.9cm}|P{1.2cm}|P{0.9cm}}0.0299&0.0335&0.0325&0.0333&0.0289\end{tabular} \\\hline
$\text{AUC}_{\text{RMSE}} \downarrow$& \begin{tabular}{P{1.4cm}|P{0.8cm}|P{0.9cm}|P{1.2cm}|P{0.9cm}}0.2512&0.2722&0.2688&0.2726&0.2487\end{tabular} & \begin{tabular}{P{1.4cm}|P{0.8cm}|P{0.9cm}|P{1.2cm}|P{0.9cm}}0.2574&0.2741&0.2702&0.2733&0.2512\end{tabular} \\\hline
$\text{AUC}_{\text{RMSE }\textit{log}} \downarrow$& \begin{tabular}{P{1.4cm}|P{0.8cm}|P{0.9cm}|P{1.2cm}|P{0.9cm}}0.0861&0.0940&0.0927&0.0938&0.0849\end{tabular} & \begin{tabular}{P{1.4cm}|P{0.8cm}|P{0.9cm}|P{1.2cm}|P{0.9cm}}0.0880&0.0946&0.0931&0.0941&0.0857\end{tabular} \\\hline
\midrule
${{\delta}_{1}}^* \uparrow$& \begin{tabular}{P{1.4cm}|P{0.8cm}|P{0.9cm}|P{1.2cm}|P{0.9cm}}0.9830&0.9752&0.9765&0.9754&0.9833\end{tabular} & \begin{tabular}{P{1.4cm}|P{0.8cm}|P{0.9cm}|P{1.2cm}|P{0.9cm}}0.9821&0.9745&0.9761&0.9752&0.9827\end{tabular} \\\hline
${{\delta}_{2}}^* \uparrow$& \begin{tabular}{P{1.4cm}|P{0.8cm}|P{0.9cm}|P{1.2cm}|P{0.9cm}}0.9960&0.9942&0.9943&0.9940&0.9959\end{tabular} & \begin{tabular}{P{1.4cm}|P{0.8cm}|P{0.9cm}|P{1.2cm}|P{0.9cm}}0.9957&0.9941&0.9957&0.9942&0.9941\end{tabular} \\\hline
${{\delta}_{3}}^* \uparrow$& \begin{tabular}{P{1.4cm}|P{0.8cm}|P{0.9cm}|P{1.2cm}|P{0.9cm}}0.9987&0.9982&0.9983&0.9982&0.9986\end{tabular} & \begin{tabular}{P{1.4cm}|P{0.8cm}|P{0.9cm}|P{1.2cm}|P{0.9cm}}0.9988&0.9982&0.9986&0.9983&0.9982\end{tabular} \\\hline
$\text{AbsRel}^* \downarrow$& \begin{tabular}{P{1.4cm}|P{0.8cm}|P{0.9cm}|P{1.2cm}|P{0.9cm}}0.0375&0.0472&0.0450&0.0467&0.0372\end{tabular} & \begin{tabular}{P{1.4cm}|P{0.8cm}|P{0.9cm}|P{1.2cm}|P{0.9cm}}0.0384&0.0475&0.0453&0.0467&0.0378\end{tabular} \\\hline
$\text{SqRel}^* \downarrow$& \begin{tabular}{P{1.4cm}|P{0.8cm}|P{0.9cm}|P{1.2cm}|P{0.9cm}}0.0151&0.0210&0.0203&0.0221&0.0151\end{tabular} & \begin{tabular}{P{1.4cm}|P{0.8cm}|P{0.9cm}|P{1.2cm}|P{0.9cm}}0.0157&0.0213&0.0208&0.0214&0.0158\end{tabular} \\\hline
$\text{RMSE}^* \downarrow$& \begin{tabular}{P{1.4cm}|P{0.8cm}|P{0.9cm}|P{1.2cm}|P{0.9cm}}0.1858&0.2191&0.2131&0.2181&0.1876\end{tabular} & \begin{tabular}{P{1.4cm}|P{0.8cm}|P{0.9cm}|P{1.2cm}|P{0.9cm}}0.1894&0.2200&0.2134&0.2182&0.1898\end{tabular} \\\hline
${\text{RMSE }\textit{log}}^* \downarrow$& \begin{tabular}{P{1.4cm}|P{0.8cm}|P{0.9cm}|P{1.2cm}|P{0.9cm}}0.0627&0.0749&0.0726&0.0744&0.0627\end{tabular} & \begin{tabular}{P{1.4cm}|P{0.8cm}|P{0.9cm}|P{1.2cm}|P{0.9cm}}0.0639&0.0753&0.0729&0.0744&0.0635\end{tabular} \\\hline
  \end{tabular}
\caption{Area Under Curve (AUC) computed using various depth estimation metrics and maximum scores achieved in total number of clicks (denoted with "*") on \textbf{NYU-Depth-V2} for all backpropagating refinement settings. Since BTSNet has a G-BRS layout that only contains 1 insertion, we omit the \#layers in this table. For clarification, RGB-BRS does not utilize G-BRS layers and other G-BRS layout contains 1 G-BRS layer.}
\label{tab:table_scores_nyu}
\end{table*}
\begin{table*}[t]
  \small
  \centering
  \begin{tabular}{c|P{2.5cm}|P{2.5cm}|P{2.5cm}|P{2.5cm}|P{2.5cm}}
  \toprule
\#Layers&RGB-BRS&G-BRS-sb&G-BRS-bmsb&G-BRS-bmsb-m&G-BRS-bmconv\\\hline
\midrule
L0&1.542&-&-&-&-\\\hline
L1&-&0.687&0.601&0.450&0.583\\\hline
  \end{tabular}
\caption{SPC for all backpropagating refinement layouts with $\mathcal{L}_{c}$ on \textbf{SBD}.}
\label{tab:table_spc_sbd}
\end{table*}
\begin{table*}[t]
  \small
  \centering
  \begin{tabular}{c|P{2.5cm}|P{2.5cm}|P{2.5cm}|P{2.5cm}|P{2.5cm}}
  \toprule
\#Layers&RGB-BRS&G-BRS-sb&G-BRS-bmsb&G-BRS-bmsb-m&G-BRS-bmconv\\\hline
\midrule
L0&8.362&-&-&-&-\\\hline
L1&-&5.627&5.641&5.633&5.633\\\hline
L2&-&5.649&5.652&5.665&5.661\\\hline
L3&-&5.691&5.716&5.718&5.727\\\hline
  \end{tabular}
\caption{SPC for all backpropagating refinement layouts with $\mathcal{L}_{c}$ on \textbf{Cityscapes}.}
\label{tab:table_spc_cityscapes}
\end{table*}
\begin{table*}[t]
  \small
  \centering
  \begin{tabular}{c|P{2.5cm}|P{2.5cm}|P{2.5cm}|P{2.5cm}|P{2.5cm}}
  \toprule
\#Layers&RGB-BRS&G-BRS-sb&G-BRS-bmsb&G-BRS-bmsb-m&G-BRS-bmconv\\\hline
\midrule
L0&8.125&-&-&-&-\\\hline
L1&-&5.153&5.133&5.145&5.134\\\hline
L2&-&5.159&5.153&5.165&5.159\\\hline
L3&-&5.222&5.224&5.233&5.231\\\hline
  \end{tabular}
\caption{SPC for all backpropagating refinement layouts with $\mathcal{L}_{c}$ on \textbf{Mapillary Vista}.}
\label{tab:table_spc_mapillary}
\end{table*}
\begin{table*}[t]
  \small
  \centering
  \begin{tabular}{c|P{2.5cm}|P{2.5cm}|P{2.5cm}|P{2.5cm}|P{2.5cm}}
  \toprule
\#Layers&RGB-BRS&G-BRS-sb&G-BRS-bmsb&G-BRS-bmsb-m&G-BRS-bmconv\\\hline
\midrule
L0&2.473&-&-&-&-\\\hline
L1&-&1.350&1.338&1.347&1.344\\\hline
L2&-&1.348&1.346&1.359&1.355\\\hline
L3&-&1.360&1.367&1.388&1.375\\\hline
  \end{tabular}
\caption{SPC for all backpropagating refinement layouts with $\mathcal{L}_{c}$ on \textbf{Composition-1k}.}
\label{tab:table_spc_comp1k}
\end{table*}
\begin{table*}[t]
  \small
  \centering
  \begin{tabular}{c|P{2.5cm}|P{2.5cm}|P{2.5cm}|P{2.5cm}|P{2.5cm}}
  \toprule
\#Layers&RGB-BRS&G-BRS-sb&G-BRS-bmsb&G-BRS-bmsb-m&G-BRS-bmconv\\\hline
\midrule
L0&3.205&-&-&-&-\\\hline
L1&-&2.040&2.040&2.045&2.055\\\hline
  \end{tabular}
\caption{SPC for all backpropagating refinement layouts with $\mathcal{L}_{c}$ on \textbf{NYU-Depth-V2}.}
\label{tab:table_spc_nyu}
\end{table*}
\renewcommand{\arraystretch}{1}
\end{document}